\documentclass{article}
\usepackage[final]{colm2026_conference}

\usepackage{microtype}
\usepackage{hyperref}
\usepackage{url}
\usepackage{booktabs}
\usepackage{multirow}
\usepackage{graphicx}
\usepackage{subcaption}
\usepackage{amsmath}
\usepackage[table]{xcolor}
\usepackage{enumitem}
\usepackage{array}

\usepackage{lineno}
\usepackage{pifont}
\usepackage{wrapfig}
\usepackage{tcolorbox}
\newtcolorbox{promptbox}{
  colback=gray!5, colframe=gray!50,
  fontupper=\small\ttfamily,
  boxrule=0.4pt, arc=2pt, left=6pt, right=6pt, top=4pt, bottom=4pt
}

\definecolor{darkblue}{rgb}{0, 0, 0.5}
\hypersetup{colorlinks=true, citecolor=darkblue, linkcolor=darkblue, urlcolor=darkblue}

\title{Overconfidence and Calibration in Medical VQA: \\ Empirical Findings and Hallucination-Aware Mitigation}

\author{%
\begin{tabular}[t]{@{}p{0.50\textwidth}@{\hspace{0.05\textwidth}}p{0.45\textwidth}@{}}
\bf Ji Young Byun \newline
\mdseries Johns Hopkins University, School of Medicine \newline
\texttt{jbyun13@jhu.edu}
&
\bf Young-Jin Park \newline
\mdseries Massachusetts Institute of Technology \newline
\texttt{youngp@mit.edu}
\\
\\
\bf Jean-Philippe Corbeil \newline
\mdseries Microsoft Healthcare \& Life Sciences \newline
\texttt{jcorbeil@microsoft.com}
&
\bf Asma Ben Abacha \newline
\mdseries Microsoft Healthcare \& Life Sciences \newline
\texttt{abenabacha@microsoft.com}
\end{tabular}
}

\begin{document}

\ifcolmsubmission
\linenumbers
\fi

\maketitle

\vspace{-2em}
\begin{center}
\texttt{Project Page:} \url{https://jiyoungbyun.github.io/vlm-hac}
\end{center}
\vspace{1.5em}

\begin{abstract}

As vision-language models (VLMs) are increasingly deployed in clinical decision support, more than accuracy is required: knowing when to trust their predictions is equally critical. 
Yet, a comprehensive and systematic investigation into the overconfidence of these models remains notably scarce in the medical domain.
We address this gap through a comprehensive empirical study of confidence calibration in VLMs, spanning three model families (Qwen3-VL, InternVL3, LLaVA-NeXT), three model scales (2B--38B), and multiple confidence estimation prompting strategies, across three medical visual question answering (VQA) benchmarks. Our study yields three key findings:
First, overconfidence persists across model families and is not resolved by scaling or prompting, such as chain-of-thought and verbalized confidence variants. Second, simple post-hoc calibration approaches, such as Platt scaling, reduce calibration error and consistently outperform the prompt-based strategy. Third, due to their (strict) monotonicity, these post-hoc calibration methods are inherently limited in improving the discriminative quality of predictions, leaving AUROC at the same level. 
Motivated by these findings, we investigate hallucination-aware calibration (HAC), which incorporates vision-grounded hallucination detection signals as complementary inputs to refine confidence estimates. We find that leveraging these hallucination signals improves both calibration and AUROC, with the largest gains on open-ended questions. On closed-ended questions, format-specific signals such as the logit margin show more informative.
Overall, our findings suggest post-hoc calibration as standard practice for medical VLM deployment over raw confidence estimates, and highlight the practical usefulness of hallucination signals to enable more reliable use of VLMs in medical VQA.

\end{abstract}

\section{Introduction}
\label{sec:intro}

Vision–language models (VLMs) are increasingly being explored for medical applications such as clinical visual question answering (VQA)~\citep{zhang2024development, bae2023ehrxqa, yim2024dermavqa, li2023llava}, radiology report generation~\citep{jing2025reason, de2025padchest}, and diagnostic decision support~\citep{irvin2019chexpert, bustos2020padchest, chambon2024chexpert, zhang2023biomedclip, tu2024towards, byun2026test}. 

While confidence calibration has been extensively studied in the general domain~\citep{tian2023just, xiong2024llms, geng2024survey, liu2025uncertainty}, systematic investigation in medicine remains scarce. In practice, however, the alignment between a model’s confidence and its actual accuracy matters. A model that assigns 90\% confidence to an incorrect diagnosis may lead a clinician to forgo lab tests or delay appropriate treatment.

Recent efforts have begun to address this gap, but have focused on more complex interventions, such as reinforcement learning~\citep{kriz2025prompt4trust} or multi-agent debate~\citep{pandey2026refine}, while often neglecting a foundational examination of simple post-hoc calibration approaches. The most direct empirical study to date~\citep{du2025confidence} is limited to closed-ended questions, a single model scale (7B), verbalized confidence only, and its proposed remedy is itself prompt-based. As a result, the community lacks empirical grounding on even basic questions: Are modern VLMs well-calibrated in medical contexts? Does scaling or prompting improve reliability? And if not, what principled methodology should serve as the baseline for reliable medical VLM deployment?

We address these questions through a comprehensive empirical evaluation and identify three key findings. First, overconfidence is a systematic property across all models and scales (Section~\ref{sec:pervasive_overconfidence} and \ref{sec:scale}). Second, no prompting strategy---including Chain-of-Thought (CoT)~\citep{wei2022chain}, penalty-based~\citep{ni2024llms}, and multi-stage approaches~\citep{tian2023just,xiong2024llms}---robustly improves calibration (Section~\ref{sec:prompting}). Third, simple post-hoc calibration substantially reduces calibration error, suggesting the importance of recalibration as a baseline practice in medical VLM deployment.

Standard post-hoc methods apply monotonic transformations that preserve the ranking of predictions, making them inherently limited in their ability to enhance discriminative quality. This highlights a key limitation of monotonic post-hoc calibration methods: No matter how well the scores are recalibrated, a confidently incorrect prediction will always be ranked above a less confident but correct one. Overcoming this ceiling requires incorporating signals that are informative about prediction correctness yet complementary to confidence.

A separate line of work has developed vision-grounded hallucination detection methods for VLMs, including perturbation-based approaches validated on medical VQA~\citep{liao2025vision, gautam2025hedge}. Hallucination detection approaches capture whether a prediction is grounded in the input image. To our knowledge, their potential for refining confidence calibration has not been explored. This motivates our investigation of Hallucination-Aware Calibration (HAC), which incorporates hallucination detection signals to enable reranking of predictions, improving both calibration and discriminative quality.

In short, our key contributions are:
\begin{itemize}[itemsep=2pt, topsep=0pt, parsep=0pt, leftmargin=*]
    \item We conduct a comprehensive empirical study of confidence calibration in medical VQA, spanning three VLM families and model scales from 2B to 38B, both open-ended and closed-ended questions, and over eight confidence estimation strategies. We establish that overconfidence is pervasive and not mitigated by scaling or prompting.
    \item We demonstrate that simple post-hoc methods such as Platt scaling consistently outperform prompt engineering for VLM calibration, despite receiving comparatively little attention in the recent literature.
    \item We introduce HAC, a signal-agnostic post-hoc framework that incorporates hallucination detection signals to refine confidence estimates. HAC enhances discriminative quality particularly on open-ended questions; on closed-ended questions, substituting a format-appropriate signal (yes/no logit margin) recovers the gains, suggesting a promising direction for integrating complementary error signals into calibration.
    \item We provide practical observations and cost--quality tradeoffs for practitioners deploying medical VLMs, including ablations over hallucination proxies and sample budgets.
\end{itemize}

\section{Related work}
\label{sec:related}

\paragraph{Confidence Estimation in Generative Models.}
As generative models operate over an open-ended output space, confidence estimation becomes non-trivial. Several paradigms have been proposed. \emph{Sampling-based confidence} approximates predictive distribution via Monte Carlo estimation from multiple independent generations~\citep{neal1993probabilistic,wang2022self}; \citet{kuhn2023semantic} strengthened this via semantic entropy, which clusters semantically equivalent outputs before computing consistency scores, offering a principled solution to the open-ended output space problem. \emph{Logit-based confidence} extracts confidence from the output distribution over answer tokens~\citep{desai2020calibration, zhao2021calibrate}; while computationally efficient, it is largely limited to closed-form settings. \emph{Verbalized confidence} prompts the model to report a numeric score~\citep{kadavath2022language, tian2023just, xiong2024llms, ni2024llms}; it is simple and applicable to black-box models but provides no theoretical guarantee and may conflate instruction following with genuine uncertainty.

\paragraph{Hallucination Detection in Generative Models.} 
Hallucination is a failure mode in large language model (LLM)~\citep{xiao2021hallucination}: models produce false outputs that are unfaithful or factually incorrect. Prior work has shown that LLMs tend to produce semantically inconsistent outputs under uncertainty about the prompt~\citep{kuhn2023semantic}, and confident models are more likely to disregard adversarial hints paired with the input~\citep{yadkori2024believe}. However, hallucination detection has been studied primarily in language models, with VLMs remaining relatively underexplored. Recent work has begun to address this gap by proposing perturbation-based approaches that detect hallucination by contrasting output distributions before and after perturbing the input image, validating their methods on medical VQA benchmarks~\citep{liao2025vision, gautam2025hedge}.

\paragraph{Overconfidence in Medical VQA Benchmarks.} \label{sec:related:overconfidence_medical}
A growing body of work has adapted general VLMs to clinical tasks via domain-specific supervised fine-tuning~\citep{li2024llava, zhang2023pmc}, but calibration has received little attention. \cite{savage2024large} empirically show that medical LLMs often exhibit overconfidence, \cite{shen2025exposing} report an accuracy--reliability mismatch in pathology VQA; and \cite{du2025confidence} show that fine-tuning on radiology data improves accuracy while often degrading calibration.
Prior work has largely addressed this through prompt-based (i.e., verbalized) confidence estimation, whereas the role of post-hoc calibration methods in medical VLMs remains underexplored.

\section{Experimental Design}

\paragraph{Models \& Datasets.} We evaluate three open-source VLM families that are general-purpose yet have seen adoption in medical vision tasks: Qwen3-VL-Instruct (2B, 8B, and 32B)~\citep{Qwen-VL, qwen3technicalreport}, InternVL3-Instruct (2B, 8B, and 38B)~\citep{chen2024internvl}, and LLaVA-NeXT (7B and 34B)~\citep{liu2024llavanext}. This selection spans a representative range of scales (2B--38B) and architectures. We evaluate on three established medical VQA benchmarks: (1) VQA-RAD~\citep{lau2018vqarad} consists of closed- and open-ended questions on radiology images covering modality, plane, organ system, and abnormality, (2) SLAKE~\citep{liu2021slake} is a bilingual dataset\footnote{We only use English questions to ensure cross-dataset consistency.} on organ identification, position, modality, and abnormality across X-ray, CT, and MRI, and (3) VQA-Med-2019~\citep{ImageCLEFVQA-Med2019} contains radiology images with questions covering modality, plane, organ system, and abnormality. Detailed statistics are in Appendix~\ref{app:dataset_stats}.

To enable fair comparison across question types while avoiding dataset-specific imbalance, we construct a pooled evaluation set by aggregating samples from VQA-RAD, SLAKE-EN, and VQA-Med within each type and evaluating on their union. We report results on this pooled set in the main text, with per-dataset results provided in the Appendix.

\paragraph{Confidence Measurements.}
Let $\mathcal{X}$ and $\mathcal{Y}$ denote the input and output spaces, respectively.
Consider a model parameterized by $\theta$, which defines a conditional distribution: $P_\theta(y \mid x)$.
Given an input $x \in \mathcal{X}$, the corresponding predicted label is $\hat{y} = \arg\max_{y \in \mathcal{Y}} P_\theta(y \mid x).$
The goal is to estimate the confidence of the model prediction: $P_\theta(y = \hat{y} \mid x).$

Prior work typically adopts three standardized approaches: sampling-based and verbalized methods, as discussed in the previous section.
We use $N=20$ generations for the sampling approach, with an ablation study on $N$ in Appendix~\ref{sec:sample_size_ablation}.
In this paper, we explore over eight estimation strategies; we defer the details, including respective prompts, to Appendix~\ref{app:confidence}. 

For open-form questions, multiple linguistic expressions can have the same meaning. Instead of pattern matching, we cluster semantically equivalent outputs to a common canonical form~\citep{kuhn2023semantic, farquhar2024detecting}. See Appendix~\ref{app:cluster} for details.

\paragraph{Evaluation Metrics.} 

We assess calibration using the expected calibration error (ECE)~\citep{naeini2015obtaining, guo2017calibration}, which measures the average gap between confidence and accuracy across confidence bins.
We report adaptive calibration error (ACE)~\citep{nixon2019measuring} as well, which uses population-uniform bins to ensure more reliable estimation across all confidence ranges.
We additionally report Area Under the Receiver Operating Characteristic Curve (AUROC)~\citep{bradley1997use} to evaluate the confidence scores' diagnostic capability in distinguishing correct predictions. For AUROC, tied confidence scores are assigned the same threshold, equivalent to averaging over all possible orderings.
We apply the LLM-as-a-judge framework for accuracy evaluation. Details, including the judge prompt, agreement with larger judges, and human expert validation, are in Appendix~\ref{app:llm_judge}.

We remark that ECE can be inadequate for measuring the reliability of severely overconfident models. In this case, most predictions concentrate in a few bins near confidence $1.0$, making the effective size of the remaining bins too small to yield reliable estimates. While ECE serves reliably when confidence scores are broadly distributed across $[0, 1]$, ACE adaptively partitions predictions into equal-mass bins, ensuring each bin is sufficiently populated. Following the literature~\citep{nixon2019measuring, roelofs2022mitigating}, we use ACE as the primary metric for our analysis in the main body.

\section{Are VLMs Well-Calibrated for Medical VQA?} \label{sec:overconfidence}

While state-of-the-art VLMs have achieved notable performance in the general domain, recent works have highlighted overconfidence as a key failure mode~\citep{groot2024overconfidence, du2025confidence}. As many medical applications increasingly adopt such general-purpose models~\cite{gilal2025pathvlm, lee2025cxr, brin2025assessing}, understanding their reliability becomes critical. In safety-critical settings, poor calibration can lead to catastrophic decision making. However, calibration in medical VLMs remains underexplored.

To this end, we first investigate whether overconfidence remains a persistent issue in recent open-source models. Specifically, this analysis aims to address the following research questions: (1) Are general-purpose VLMs well-calibrated in medical contexts?, (2) Are larger models better calibrated?, (3) Do prompting strategies lead to different calibration qualities?

\subsection{Overconfidence is Pervasive Across Models in Medical VQA Benchmarks}
\label{sec:pervasive_overconfidence}

\begin{figure}[h!]
\centering
\begin{minipage}[c]{0.53\linewidth}
    \centering
    \includegraphics[width=\linewidth]{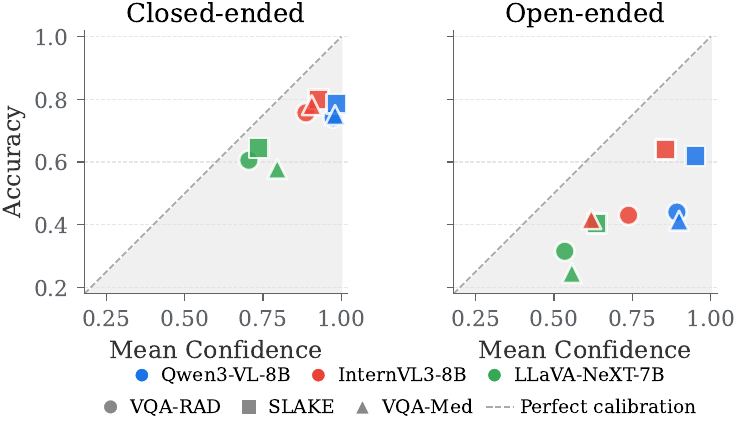}
    \captionof{figure}{Mean confidence vs.\ accuracy for different question types on medical VQA benchmarks (VQA-RAD, SLAKE-EN, VQA-Med). All models
    fall below the diagonal (grey region) across datasets, indicating consistent \textbf{overconfidence}.}
    \label{fig:confidence_accuracy}
\end{minipage}
\hfill
\begin{minipage}[c]{0.44\linewidth}
    \centering
    \scriptsize
    \setlength{\tabcolsep}{4pt}
    \captionof{table}{Accuracy, mean confidence, ECE, and ACE on the pooled medical VQA benchmarks. Gap $=$ Conf.\ $-$ Acc.\ denotes the overconfidence gap; all models exhibit systematic \textbf{overconfidence}, while scaling does not mitigate it.}
    \label{tab:scaling_main}
    \begin{tabular}{@{}lccccc@{}}
    \toprule
    \textbf{Model} & \textbf{Acc.} & \textbf{Conf.} & \textbf{Gap} & \textbf{ECE} & \textbf{ACE} \\
    \midrule
    Qwen3-VL-2B          & .562 & .860 & .298 & .316 & .305 \\
    Qwen3-VL-8B          & .605 & .943 & .338 & .340 & .338 \\
    Qwen3-VL-32B         & .651 & .954 & .304 & .307 & .303 \\
    \midrule
    InternVL3-2B         & .554 & .774 & .220 & .227 & .231 \\
    InternVL3-8B         & .618 & .811 & .192 & .204 & .202 \\
    InternVL3-38B        & .549 & .767 & .218 & .231 & .237 \\
    \midrule
    LLaVA-NeXT-7B        & .434 & .640 & .206 & .213 & .226 \\
    LLaVA-NeXT-34B       & .472 & .751 & .279 & .285 & .293 \\
    \bottomrule
    \end{tabular}
\end{minipage}

\end{figure}

Figure~\ref{fig:confidence_accuracy} shows mean confidence against accuracy for all models on the pooled VQA-RAD, SLAKE-EN, and VQA-Med datasets, covering three model families: InternVL3, LLaVA-NeXT, and Qwen3-VL. Both use the sampling-based approach as the primary method for its statistical rigor; detailed experimental setup and comprehensive results, including per-dataset statistics, are reported in Appendix~\ref{app:overconfidence_scaling_extended}.

Across all experimental conditions, every point falls below the diagonal (grey region), empirically confirming that \emph{overconfidence is a systematic property of general-purpose VLMs in medical contexts}, rather than an artifact of any particular architecture and scale. Furthermore, in line with \citet{chhikara2025mind}, we find the overconfidence gap is larger for open-ended questions than for closed-ended ones, as structured answer choices (e.g., yes/no) provide external constraints on the output space that are absent in free-form generation.

\subsection{Larger Models Are Not Necessarily Better Calibrated}
\label{sec:scale}

Table~\ref{tab:scaling_main} summarizes accuracy, mean confidence, overconfidence gap, ECE, and ACE on the medical VQA benchmarks. Per-dataset results are in Tables~\ref{tab:scaling_analysis_rad_vqa}, \ref{tab:scaling_analysis_slake}, \ref{tab:scaling_analysis_vqa_med_2019}, and \ref{tab:scaling_analysis} in the Appendix.
While larger models tend to achieve higher accuracy, this does not translate into better calibration. The overconfidence gap remains uniformly high regardless of model size.
More precisely, no meaningful improvement in ACE was observed from scaling: For within-family transitions from 2B to 7/8B models, 4 of 6 comparisons (2 families × 3 question-type splits) showed slight improvement (mean $\Delta$ACE $= -0.002$), while from 7/8B to 30B+ models, 5 of 9 comparisons (3 families × 3 splits) showed worse calibration (mean $\Delta$ACE $= +0.022$). Overall, 8 of 15 comparisons improved and 7 worsened (mean $\Delta$ACE $= +0.012$).
These results indicate that \emph{scaling alone does not address the fundamental overconfidence problem in VLMs} consistent with prior work on LLM uncertainty~\citep{huang-etal-2024-calibrating,chhikara2025mind,testoni-calixto-2026-mind}, motivating alternative mitigation strategies.

\subsection{No Prompting Strategy Consistently Improves Calibration}
\label{sec:prompting}

\begin{figure}[h!]
\centering
\begin{subfigure}[b]{0.95\linewidth}
    \includegraphics[width=\linewidth]{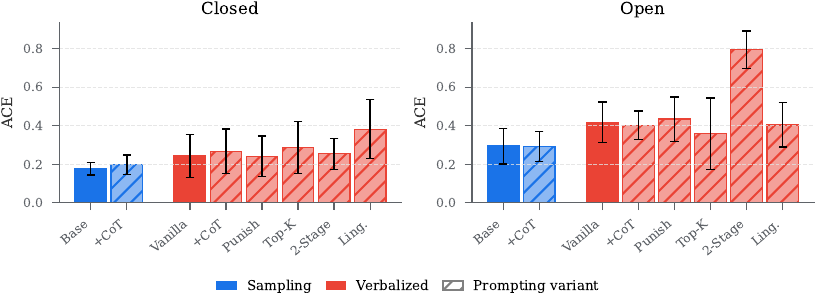}
\end{subfigure}
\caption{ACE across sampling-based and verbalized confidence extraction methods and their prompting variants (\small{Respective prompts are in Appendix~\ref{app:prompts}.}), evaluated on the pooled medical VQA benchmarks and averaged across the 7/8B models. Except for Top-K variants in the verbalized approach, prompting variants do not consistently improve calibration. \small{Full results, including per-model and per-dataset breakdowns with ECE and AUROC, are in Appendix~\ref{app:overconfidence_promting_extended}.}}
\label{fig:comp_prompting}
\end{figure}

Figure~\ref{fig:comp_prompting} reports ACE across different confidence extraction methods, averaged across 7/8B models and datasets. 
The central finding is that \emph{no prompting strategy consistently improves calibration}. CoT offers little to no benefit, and most verbalized variants fail to improve over vanilla prompting, often worsening calibration. Overall, the sampling approach achieves better calibration than the verbalized approach. However, these gains come with increased computational cost, which we analyze in detail in Section~\ref{sec:compute}.

\section{Mitigating Overconfidence via Post-hoc Calibration}
\label{sec:posthoc}

In Section~\ref{sec:overconfidence}, we observed systematic overconfidence across model families, scales, and prompting strategies. This raises the question: \textbf{Can we correct this overconfidence after the model makes its predictions?} Even when the confidence values are poorly calibrated, their \emph{relative ordering} may remain informative. This property allows us to adjust confidence scores post hoc, improving reliability without modifying the underlying model or prompt.

To this end, we investigate the efficacy of standard post-hoc calibration approaches: Platt scaling~\citep{platt1999probabilistic}, isotonic regression~\citep{zadrozny2002transforming}, and histogram binning~\citep{zadrozny2001obtaining}. These methods transform estimated confidence scores into better-calibrated values by fitting a transformation function that minimizes the calibration error on a validation set.
For instance, Platt scaling adjusts confidence scores using a logistic regression model in a strictly monotonic manner (i.e., preserving their relative ordering).
We refer the reader to the respective original works for further details.

\subsection{Empirical Evaluation on Standard Post-hoc Calibration Methods} 

For this analysis, we conduct 5-fold cross-validation, tuning calibration parameters (e.g., scaling factor) on the held-out validation fold and evaluating on each test fold. We report average scores across folds (Figure~\ref{fig:posthoc_bars}), with standard deviations provided in the Appendices (Tables~\ref{tab:hac_ace_all} and \ref{tab:hac_ece_all}), comparing performance before and after applying post-hoc calibration. 

Regardless of the specific post-hoc calibration approach, calibration error consistently decreases. For instance, averaging across 7--8B models, verbalized ECE drops from 0.243 to 0.038 and 0.029 (closed) and from 0.419 to 0.042 and 0.035 (open), under Platt scaling and isotonic regression, respectively; ACE shows a similar pattern, dropping from 0.243 to 0.109 and 0.100 (closed) and from 0.418 to 0.100 and 0.096 (open). 
This improvement surpasses the gains from verbalized prompting strategies explored in the previous section: for example, in Figure~\ref{fig:comp_prompting}, Top-K variant only reduces ECE to 0.356 (open), and ACE to 0.358 (open)---far above the post-hoc calibrated levels. 
In other words, post-hoc calibration yields larger reductions in calibration error than prompt engineering alone.

\begin{figure}[t!]
\centering
\includegraphics[width=\linewidth]{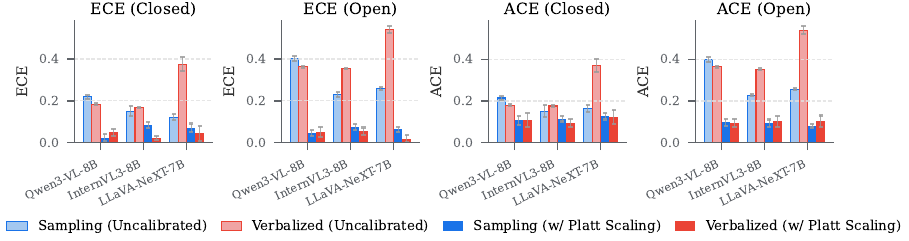}
\caption{Calibration errors (ECE and ACE) before and after post-hoc calibration (Platt scaling) for sampling-based and verbalized confidence on closed- and open-ended questions, evaluated on the pooled datasets. \small{Full results are in Tables~\ref{tab:hac_ace_all} and \ref{tab:hac_ece_all} (Appendix~\ref{app:calibration}).}}
\label{fig:posthoc_bars}
\end{figure}

\section{Vision-Grounded Hallucination-Aware Calibration} \label{sec:hac}

A key limitation shared by standard post-hoc calibration methods is that they apply monotonic transformations to confidence scores. By definition, strict monotonicity preserves the rank ordering of predictions. Consequently, methods like Platt scaling are fundamentally unable to correct the ranking. To overcome this limitation, we incorporate signals that are informative about prediction correctness yet \emph{orthogonal} to confidence. 

We posit that hallucination scores offer such an orthogonal signal. Specifically, hallucinations are a common failure mode in VLMs, where predictions are not grounded in the input image but instead arise from spurious correlations or prior biases. Such hallucinations are known to be driven by strong language priors~\citep{leng2024mitigating, wang2024mitigating}. Combined with the observations in Section~\ref{sec:overconfidence} on VLM overconfidence, predictions with high confidence yet large hallucination scores are likely to have lower accuracy, and should therefore be suppressed relative to non-hallucinated overconfident predictions. Despite this connection, hallucination signals have not been widely explored for confidence calibration.

To this end, we introduce \textbf{Hallucination-Aware Calibration (HAC)}, a simple approach that incorporates vision-grounded hallucination detection signals into the calibration pipeline. 
Given a precomputed hallucination score (described below),  HAC proceeds as follows.

\paragraph{Precomputing Hallucination Score.} First, we compute a vision-grounded hallucination score. We adopt the VASE~\citep{liao2025vision} approach based on its state-of-the-art performance in medical VQA. Specifically, VASE samples $10$ generations from the original image and a weakly perturbed image (preserving clinical validity), respectively. It then calculates the entropy of the contrastive semantic predictive distribution to assess hallucination---a higher VASE score indicates a greater likelihood of hallucination.

\paragraph{Hallucination-informed scoring.}
We combine the raw confidence score $c$ with a hallucination score $h$  via a two-dimensional Platt model:
\begin{equation}
  s_{\text{platt}}(c, h) = \sigma(a \cdot c + b \cdot h + d),
\end{equation}
where $\sigma(\cdot)$ denotes the sigmoid function, and parameters $a, b, d$ are fit by minimizing the negative log-likelihood on a held-out validation dataset. We denote this variant \textbf{HAC-Platt}.

We further propose an alternative formulation, \textbf{HAC-Gate}:
\begin{equation}
  s_{\text{gate}}(c, h) = c \cdot \sigma(-\alpha \cdot h + \beta).
\end{equation}
The sigmoid gate attenuates confidence when $h$ is large.

\subsection{Empirical Evaluation on HAC Variants} \label{sec:hac_baseline}

To provide the empirical support for HAC, we compare HAC against calibration baselines along two complementary dimensions: discriminative capability and calibration error, ensuring improvements in both ranking quality and the fidelity of confidence estimates.

A natural question is how much labeled calibration data HAC requires. We find HAC to be highly data-efficient: as few as $10$--$25$ calibration examples already recover most of the benefit, and both discrimination (AUROC) and calibration error (ACE) reach near-asymptotic values beyond roughly $100$--$200$ examples, so only modest labeled data is needed (Appendix~\ref{app:calsize}, Figures~\ref{fig:calsize_auroc} and~\ref{fig:calsize_ace}).

\begin{table}[t]
\centering
% \caption{AUROC ($\uparrow$) comparison across calibration methods on the pooled dataset. Each cell shows sampling\,/\,verbalized confidence. \small{Full results are in Table~\ref{tab:hac_auroc_all} (Appendix~\ref{app:calibration}).}} % \textbf{Bold} = best per row.
\caption{AUROC ($\uparrow$) comparison across post-hoc calibration methods on the pooled datasets. Platt scaling preserves the same AUROC as uncalibrated scores due to its strictly monotonic transformation. Isotonic regression also preserves nearly identical AUROC due to its non-decreasing mapping. In contrast, HAC improves AUROC by incorporating hallucination signals that rerank predictions. On average across all 8 models, HAC improves AUROC by 5.3 pp over uncalibrated baselines, with notably larger gains on open-ended questions (+7.3 pp), especially for verbalized confidence (+10.1 pp). \small{Full results are in Table~\ref{tab:hac_auroc_all} (Appendix~\ref{app:auroc}).}}
\label{tab:hac_auroc}
\resizebox{\textwidth}{!}{%
\begin{tabular}{@{}ll|cc|cc|cc|cc|cc@{}}
\toprule
 &  & \multicolumn{2}{c}{Uncalibrated} & \multicolumn{2}{c}{Platt Scaling} & \multicolumn{2}{c}{Isotonic Regr.} & \multicolumn{2}{c}{\textbf{HAC-Platt}} & \multicolumn{2}{c}{\textbf{HAC-Gate}} \\
\cmidrule(lr){3-4} \cmidrule(lr){5-6} \cmidrule(lr){7-8} \cmidrule(lr){9-10} \cmidrule(lr){11-12}
 &  & Samp. & Verb. & Samp. & Verb. & Samp. & Verb. & Samp. & Verb. & Samp. & Verb. \\
\midrule
\multirow{3}{*}{\rotatebox[origin=c]{90}{Closed}} & Qwen3-VL-8B & \cellcolor[rgb]{0.89,0.96,0.87} 0.551 & \cellcolor[rgb]{0.84,0.94,0.85} 0.600 & \cellcolor[rgb]{0.89,0.96,0.87} 0.551 & \cellcolor[rgb]{0.84,0.94,0.85} 0.600 & \cellcolor[rgb]{0.88,0.95,0.87} 0.556 & \cellcolor[rgb]{0.83,0.93,0.85} 0.604 & \cellcolor[rgb]{0.80,0.92,0.85} \textbf{0.639} & \cellcolor[rgb]{0.83,0.93,0.85} \textbf{0.608} & \cellcolor[rgb]{0.80,0.92,0.85} \textbf{0.639} & \cellcolor[rgb]{0.83,0.93,0.85} \textbf{0.608} \\
 & InternVL3-8B & \cellcolor[rgb]{0.76,0.91,0.86} 0.663 & \cellcolor[rgb]{0.91,0.96,0.89} 0.522 & \cellcolor[rgb]{0.76,0.91,0.86} 0.663 & \cellcolor[rgb]{0.91,0.96,0.89} 0.522 & \cellcolor[rgb]{0.77,0.91,0.86} 0.660 & \cellcolor[rgb]{0.91,0.96,0.89} 0.522 & \cellcolor[rgb]{0.74,0.90,0.86} 0.677 & \cellcolor[rgb]{0.89,0.96,0.88} \textbf{0.545} & \cellcolor[rgb]{0.74,0.90,0.86} \textbf{0.678} & \cellcolor[rgb]{0.89,0.96,0.88} \textbf{0.545} \\
 & LLaVA-NeXT-7B & \cellcolor[rgb]{0.84,0.94,0.85} \textbf{0.598} & \cellcolor[rgb]{0.88,0.95,0.87} 0.558 & \cellcolor[rgb]{0.84,0.94,0.85} \textbf{0.598} & \cellcolor[rgb]{0.88,0.95,0.87} 0.558 & \cellcolor[rgb]{0.85,0.94,0.86} 0.587 & \cellcolor[rgb]{0.88,0.95,0.87} \textbf{0.562} & \cellcolor[rgb]{0.84,0.94,0.85} 0.597 & \cellcolor[rgb]{0.89,0.96,0.88} 0.547 & \cellcolor[rgb]{0.84,0.94,0.85} \textbf{0.598} & \cellcolor[rgb]{0.88,0.95,0.87} 0.556 \\
\midrule
\multirow{3}{*}{\rotatebox[origin=c]{90}{Open}} & Qwen3-VL-8B & \cellcolor[rgb]{0.79,0.92,0.85} 0.643 & \cellcolor[rgb]{0.70,0.88,0.88} 0.710 & \cellcolor[rgb]{0.79,0.92,0.85} 0.643 & \cellcolor[rgb]{0.70,0.88,0.88} 0.710 & \cellcolor[rgb]{0.79,0.92,0.85} 0.642 & \cellcolor[rgb]{0.70,0.88,0.88} 0.710 & \cellcolor[rgb]{0.65,0.85,0.89} \textbf{0.750} & \cellcolor[rgb]{0.62,0.84,0.90} \textbf{0.771} & \cellcolor[rgb]{0.65,0.85,0.89} \textbf{0.750} & \cellcolor[rgb]{0.66,0.86,0.89} 0.743 \\
 & InternVL3-8B & \cellcolor[rgb]{0.63,0.84,0.90} 0.764 & \cellcolor[rgb]{0.82,0.93,0.85} 0.614 & \cellcolor[rgb]{0.63,0.84,0.90} 0.764 & \cellcolor[rgb]{0.82,0.93,0.85} 0.614 & \cellcolor[rgb]{0.63,0.85,0.90} 0.763 & \cellcolor[rgb]{0.82,0.93,0.84} 0.620 & \cellcolor[rgb]{0.60,0.82,0.90} 0.789 & \cellcolor[rgb]{0.66,0.86,0.89} \textbf{0.742} & \cellcolor[rgb]{0.60,0.82,0.89} \textbf{0.793} & \cellcolor[rgb]{0.67,0.86,0.89} 0.736 \\
 & LLaVA-NeXT-7B & \cellcolor[rgb]{0.71,0.89,0.88} 0.706 & \cellcolor[rgb]{0.84,0.94,0.85} 0.603 & \cellcolor[rgb]{0.71,0.89,0.88} 0.706 & \cellcolor[rgb]{0.84,0.94,0.85} 0.603 & \cellcolor[rgb]{0.71,0.89,0.87} 0.703 & \cellcolor[rgb]{0.84,0.94,0.85} 0.599 & \cellcolor[rgb]{0.68,0.87,0.88} \textbf{0.725} & \cellcolor[rgb]{0.71,0.89,0.87} \textbf{0.703} & \cellcolor[rgb]{0.69,0.87,0.88} 0.721 & \cellcolor[rgb]{0.71,0.89,0.87} 0.702 \\
\bottomrule
\end{tabular}}
\end{table}

\begin{table}[t]
\centering
\caption{AUROC ($\uparrow$) comparison of HAC against uncertainty baselines on the pooled datasets. Baseline uncertainty measures discriminate weakly on closed-ended questions, never exceeding AUROC~$0.661$. In contrast, HAC attains the best overall AUROC, with the largest gains on open-ended questions (up to $+10.2$~pp over the strongest baseline for Qwen3-VL-8B).}

\label{tab:hac_auroc_baselines}
\resizebox{\textwidth}{!}{%
\begin{tabular}{@{}ll|cccccc|cc@{}}
\toprule
 &  & \multirow{2}{*}{\shortstack{Semantic\\Entropy}} & \multirow{2}{*}{\shortstack{Predictive\\Entropy}} & \multirow{2}{*}{\shortstack{Majority\\Vote}} & \multirow{2}{*}{\shortstack{Distinct\\Answers}} & \multirow{2}{*}{\shortstack{Eigen\\Score}} & \multirow{2}{*}{UMPIRE} & \multicolumn{2}{c}{\textbf{HAC-Platt}} \\
\cmidrule(lr){9-10}
 &  &  &  &  &  &  &  & Samp. & Verb. \\
\midrule
\multirow{3}{*}{\rotatebox[origin=c]{90}{Closed}} & Qwen3-VL-8B & \cellcolor[rgb]{0.91,0.96,0.89} 0.530 & \cellcolor[rgb]{0.89,0.96,0.88} 0.545 & \cellcolor[rgb]{0.89,0.96,0.88} 0.545 & \cellcolor[rgb]{0.89,0.96,0.88} 0.543 & \cellcolor[rgb]{0.86,0.95,0.86} 0.577 & \cellcolor[rgb]{0.85,0.94,0.86} 0.593 & \cellcolor[rgb]{0.80,0.92,0.85} \textbf{0.639} & \cellcolor[rgb]{0.83,0.93,0.85} 0.606 \\
 & InternVL3-8B & \cellcolor[rgb]{0.84,0.94,0.85} 0.601 & \cellcolor[rgb]{0.77,0.91,0.86} 0.661 & \cellcolor[rgb]{0.77,0.91,0.86} 0.661 & \cellcolor[rgb]{0.79,0.92,0.85} 0.644 & \cellcolor[rgb]{0.91,0.97,0.90} 0.509 & \cellcolor[rgb]{0.83,0.93,0.85} 0.606 & \cellcolor[rgb]{0.74,0.90,0.86} \textbf{0.677} & \cellcolor[rgb]{0.90,0.96,0.88} 0.542 \\
 & LLaVA-NeXT-7B & \cellcolor[rgb]{0.83,0.93,0.85} \textbf{0.606} & \cellcolor[rgb]{0.84,0.94,0.85} 0.598 & \cellcolor[rgb]{0.84,0.94,0.85} 0.598 & \cellcolor[rgb]{0.91,0.96,0.89} 0.531 & \cellcolor[rgb]{0.89,0.96,0.87} 0.549 & \cellcolor[rgb]{0.88,0.95,0.87} 0.555 & \cellcolor[rgb]{0.84,0.94,0.85} 0.597 & \cellcolor[rgb]{0.90,0.96,0.88} 0.541 \\
\midrule
\multirow{3}{*}{\rotatebox[origin=c]{90}{Open}} & Qwen3-VL-8B & \cellcolor[rgb]{0.76,0.91,0.86} 0.666 & \cellcolor[rgb]{0.77,0.91,0.86} 0.661 & \cellcolor[rgb]{0.77,0.91,0.85} 0.654 & \cellcolor[rgb]{0.77,0.91,0.86} 0.662 & \cellcolor[rgb]{0.76,0.91,0.86} 0.668 & \cellcolor[rgb]{0.92,0.97,0.90} 0.505 & \cellcolor[rgb]{0.65,0.85,0.89} 0.750 & \cellcolor[rgb]{0.62,0.84,0.90} \textbf{0.770} \\
 & InternVL3-8B & \cellcolor[rgb]{0.65,0.85,0.89} 0.751 & \cellcolor[rgb]{0.62,0.84,0.90} 0.770 & \cellcolor[rgb]{0.65,0.85,0.89} 0.750 & \cellcolor[rgb]{0.62,0.84,0.90} 0.770 & \cellcolor[rgb]{0.92,0.97,0.91} 0.413 & \cellcolor[rgb]{0.73,0.90,0.87} 0.684 & \cellcolor[rgb]{0.60,0.82,0.90} \textbf{0.789} & \cellcolor[rgb]{0.65,0.86,0.89} 0.747 \\
 & LLaVA-NeXT-7B & \cellcolor[rgb]{0.68,0.87,0.88} 0.723 & \cellcolor[rgb]{0.69,0.87,0.88} 0.717 & \cellcolor[rgb]{0.73,0.90,0.87} 0.689 & \cellcolor[rgb]{0.70,0.88,0.88} 0.715 & \cellcolor[rgb]{0.73,0.90,0.87} 0.689 & \cellcolor[rgb]{0.73,0.90,0.87} 0.689 & \cellcolor[rgb]{0.68,0.87,0.88} \textbf{0.725} & \cellcolor[rgb]{0.71,0.89,0.87} 0.703 \\
\bottomrule
\end{tabular}}
\end{table}

\paragraph{When Does HAC Improve Predictive Discrimination?}

Table~\ref{tab:hac_auroc} shows the AUROC change after post-hoc calibration for both sampling-based and verbalized confidence. Platt scaling preserves AUROC by definition, and isotonic regression exhibited negligible change. In contrast, \textbf{HAC consistently improves AUROC on open-ended questions} for both confidence types. On closed-ended questions, HAC yields inconsistent results.

Within open-ended questions, the AUROC gain from HAC is most pronounced for verbalized confidence. This is expected: HAC acts as a complementary signal, and its marginal contribution is largest when the uncalibrated confidence alone is a weak discriminator. Sampling-based confidence already achieves relatively high AUROC by aggregating multiple responses---a process inherently correlated with language-grounded hallucination detection signals such as semantic entropy~\citep{farquhar2024detecting}---leaving less room for VASE to add discriminative value. Verbalized confidence, by contrast, tends to start from a lower baseline AUROC, making it a more receptive target for hallucination-aware correction.

The inconsistency on closed-ended questions reflects the informativeness of VASE, rather than a limitation of the HAC framework. VASE correlates strongly with correctness on open-ended questions (Pearson $r=-0.48$ for sampling, $-0.46$ for verbalized) but weakly on closed-ended ones ($r = -0.20$ and $-0.16$, respectively). This is consistent with \citet{liao2025vision}, which validates VASE primarily on open-ended questions. Because HAC is signal-agnostic, VASE can be replaced with a format-appropriate signal. The internal logit margin (HAC-Logit) consistently improves closed-ended AUROC over HAC-VASE (Table~\ref{tab:logit-margin}; +0.150 for Qwen3-VL-8B, +0.063 for InternVL3-8B).

We also compare HAC against training-free uncertainty baselines spanning semantic agreement~\citep{kuhn2023semantic}, lexical agreement~\citep{wang2022self, lin2023generating}, and embedding geometry~\citep{chen2024inside, lau2026uncertainty}. None matches HAC across question types, and unlike HAC these scores are unbounded and give no calibrated probabilities (Table~\ref{tab:hac_auroc_baselines}).

\paragraph{Calibration Gains.} Table~\ref{tab:hac_ace} reports ACE under the same setup. As expected, uncalibrated baselines were systematically overconfident. Standard post-hoc methods and HAC reduce these errors substantially. Overall, HAC achieves calibration gains comparable to standard post-hoc methods, including on closed-ended questions, where its AUROC gains are limited. Therefore, we conclude: \textbf{HAC does not sacrifice calibration quality to achieve its discriminative (AUROC) gains.}

\paragraph{Ablation and Further Analysis.} 
We present additional results on the ablation and further analysis of HAC in Appendix~\ref{app:hac_ablation}. The findings are as follows:

First, we evaluate whether HAC-Platt calibration parameters learned on one dataset transfer to another. We tuned HAC parameters on each dataset and tested on the remaining ones; overall, HAC was relatively insensitive to the choice of calibration dataset (Figures~\ref{fig:transfer_ace}--\ref{fig:transfer_auroc}). 

Second, we inspect the learned HAC-Platt parameters (Table~\ref{tab:hac_params}). The signs $\hat{a} \geq 0$ and $\hat{b} \leq 0$ held across all 160 fits (8 models $\times$ 5 folds $\times$ 2 confidence types $\times$ 2 question types), confirming that higher base confidence consistently increases and higher hallucination scores consistently decrease the calibrated output. The hallucination penalty $\hat{b}$ was on average stronger for open-ended questions than for closed-ended ones, consistent with the greater risk of hallucination in free-form answers. 

Third, we replace VASE in HAC-Platt with Semantic Entropy (SE)~\citep{farquhar2024detecting} or RadFlag~\citep{zhang2024radflag} (Figure~\ref{fig:ablation_hedge}). Both give comparable or slightly better ACE on closed-ended questions but weaker AUROC, confirming that HAC is signal-agnostic and VASE is the strongest single signal in our setting.

Fourth, we vary the number of labeled examples used to fit the calibration parameters (Figures~\ref{fig:calsize_auroc}--~\ref{fig:calsize_ace}). HAC is effective with small calibration sets, often improving AUROC with just 10--50 labels and stabilizing with around 100--200 examples.

\begin{table}[t]
\centering
\caption{ACE ($\downarrow$) comparison across calibration methods on the pooled dataset. Each cell shows sampling or verbalized confidence. Overall, both HAC and standard calibration approaches achieve notable calibration gains comparable to uncalibrated baselines.
\small{Full results are in Table~\ref{tab:hac_ace_all} (Appendix~\ref{app:calibration}).}} % \textbf{Bold} = best per row. 
\label{tab:hac_ace}
\resizebox{\textwidth}{!}{%
\begin{tabular}{@{}ll|cc|cc|cc|cc|cc@{}}
\toprule
 &  & \multicolumn{2}{c}{Uncalibrated} & \multicolumn{2}{c}{Platt Scaling} & \multicolumn{2}{c}{Isotonic Regr.} & \multicolumn{2}{c}{\textbf{HAC-Platt}} & \multicolumn{2}{c}{\textbf{HAC-Gate}} \\
\cmidrule(lr){3-4} \cmidrule(lr){5-6} \cmidrule(lr){7-8} \cmidrule(lr){9-10} \cmidrule(lr){11-12}
 &  & Samp. & Verb. & Samp. & Verb. & Samp. & Verb. & Samp. & Verb. & Samp. & Verb. \\
\midrule
\multirow{3}{*}{\rotatebox[origin=c]{90}{Closed}} & Qwen3-VL-8B & \cellcolor[rgb]{0.94,0.97,0.78} 0.214 & \cellcolor[rgb]{0.87,0.95,0.72} 0.180 & \cellcolor[rgb]{0.69,0.87,0.67} 0.108 & \cellcolor[rgb]{0.70,0.87,0.67} 0.109 & \cellcolor[rgb]{0.66,0.85,0.66} \textbf{0.096} & \cellcolor[rgb]{0.65,0.85,0.66} \textbf{0.095} & \cellcolor[rgb]{0.67,0.86,0.66} 0.101 & \cellcolor[rgb]{0.69,0.87,0.67} 0.108 & \cellcolor[rgb]{0.67,0.86,0.66} 0.101 & \cellcolor[rgb]{0.71,0.87,0.67} 0.115 \\
 & InternVL3-8B & \cellcolor[rgb]{0.81,0.92,0.68} 0.152 & \cellcolor[rgb]{0.87,0.94,0.72} 0.176 & \cellcolor[rgb]{0.71,0.88,0.67} 0.116 & \cellcolor[rgb]{0.65,0.85,0.66} 0.094 & \cellcolor[rgb]{0.68,0.86,0.66} \textbf{0.102} & \cellcolor[rgb]{0.64,0.85,0.66} \textbf{0.093} & \cellcolor[rgb]{0.71,0.87,0.67} 0.114 & \cellcolor[rgb]{0.66,0.85,0.66} 0.096 & \cellcolor[rgb]{0.70,0.87,0.67} 0.113 & \cellcolor[rgb]{0.65,0.85,0.66} 0.095 \\
 & LLaVA-NeXT-7B & \cellcolor[rgb]{0.84,0.93,0.70} 0.165 & \cellcolor[rgb]{0.99,0.77,0.63} 0.371 & \cellcolor[rgb]{0.74,0.89,0.67} 0.127 & \cellcolor[rgb]{0.73,0.88,0.67} 0.123 & \cellcolor[rgb]{0.76,0.90,0.67} 0.131 & \cellcolor[rgb]{0.70,0.87,0.67} \textbf{0.111} & \cellcolor[rgb]{0.73,0.89,0.67} \textbf{0.124} & \cellcolor[rgb]{0.78,0.91,0.68} 0.140 & \cellcolor[rgb]{0.76,0.90,0.67} 0.131 & \cellcolor[rgb]{0.84,0.93,0.70} 0.163 \\
\midrule
\multirow{3}{*}{\rotatebox[origin=c]{90}{Open}} & Qwen3-VL-8B & \cellcolor[rgb]{0.98,0.69,0.59} 0.400 & \cellcolor[rgb]{0.99,0.79,0.64} 0.363 & \cellcolor[rgb]{0.66,0.85,0.66} 0.098 & \cellcolor[rgb]{0.65,0.85,0.66} 0.095 & \cellcolor[rgb]{0.65,0.85,0.66} 0.094 & \cellcolor[rgb]{0.63,0.84,0.65} 0.087 & \cellcolor[rgb]{0.63,0.84,0.65} \textbf{0.086} & \cellcolor[rgb]{0.59,0.82,0.64} \textbf{0.075} & \cellcolor[rgb]{0.63,0.84,0.65} 0.086 & \cellcolor[rgb]{0.59,0.82,0.64} 0.075 \\
 & InternVL3-8B & \cellcolor[rgb]{0.96,0.98,0.81} 0.228 & \cellcolor[rgb]{0.99,0.82,0.66} 0.352 & \cellcolor[rgb]{0.65,0.85,0.66} 0.094 & \cellcolor[rgb]{0.68,0.86,0.66} 0.102 & \cellcolor[rgb]{0.64,0.84,0.66} \textbf{0.090} & \cellcolor[rgb]{0.65,0.85,0.66} 0.094 & \cellcolor[rgb]{0.65,0.85,0.66} 0.094 & \cellcolor[rgb]{0.63,0.84,0.65} \textbf{0.087} & \cellcolor[rgb]{0.66,0.85,0.66} 0.097 & \cellcolor[rgb]{0.64,0.84,0.66} 0.091 \\
 & LLaVA-NeXT-7B & \cellcolor[rgb]{1.00,0.99,0.85} 0.256 & \cellcolor[rgb]{0.81,0.45,0.53} 0.539 & \cellcolor[rgb]{0.61,0.83,0.65} 0.082 & \cellcolor[rgb]{0.68,0.86,0.66} 0.104 & \cellcolor[rgb]{0.56,0.80,0.64} \textbf{0.068} & \cellcolor[rgb]{0.69,0.86,0.67} 0.106 & \cellcolor[rgb]{0.65,0.85,0.66} 0.096 & \cellcolor[rgb]{0.61,0.83,0.65} \textbf{0.083} & \cellcolor[rgb]{0.59,0.82,0.64} 0.077 & \cellcolor[rgb]{0.62,0.83,0.65} 0.084 \\
\bottomrule
\end{tabular}}
\end{table}

\section{Conclusion and Discussion}

We presented a comprehensive empirical study of calibration in medical VQA across three VLM families, spanning scales
from 2B to 38B, and multiple confidence estimation paradigms. Overconfidence is pervasive and persists regardless of scale
or prompting strategy (Section~\ref{sec:pervasive_overconfidence}). Post-hoc calibration consistently reduces calibration error (Section~\ref{sec:posthoc}), but cannot improve discriminative quality due to its monotonicity. To address this, we introduced HAC, which incorporates hallucination detection signals to
rerank predictions, improving AUROC by 5.3 percentage points on average (7.3pp on open-ended questions), without sacrificing calibration performance (Section~\ref{sec:hac}).

\subsection{Discussion of Computational Cost Trade-Offs} \label{sec:compute}

While performance is the primary concern, computational efficiency cannot be overlooked in medical deployment, where time-critical decision support and resource-constrained settings impose strict latency and cost constraints.

For LLMs, inference cost is dominated by autoregressive decoding: input tokens are processed in parallel, whereas output tokens are generated sequentially, incurring higher per-token latency~\citep{pope2023efficiently}. We therefore report the number of generated tokens as a proxy for computational cost during inference. 

\begin{table}[h]
\centering
\caption{Computational cost per question. $N$: the number of required generations per question. \#Tok/Gen reports the mean{\scriptsize$\pm$}std of output tokens per generation.}
\label{tab:token_cost}
\small
\setlength{\tabcolsep}{4pt}
\begin{tabular}{@{}ll c rr rr@{}}
\toprule
& & & \multicolumn{2}{c}{\textbf{Closed}} & \multicolumn{2}{c}{\textbf{Open}} \\
\cmidrule(lr){4-5} \cmidrule(lr){6-7}
& Method & $N$ & \multicolumn{1}{c}{\#Tok/Gen} & \multicolumn{1}{c}{Total} & \multicolumn{1}{c}{\#Tok/Gen} & \multicolumn{1}{c}{Total} \\
\midrule
\multirow{2}{*}{\rotatebox[origin=c]{90}{\tiny Samp.}}
& Base       & 10--20 & 2.9{\tiny$\pm$3.7}     & 29.0  & 7.2{\tiny$\pm$52.1}    & 72.0 \\
& CoT        & 10--20 & 146.2{\tiny$\pm$367.4}  & 1462.0 & 144.5{\tiny$\pm$408.0} & 1445.0 \\
\midrule
\rotatebox[origin=c]{90}{\tiny Hal.}
& VASE       & 20  & 2.4{\tiny$\pm$1.5}      & 48.0  & 8.1{\tiny$\pm$6.2}     & 162.0 \\
\midrule
\multirow{6}{*}{\rotatebox[origin=c]{90}{\tiny Verbalized}}
& Vanilla      & 1 & 25.8{\tiny$\pm$24.1}    & 25.8  & 39.6{\tiny$\pm$29.6}   & 39.6 \\
& Vanilla+CoT  & 1 & 138.1{\tiny$\pm$51.7}   & 138.1 & 141.0{\tiny$\pm$52.7}  & 141.0 \\
& Punish       & 1 & 28.7{\tiny$\pm$24.0}    & 28.7  & 43.8{\tiny$\pm$28.8}   & 43.8 \\
& Linguistic   & 1 & 24.2{\tiny$\pm$21.8}    & 24.2  & 28.0{\tiny$\pm$20.9}   & 28.0 \\
& Top-$k$      & 1 & 92.2{\tiny$\pm$58.7}    & 92.2  & 92.0{\tiny$\pm$64.7}   & 92.0 \\
& Two-stage    & 2 & 81.4{\tiny$\pm$42.8}    & 162.8 & 84.0{\tiny$\pm$60.7}   & 168.0 \\
\bottomrule
\end{tabular}
\label{tab:cost}
\end{table}

We investigate average token counts per generation for each method in Tables~\ref{tab:cost}. Primarily, we confirm that sophisticated prompting strategies increase the inference cost as expected. Notably, CoT prompting increases cost by 20 to 50 times, as it requires generating intermediate reasoning tokens. Verbalized methods also produce more tokens per generation than sampling because they output a confidence score alongside their answer.

To assess overall expense, the number of generations $N$ should also be considered. Verbalized methods require a single forward pass (or two for two-stage prompting), whereas sampling-based methods need at least $N=10$ samples to achieve a reliable estimate (Appendix~\ref{sec:sample_size_ablation}). In contrast, the base sampling method with $N=20$ yields a cumulative token count comparable to two verbalized calls, and lower than a single Top-k call.

For post-hoc calibration and HAC, fitting the calibration model itself is negligible in cost, as it operates on scalar scores on a CPU in near real time. At inference time, applying the learned mapping adds virtually no latency, as it is a simple scalar transformation per sample. To improve both calibration and discriminative quality, HAC additionally leverages hallucination scores from VASE, which requires 20 generations per sample (10 each for the original and perturbed image). Hence, exploring sampling-free yet reliable hallucination detection methods is a promising direction for further reducing this overhead.

\subsection{Toward Reliable Deployment: Guidance and Open Challenges}
\label{sec:discussion}

\paragraph{Practical Recommendations.}
Our findings establish standard post-hoc calibration (e.g., Platt scaling) as a minimum baseline for medical VLM deployment as it consistently reduces calibration error at negligible cost. The choice between sampling-based and verbalized approaches should be decided by the goal. For example, sampling-based confidence benefits from multi-sample aggregation, which provides statistically grounded estimates, but incurs compute cost that scales with $N$ (Table~\ref{tab:cost}). With sufficient serving bandwidth, hybrid strategies that aggregate verbalized scores across multiple generations are also feasible and may combine the strengths of both paradigms. When discriminative quality is the priority, HAC offers the largest gains---improving AUROC by 5.3 percentage points on average and up to 7.3pp on open-ended questions---by incorporating hallucination signals to rerank predictions, which are most relevant to real-world clinical use. Although HAC requires additional generations for hallucination estimation, this overhead is mitigable in practice, as the generations are independent and can be fully parallelized without increasing latency.

\paragraph{Limitations.}
While our evaluation focuses on general-purpose VLMs, the calibration of domain-specific fine-tuned models (e.g., Llava-Med) could be further explored. Second, although VQA-Med covers all eight primary imaging modality categories, SLAKE and VQA-RAD include only a subset (X-ray, CT, and MRI), which may limit the generalizability of our findings to underrepresented modalities. Third, no single auxiliary signal serves both answer formats: VASE is more effective for open-ended questions, while the logit margin performs better for closed-ended questions. Additionally, developing a more cost-effective vision-grounded hallucination detection method remains an important direction.

\paragraph{Future Work.}
Furthermore, while verbalized confidence estimation strategies, Top-$k$ and Two-stage methods, already incorporate elements of a sampling-based approach, exploring hybrid confidence estimation strategies that combine verbalized and sampling-based paradigms may be worthwhile when sufficient bandwidth is available. In addition, incorporating other informative signals---such as statistics from reasoning traces~\citep{testoni-calixto-2026-mind}---beyond hallucination is a promising direction. Moreover, developing sampling-free yet reliable hallucination detection methods would further improve robustness. Lastly, although our evaluation relies on benchmark datasets, validating these findings in real clinical settings is essential to bridge the gap between benchmark-level observations and real-world reliability.

\section*{Acknowledgments}
The authors thank Paul Vozila for his valuable feedback on the manuscript.
Ji Young Byun was supported in part by a discretionary fund at John Hopkins University's Whiting School of Engineering.
Young-Jin Park was supported in part by the MIT-IBM Watson AI Lab.

\bibliography{colm2026_conference}
\bibliographystyle{colm2026_conference}

\clearpage
\appendix

\section{Further Related Work: Calibration in Deep Learning and Generative Models.}

Modern deep neural networks are often observed to be miscalibrated: while achieving high accuracy, they tend to produce overconfident predictions. \citet{guo2017calibration} provided the characterization of this phenomenon and introduced \emph{temperature scaling} as a post-hoc remedy that divides logits by a learned scalar before the softmax. Subsequent post-hoc methods include Platt scaling~\citep{platt1999probabilistic}, isotonic regression~\citep{zadrozny2002transforming}, and histogram binning~\citep{zadrozny2001obtaining}, all of which rescale outputs after training without altering the model itself. 

However, measuring confidence in modern generative models remains an open challenge~\citep{lin2023generating, geng2024survey, liu2025uncertainty}. Unlike traditional classifiers, which produce probabilities over a fixed output space, VLMs generate free-form text through autoregressive decoding. This makes the notion of a single prediction probability ambiguous. 

This ambiguity gives rise to two concrete difficulties in practice. First, the same correct answer can surface in many linguistic forms (e.g., "left ventricle," "the left ventricular region," or simply "left"), so token-level probabilities do not cleanly aggregate into answer-level confidence. Moreover, models that generate intermediate reasoning steps before committing to a final answer further complicate estimation, as confidence must be estimated over a sequence of tokens that includes both reasoning and response~\citep{fu2025deep}. Second, many recent studies instead rely on verbalized confidence~\citep{tian2023just, xiong2024llms, ni2024llms, du2025confidence,xuan-etal-2025-seeing}, prompting models to explicitly report a confidence label or numerical value. While convenient, these approaches lack theoretical grounding and may conflate instruction following with genuine uncertainty estimation.

Moreover, models that generate intermediate reasoning steps before committing to a final answer further complicate estimation, as confidence must be estimated over a sequence of tokens that includes both reasoning and response~\citep{fu2025deep}. 
Beyond aggregation, many recent studies instead rely on verbalized confidence~\citep{tian2023just, xiong2024llms, ni2024llms, du2025confidence,xuan-etal-2025-seeing}, prompting models to explicitly report a confidence label or numerical value. While convenient, this approach lacks theoretical grounding and may conflate instruction following with genuine uncertainty estimation.

While a few lines of work suggest a more principled calibration in LLM generative setups~\citep{shen2024thermometer, park2025know}, calibration in the vision-language domain remains underexplored, with general-domain VLMs consistently exhibiting poor alignment between confidence and accuracy~\citep{groot2024overconfidence,pmlr-v235-tu24a}.

\section{Dataset Statistics}
\label{app:dataset_stats}

We provide detailed dataset statistics of the medical VQA datasets used in our empirical analysis in Table~\ref{tab:da}. Note that the SLAKE dataset provides its own CLOSED/OPEN labels, where CLOSED includes not only yes/no questions but also \emph{closed-choice} questions such as ``Which organ is abnormal, heart or lung?'' Similarly, VQA-Med-2019 contains comparable closed-choice questions (e.g., ``Is this image modality T1, T2, or FLAIR?'') without explicitly providing such labels. Meanwhile, VQA-RAD classifies all non-yes/no questions as open-ended. Since the boundary between closed-ended and open-ended questions is either undefined or ambiguously defined across benchmarks, we adopt a stricter criterion and define closed-ended questions as solely yes/no questions throughout our analysis, ensuring consistent categorization across all datasets.

\begin{table}[h]
\centering
\caption{Statistics of the test splits of medical VQA datasets with open- and closed-ended (yes/no) questions.} \label{tab:da}
\begin{tabular}{lcccc}
\toprule
\textbf{Dataset} & \textbf{\# Images} & \textbf{Open-ended} & \textbf{Closed-ended} & \textbf{Total QA} \\
\midrule
VQA-RAD      & 203  & 200 & 251 & 451 \\
SLAKE-EN     & 96   & 706 & 355 & 1061 \\
VQA-Med-2019 & 500  & 436 & 64 & 500 \\
\midrule
Pooled       & 799  & 1342 & 670 & 2012 \\
\bottomrule
\end{tabular}
\end{table}

\section{Confidence Measurements.} \label{app:confidence}

\newcommand{\parse}{\mathrm{parse}}

Let $\mathcal{X}$ and $\mathcal{Y}$ denote the input and output spaces, respectively.
Consider a model parameterized by $\theta$, which defines a conditional distribution: $P_\theta(y \mid x)$.
Given an input $x \in \mathcal{X}$, the corresponding predicted label is $\hat{y} = \arg\max_{y \in \mathcal{Y}} P_\theta(y \mid x).$
The goal is to estimate the confidence of the model prediction: $P_\theta(y = \hat{y} \mid x).$

Unlike the standard supervised learning setup, generative models have an open-ended output space.
As a result, neither training the model to be uncertainty-aware (i.e., to output a distribution $P_\theta$ rather than a point estimate) nor defining $\hat{y}$ is straightforward.
To address this challenge, prior work typically adopts three standardized approaches: sampling-based, logit-based, and verbalized methods.

\subparagraph{Sampling-based confidence.}
The sampling-based approach approximates $P_\theta$ via Monte Carlo sampling \citep{neal1993probabilistic, wang2022self}.
More specifically, for each question $q$ with image $I$, we draw $N$ independent generations $\{a_1, \ldots, a_N\} \sim P(\cdot \mid q, I)$ and define confidence as the empirical frequency of the plurality answer $\hat{a}$:
\begin{equation*}
    c_{\text{samp}}(q) = \frac{1}{N} \sum_{i=1}^{N} \mathbf{1}[\text{parse}(a_i) = \hat{a}].
\end{equation*}
where $\parse(\cdot)$ is a function mapping textual answers to a canonical form, and $\hat{a} = \arg\max_a \sum_{i=1}^N \mathbf{1}[\parse(a_i) = a]$.
While this estimator is \emph{unbiased}, it requires a large number of samples for precise estimation, which can become a bottleneck under an expensive LLM inference regime.

For open-form questions, there may exist multiple linguistic ways to express a semantically equivalent answer. Thus, rather than grouping by pattern matching, we extend $\parse(\cdot)$ to map semantically similar answers into the same canonical form~\citep{kuhn2023semantic, farquhar2024detecting}. While the original paper employs DeBERTa (0.9B) \citep{he2020deberta} to perform semantic clustering, its NLI-based training on general-domain corpora may not generalize well to domain-specific equivalences; we instead use Qwen/Qwen3-4B-Instruct-2507, which demonstrates competitive performance on medical benchmarks~\citep{qwen3technicalreport}.

\subparagraph{Verbalized confidence.}
We evaluate six variants of verbalized confidence prompting strategy, each instructing the model to report a numerical confidence score alongside its answer:
\begin{itemize}
    \item \textit{Vanilla}: The model is asked to provide a confidence score (0--100\%) alongside its answer~\citep{tian2023just, xiong2024llms}.
    \item \textit{Vanilla+CoT}: Same as Vanilla, with the model additionally instructed to reason step-by-step prior to answering \citep{wei2022chain}.
    \item \textit{Two-Stage}: The model first generates an answer; confidence is then elicited in a separate subsequent query~\citep{xiong2024llms}.
    \item \textit{Punish}: The model is explicitly warned that overconfident incorrect answers will be penalized~\citep{ni2024llms}.
    \item \textit{Top-K}: The model is asked to produce its top-3 candidate answers along with associated probability estimates~\citep{tian2023just}. The highest-probability candidate is taken as the predicted answer and its probability as the confidence score.
    \item \textit{Linguistic}: The model expresses confidence using qualitative terms (e.g., \textit{very likely}), which are mapped to numerical bins~\citep{tian2023just}.
\end{itemize}

 \subparagraph{Logit-based confidence.}
Logit-based methods extract confidence directly from the output distribution over answer tokens.
For instance, for closed-form yes/no questions, we aggregate over corresponding tokens via log-sum-exp:
\begin{equation}
    c_{\text{logit}}(q) = \mathrm{softmax}(\hat{z}_{\text{yes}},\, \hat{z}_{\text{no}}),
\end{equation}
where $\hat{z}_{\text{yes}} = \log \sum_{t \in \mathcal{T}_{\text{yes}}} e^{z_t}$ (and similarly for $\hat{z}_{\text{no}}$) and $\mathcal{T}_{\text{yes}}$ is the set of tokens corresponding to ``yes'' (e.g., \texttt{yes}, \texttt{Yes}, \texttt{YES}).
This approach is computationally efficient, as it does not require additional sampling or generation.
However, it is limited to closed-form settings (e.g., yes/no or multiple-choice), as open-ended questions allow diverse valid generations with varying linguistic forms.
Similarly, it does not extend to reasoning-based methods, where outputs are inherently open-ended. Due to these limitations, we do not use logit-based scores as a standalone confidence measure. However, we revisit the yes/no logit margin as an auxiliary signal for closed-ended questions in Section~\ref{sec:hac_baseline} (Table~\ref{tab:logit-margin})

\subsection{Prompts}
\label{app:prompts}

Below we list the exact prompt templates used for each confidence elicitation method. In all templates, \texttt{\{question\}} is replaced with the input question at inference time. Every prompt is prefixed with the corresponding medical image.

\subsubsection{Sampling-Based Prompts}

\paragraph{Base (no CoT).}
Used for sampling-based confidence without chain-of-thought.
\begin{promptbox}
You are a medical AI assistant. Look at the provided medical image and answer the following question.\\[4pt]
Question: \{question\}\\[4pt]
Provide only the answer.\\[2pt]
Format:\\
Answer: [your answer]
\end{promptbox}

\paragraph{Base + CoT.}
Used for sampling-based confidence with chain-of-thought reasoning.
\begin{promptbox}
You are a medical AI assistant. Look at the provided medical image and answer the following question.\\[4pt]
Question: \{question\}\\[4pt]
Think step by step. Then provide your answer.\\[2pt]
Format:\\
Reasoning: [your reasoning]\\
Answer: [your answer]
\end{promptbox}

\subsubsection{Verbalized Confidence Prompts}

\paragraph{Vanilla.}
The model is asked to provide a numeric confidence score (0--100\%) alongside its answer.
\begin{promptbox}
You are a medical AI assistant. Look at the provided medical image and answer the following question.\\[4pt]
Question: \{question\}\\[4pt]
Provide your answer and your confidence level (0-100\%) in your answer.\\[2pt]
Format:\\
Answer: [your answer]\\
Confidence: [0-100]\%
\end{promptbox}

\paragraph{Vanilla + CoT.}
Same as Vanilla, with step-by-step reasoning elicited before the answer.
\begin{promptbox}
You are a medical AI assistant. Look at the provided medical image and answer the following question.\\[4pt]
Question: \{question\}\\[4pt]
Think step by step. Then provide your answer and your confidence level (0-100\%) in your answer.\\[2pt]
Format:\\
Reasoning: [your reasoning]\\
Answer: [your answer]\\
Confidence: [0-100]\%
\end{promptbox}

\paragraph{Punish.}
The model is explicitly warned that overconfident incorrect answers will be penalized.
\begin{promptbox}
You are a medical AI assistant. Look at the provided medical image and answer the following question. You will be punished if the answer is wrong but you answer it with high confidence.\\[4pt]
Question: \{question\}\\[4pt]
Provide your answer and your confidence level (0-100\%) in your answer.\\[2pt]
Format:\\
Answer: [your answer]\\
Confidence: [0-100]\%
\end{promptbox}

\paragraph{Top-K.}
The model provides its top-3 candidate answers with associated probability estimates. The highest-probability candidate is taken as the predicted answer.
\begin{promptbox}
You are a medical AI assistant. Look at the provided medical image and answer the following question.\\[4pt]
Question: \{question\}\\[4pt]
Provide your top 3 best guesses for the answer, along with the probability (0-100\%) that each guess is correct. The probabilities should sum to 100\%.\\[2pt]
Format:\\
Guess 1: [answer] (Probability: [X]\%)\\
Guess 2: [answer] (Probability: [Y]\%)\\
Guess 3: [answer] (Probability: [Z]\%)
\end{promptbox}

\paragraph{Two-Stage.}
A two-step procedure: the model first generates an answer (Stage~1), then confidence is elicited in a separate follow-up query (Stage~2).

\noindent\textit{Stage 1:}
\begin{promptbox}
You are a medical AI assistant. Look at the provided medical image and answer the following question.\\[4pt]
Question: \{question\}\\[4pt]
Provide your answer.
\end{promptbox}

\noindent\textit{Stage 2} (where \texttt{\{answer\}} is the model's response from Stage~1):
\begin{promptbox}
Question: \{question\}\\
Proposed answer: \{answer\}\\[4pt]
How likely is the above answer to be correct? Provide a probability between 0\% and 100\%.\\[2pt]
Format:\\
Confidence: [0-100]\%
\end{promptbox}

\paragraph{Linguistic.}
The model expresses confidence using qualitative terms, which are mapped to numerical probabilities.
\begin{promptbox}
You are a medical AI assistant. Look at the provided medical image and answer the following question.\\[4pt]
Question: \{question\}\\[4pt]
Provide your answer and describe how confident you are using one of these terms: ``almost certain'', ``highly likely'', ``very good chance'', ``probable'', ``likely'', ``better than even'', ``about even'', ``unlikely'', ``improbable'', ``very good chance not'', ``highly unlikely'', ``almost certainly not''.\\[2pt]
Format:\\
Answer: [your answer]\\
Confidence: [one of the terms above]
\end{promptbox}

\noindent The linguistic terms are mapped to numerical probabilities as in Table~\ref{tab:linguistic_map}.

\begin{table}[h]
\centering
\small
\begin{tabular}{lc}
\toprule
\textbf{Term} & \textbf{Probability} \\
\midrule
Almost certain & 0.95 \\
Highly likely & 0.90 \\
Very good chance & 0.85 \\
Probable & 0.75 \\
Likely & 0.70 \\
Better than even & 0.60 \\
About even & 0.50 \\
Unlikely & 0.30 \\
Improbable & 0.20 \\
Very good chance not & 0.15 \\
Highly unlikely & 0.10 \\
Almost certainly not & 0.05 \\
\bottomrule
\end{tabular}
\caption{Mapping from linguistic confidence terms to numerical probabilities.}
\label{tab:linguistic_map}
\end{table}

\subsubsection{LLM-as-a-Judge Prompt} \label{app:llm_judge}

We use an LLM judge to evaluate whether predicted answers are semantically equivalent to the ground truth. We chose Qwen3-4B-Instruct for its computational efficiency and strong medical knowledge at small scale: in Medmarks v0.1~\citep{warner2026medmarks}, Qwen3-235B-A22B ranked as the top open-source model across 20 medical benchmarks, and even Qwen3-4B Thinking variant outperformed the average model in the 7B--19B model. This choice is consistent with prior findings that compact instruction-tuned models perform well on structured discriminative tasks \citep{abdin2024phi3technicalreporthighly, zhang2024tinyllama, qwen3technicalreport, team2025gemma}, that judge reliability depends more on task formulation than on model scale \citep{zheng2023judging, thakur-etal-2025-judging}, and that LLM-based judges correlate better with human evaluation than classical metrics in medical VQA \citep{abacha2025proceedings}.

We assess judge stability on 500 examples: our Qwen3-4B judge agrees with Qwen3-30B-A3B-Instruct-2507 and GPT-5 in 94.8\% and 92.2\% of cases, respectively. Table~\ref{tab:inter-model} details the disagreement patterns.

A human expert further annotated 125 of these examples: all 35 unanimously accepted cases, 35 randomly sampled unanimously rejected cases, and all 55 inter-judge disagreements. The expert matched the unanimous consensus in 69/70 cases (98.6\%). On the disagreement cases, the dominant pattern is GPT-5 uniquely rejecting answers both Qwen judges accept (49.1\%), with the expert siding with the Qwen judges in 23/27 such cases. Qualitatively, many GPT-5 disagreements reflected over-rejection (e.g. rejecting "no damage" or "no disease" as paraphrases of "none").

Against expert labels (Table~\ref{tab:human_validation}), we prioritize precision: a false positive inflates measured accuracy by crediting a non-equivalent answer, whereas a false negative merely makes evaluation more conservative by rejecting acceptable paraphrases. Qwen3-4B-Instruct's high precision (0.906) safeguards against over-crediting, while its lower recall (0.753) remains a limitation (Section~\ref{sec:discussion}).

Thus, for open-ended questions, we adopt llm-as-a-judge framework and the corresponding prompt is as follows:

\begin{promptbox}
Given a medical VQA question, determine if the predicted answer is semantically equivalent to the ground truth.\\[4pt]
Question: \{question\}\\
Ground truth: \{ground\_truth\}\\
Predicted: \{predicted\}\\[4pt]
Are these semantically equivalent? Reply with exactly one word: yes or no.
\end{promptbox}

\begin{table}[h]
\centering
\small
\caption{Inter-model disagreement patterns on the $N{=}55$ cases where the
three judges do not unanimously agree. \ding{51}/\ding{55} denote
accept/reject.}
\begin{tabular}{cccrr}
\toprule
Qwen3-4B & Qwen3-30B & GPT-5 & Count & \% of dis. \\
\midrule
\ding{51} & \ding{51} & \ding{55} & $27$ & $49.1$ \\
\ding{55} & \ding{51} & \ding{55} & $16$ & $29.1$ \\
\ding{55} & \ding{51} & \ding{51} & $8$  & $14.5$ \\
\ding{55} & \ding{55} & \ding{51} & $2$  & $3.6$  \\
\ding{51} & \ding{55} & \ding{55} & $2$  & $3.6$  \\
\midrule
\multicolumn{3}{l}{Total} & $55$ & $100.0$ \\
\bottomrule
\end{tabular}
\label{tab:inter-model}
\end{table}

\begin{table}[h]
\centering
\small
\caption{Human validation of the LLM judges. Judge verdicts are compared
against expert labels on $N{=}125$ annotated samples (all $70$
unanimous-agreement cases plus all $55$ inter-judge disagreements).}
\begin{tabular}{lccccccc}
\toprule
Judge & TP & FP & FN & TN & Precision & Recall & F1 \\
\midrule
Qwen3-4B (ours) & $58$ & $6$  & $19$ & $42$ & $0.906$ & $0.753$ & $0.822$ \\
Qwen3-30B       & $74$ & $12$ & $3$  & $36$ & $0.860$ & $0.961$ & $0.908$ \\
GPT-5           & $43$ & $2$  & $34$ & $46$ & $0.956$ & $0.558$ & $0.705$ \\
\bottomrule
\end{tabular}
\label{tab:human_validation}
\end{table}

\subsubsection{Answer Clustering Prompt} \label{app:cluster}

For sampling-based confidence, we cluster the $N$ sampled answers into semantically equivalent groups before computing plurality frequency. Pairwise equivalence is determined by an LLM judge (Qwen3-4B-Instruct) using the following prompt:

\begin{promptbox}
In the context of a medical VQA question, determine if two answers are semantically equivalent (same medical meaning, even if worded differently).\\[4pt]
Question: \{question\}\\
Answer A: \{answer\_a\}\\
Answer B: \{answer\_b\}\\[4pt]
Are these semantically equivalent? Reply with exactly one word: yes or no.
\end{promptbox}

\subsubsection{VASE Prompt}
We use the same Base (no CoT) prompt for answer generation. we follow the \texttt{hedge-bench} pipeline \citep{gautam2025hedge}, computing SE, RadFlag, and VASE over semantically clustered answers.\footnote{The original VASE implementation~\citep{liao2025vision} is not publicly available; \texttt{hedge-bench} provides the only open-source reimplementation.} For closed-ended questions, answers are clustered by exact yes/no matching; for open-ended questions, we use embedding-based clustering with SentenceTransformer similarity (threshold $\tau=0.85$), following the implementation provided in their GitHub. 

\clearpage
\section{Sample Size Ablation}
\label{sec:sample_size_ablation}

We conduct an ablation study on the sample size ($N$) for confidence estimation. Since sampling-based methods are unbiased estimators, confidence estimates converge as $N$ grows, but computational cost increases linearly. It is therefore important to identify the trade-off between estimation quality and sample budget.

\paragraph{Procedure.} We analytically study how calibration quality degrades at smaller sample budgets without re-running inference. We first run each model with $N{=}100$ for 2B/7B/8B models and $N{=}20$ for 30B+ models for each question $q$. Let $p = c_{\text{samp}}(q)$ denote the observed confidence for that question. For each reduced sample size $N' \in \{1, 3, 5, 10, 15, 20, 50, 100\}$ (capped at the original $N$), we simulate $K' \sim \text{Binomial}(N', p)$ and set $c' = K'/N'$. This is analytically valid because the sampling-based estimator is unbiased. We repeat this procedure $M{=}1{,}000$ times, computing ECE, ACE, and AUROC on the full evaluation set per (dataset, model, question type) group for each trial, and report the mean $\pm$ standard deviation across trials.

\begin{figure}[h!]
\centering
\includegraphics[width=\linewidth]{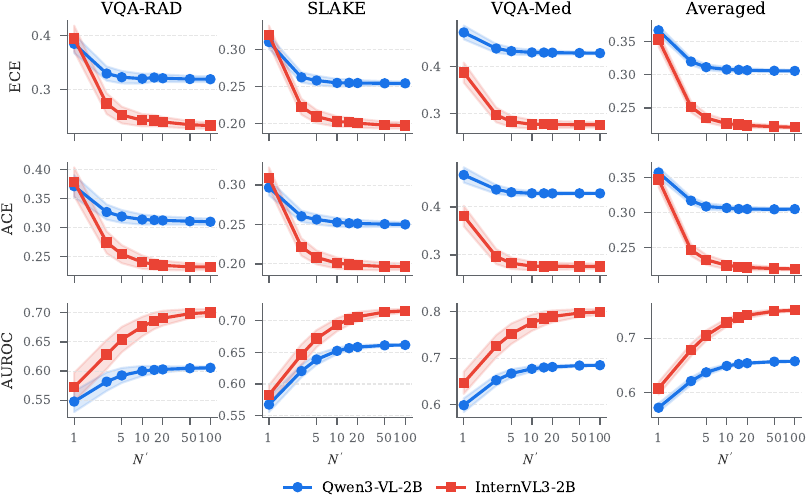}
\caption{Sample size ablation for \textbf{2B models}. ECE ($\downarrow$), ACE ($\downarrow$), and AUROC ($\uparrow$) as a function of $N'$, averaged across question types. Shaded bands show std over 1{,}000 simulations.}
\label{fig:sample_size_2b}
\end{figure}

\begin{figure}[h!]
\centering
\includegraphics[width=\linewidth]{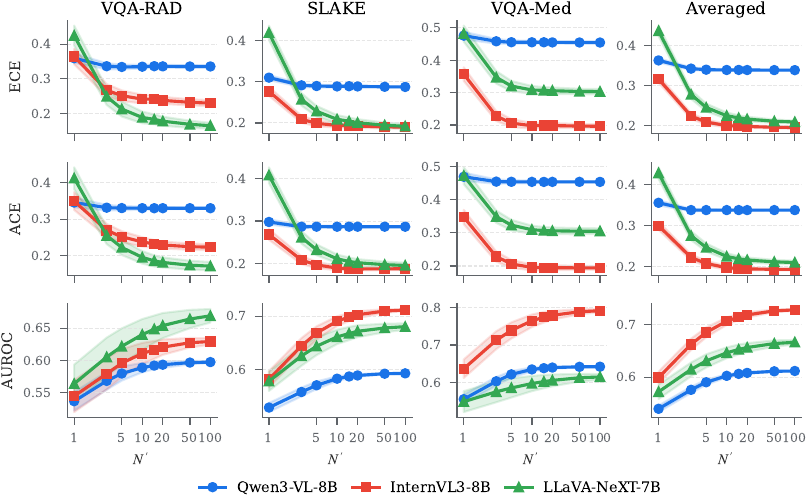}
\caption{Sample size ablation for \textbf{7/8B models}. Same setup as Figure~\ref{fig:sample_size_2b}.}
\label{fig:sample_size_7b}
\end{figure}

\begin{figure}[h!]
\centering
\includegraphics[width=\linewidth]{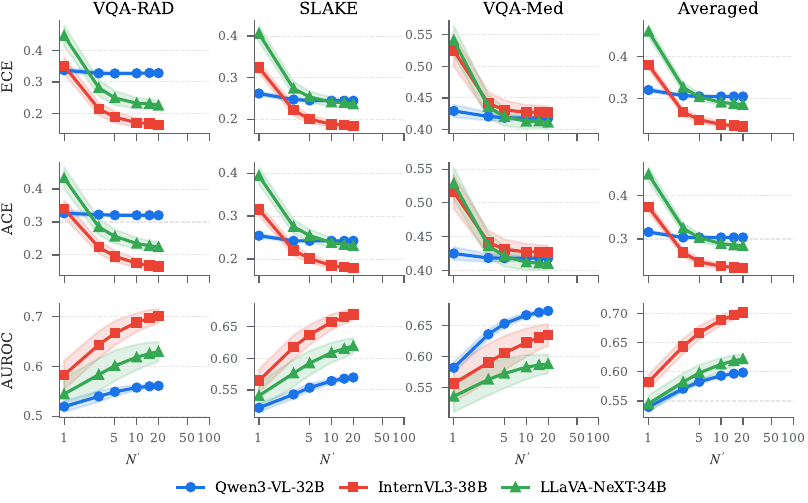}
\caption{Sample size ablation for \textbf{30B+ models}. Same setup as Figure~\ref{fig:sample_size_2b}. Note: $N' \leq 20$ since the original sample size is $N{=}20$.}
\label{fig:sample_size_30b}
\end{figure}

\paragraph{Results.} Figures~\ref{fig:sample_size_2b}--\ref{fig:sample_size_30b} show the results for 2B, 7/8B, and 30B+ model tiers, respectively. All three metrics improved monotonically with $N'$: ECE and ACE decreased as the confidence estimates became less noisy, while AUROC increased as the ranking quality of confidence scores improved. 
The simulation variance (shaded bands) also shrank rapidly with $N'$, becoming negligible by $N'{=}20$. These findings suggest that $N \geq 10$--$20$ samples suffice for stable calibration estimates, and that the additional cost of $N{=}100$ yields marginal improvement.

\clearpage

\clearpage

\section{Full Results on Overconfidence in Medical VQA (Section~\ref{sec:overconfidence})} \label{app:overconfidence_extended}

\subsection{Calibration across Model Families and Varying Scales} \label{app:overconfidence_scaling_extended}

Tables~\ref{tab:scaling_analysis_rad_vqa}--\ref{tab:scaling_analysis_vqa_med_2019} present the full results. We consistently observed systematic overconfidence across model families and datasets; scaling did not mitigate the overconfidence or miscalibration. While accuracy generally increased with larger models, reliability metrics such as ECE, ACE, and AUROC did not follow the same trend.

\subsection{Prompting Strategy Comparison} \label{app:overconfidence_promting_extended}

\begin{figure}[h!]
\centering
\begin{subfigure}[b]{0.96\linewidth}
    \includegraphics[width=\linewidth]{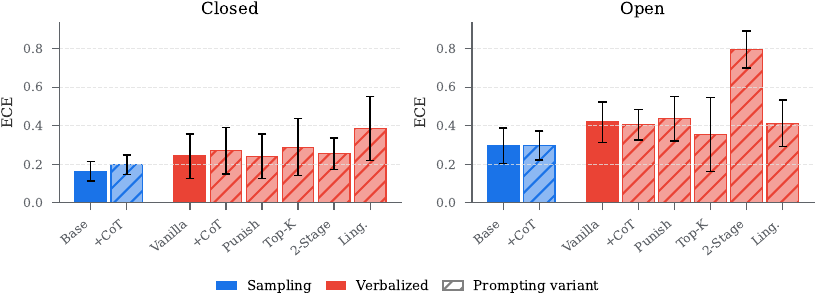}
\end{subfigure}
\caption{ECE across different confidence extraction methods, including sampling and verbalized methods and their prompting variants. The evaluation is on the pooled medical VQA benchmarks and averaged across the 7/8B models.}
\label{fig:comp_acc_ece}
\end{figure}

\begin{figure}[h!]
\centering
\begin{subfigure}[b]{0.96\linewidth}
    \includegraphics[width=\linewidth]{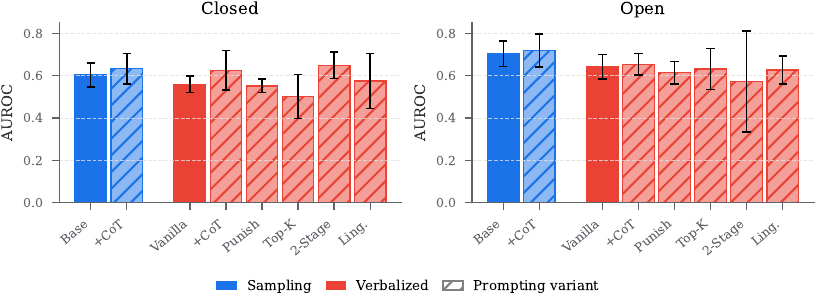}
\end{subfigure}
\caption{AUROC across different confidence extraction methods, including sampling and verbalized methods and their prompting variants. The evaluation is on the pooled medical VQA benchmarks and averaged across the 7/8B models.}
\label{fig:comp_acc_auroc}
\end{figure}

Figures~\ref{fig:comp_acc_ece} and \ref{fig:comp_acc_auroc} extend Figure~\ref{fig:comp_prompting} with ECE and AUROC, respectively.
We observe the same finding: the prompting strategy does not show consistent improvements. 

Table~\ref{tab:prompting_strategies} reports the full results. 
No single prompting variant consistently outperforms the base sampling or verbalized methods across all settings, reconfirming the finding that prompting strategies alone do not reliably improve calibration.

\begin{table}[t]
\centering
\small
\caption{Effect of model scale on accuracy, confidence, and calibration (sampling-based confidence, micro-averaged across VQA-RAD, SLAKE-EN, and VQA-Med-2019). Overconf.\ Gap = Mean Conf.\ $-$ Accuracy. $\uparrow$: higher is better, $\downarrow$: lower is better.}
\label{tab:scaling_analysis}
% \resizebox{\textwidth}{!}{%
\begin{tabular}{@{}llccccccc@{}}
\toprule
\textbf{Type} & \textbf{Model} & \textbf{Acc.} ($\uparrow$) & \textbf{Conf.} & \textbf{Gap} ($\downarrow$) & \textbf{ECE} ($\downarrow$) & \textbf{ACE} ($\downarrow$) & \textbf{AUROC} ($\uparrow$) \\
\midrule
\multirow{8}{*}{\rotatebox[origin=c]{90}{\textbf{Closed}}}
 & Qwen3-VL-2B          & 0.740 & 0.964 & 0.224 & 0.235 & 0.224 & 0.586 \\
 & Qwen3-VL-8B          & 0.766 & 0.980 & 0.214 & 0.219 & 0.214 & 0.551 \\
 & Qwen3-VL-32B         & 0.779 & 0.991 & 0.212 & 0.219 & 0.212 & 0.521 \\
\cmidrule(lr){2-8}
 & InternVL3-2B         & 0.696 & 0.857 & 0.162 & 0.181 & 0.194 & 0.693 \\
 & InternVL3-8B         & 0.782 & 0.910 & 0.128 & 0.151 & 0.152 & 0.663 \\
 & InternVL3-38B        & 0.775 & 0.842 & 0.067 & 0.106 & 0.122 & 0.662 \\
\cmidrule(lr){2-8}
 & LLaVA-NeXT-7B        & 0.624 & 0.730 & 0.106 & 0.122 & 0.165 & 0.598 \\
 & LLaVA-NeXT-34B       & 0.615 & 0.803 & 0.188 & 0.200 & 0.212 & 0.618 \\
\midrule
\multirow{8}{*}{\rotatebox[origin=c]{90}{\textbf{Open}}}
 & Qwen3-VL-2B          & 0.472 & 0.808 & 0.335 & 0.356 & 0.345 & 0.695 \\
 & Qwen3-VL-8B          & 0.525 & 0.925 & 0.400 & 0.401 & 0.400 & 0.643 \\
 & Qwen3-VL-32B         & 0.586 & 0.936 & 0.349 & 0.351 & 0.349 & 0.641 \\
\cmidrule(lr){2-8}
 & InternVL3-2B         & 0.484 & 0.732 & 0.249 & 0.250 & 0.249 & 0.782 \\
 & InternVL3-8B         & 0.537 & 0.761 & 0.225 & 0.230 & 0.228 & 0.764 \\
 & InternVL3-38B        & 0.436 & 0.730 & 0.294 & 0.294 & 0.294 & 0.740 \\
\cmidrule(lr){2-8}
 & LLaVA-NeXT-7B        & 0.339 & 0.595 & 0.256 & 0.259 & 0.256 & 0.706 \\
 & LLaVA-NeXT-34B       & 0.354 & 0.708 & 0.354 & 0.355 & 0.360 & 0.645 \\
\midrule
\multirow{8}{*}{\rotatebox[origin=c]{90}{\textbf{Overall}}}
 & Qwen3-VL-2B          & 0.562 & 0.860 & 0.298 & 0.316 & 0.305 & 0.658 \\
 & Qwen3-VL-8B          & 0.605 & 0.943 & 0.338 & 0.340 & 0.338 & 0.612 \\
 & Qwen3-VL-32B         & 0.651 & 0.954 & 0.304 & 0.307 & 0.303 & 0.601 \\
\cmidrule(lr){2-8}
 & InternVL3-2B         & 0.554 & 0.774 & 0.220 & 0.227 & 0.231 & 0.752 \\
 & InternVL3-8B         & 0.618 & 0.811 & 0.192 & 0.204 & 0.202 & 0.730 \\
 & InternVL3-38B        & 0.549 & 0.767 & 0.218 & 0.231 & 0.237 & 0.714 \\
\cmidrule(lr){2-8}
 & LLaVA-NeXT-7B        & 0.434 & 0.640 & 0.206 & 0.213 & 0.226 & 0.670 \\
 & LLaVA-NeXT-34B       & 0.472 & 0.751 & 0.279 & 0.285 & 0.293 & 0.632 \\
\bottomrule
\end{tabular}
% }
\end{table}

\begin{table}[t]
\centering
\small
\caption{Effect of model scale on accuracy, confidence, and calibration (sampling-based confidence) on VQA-RAD. Overconf.\ Gap = Mean Conf.\ $-$ Accuracy. $\uparrow$: higher is better, $\downarrow$: lower is better.}
\label{tab:scaling_analysis_rad_vqa}
% \resizebox{\textwidth}{!}{%
\begin{tabular}{@{}llccccccc@{}}
\toprule
\textbf{Type} & \textbf{Model} & \textbf{Acc.} ($\uparrow$) & \textbf{Conf.} & \textbf{Gap} ($\downarrow$) & \textbf{ECE} ($\downarrow$) & \textbf{ACE} ($\downarrow$) & \textbf{AUROC} ($\uparrow$) \\
\midrule
\multirow{8}{*}{\rotatebox[origin=c]{90}{\textbf{Closed}}}
 & Qwen3-VL-2B          & 0.689 & 0.949 & 0.260 & 0.288 & 0.281 & 0.560 \\
 & Qwen3-VL-8B          & 0.741 & 0.974 & 0.233 & 0.247 & 0.240 & 0.537 \\
 & Qwen3-VL-32B         & 0.765 & 0.975 & 0.210 & 0.230 & 0.213 & 0.551 \\
\cmidrule(lr){2-8}
 & InternVL3-2B         & 0.633 & 0.837 & 0.204 & 0.232 & 0.242 & 0.687 \\
 & InternVL3-8B         & 0.757 & 0.887 & 0.130 & 0.188 & 0.194 & 0.616 \\
 & InternVL3-38B        & 0.745 & 0.798 & 0.053 & 0.124 & 0.171 & 0.666 \\
\cmidrule(lr){2-8}
 & LLaVA-NeXT-7B        & 0.606 & 0.705 & 0.099 & 0.133 & 0.229 & 0.618 \\
 & LLaVA-NeXT-34B       & 0.558 & 0.737 & 0.179 & 0.212 & 0.261 & 0.625 \\
\midrule
\multirow{8}{*}{\rotatebox[origin=c]{90}{\textbf{Open}}}
 & Qwen3-VL-2B          & 0.400 & 0.758 & 0.358 & 0.430 & 0.401 & 0.666 \\
 & Qwen3-VL-8B          & 0.440 & 0.892 & 0.452 & 0.468 & 0.453 & 0.675 \\
 & Qwen3-VL-32B         & 0.515 & 0.974 & 0.459 & 0.465 & 0.459 & 0.577 \\
\cmidrule(lr){2-8}
 & InternVL3-2B         & 0.415 & 0.676 & 0.261 & 0.300 & 0.319 & 0.727 \\
 & InternVL3-8B         & 0.430 & 0.738 & 0.308 & 0.350 & 0.361 & 0.654 \\
 & InternVL3-38B        & 0.450 & 0.680 & 0.230 & 0.279 & 0.261 & 0.769 \\
\cmidrule(lr){2-8}
 & LLaVA-NeXT-7B        & 0.315 & 0.534 & 0.219 & 0.297 & 0.305 & 0.745 \\
 & LLaVA-NeXT-34B       & 0.354 & 0.674 & 0.320 & 0.342 & 0.343 & 0.713 \\
\midrule
\multirow{8}{*}{\rotatebox[origin=c]{90}{\textbf{Overall}}}
 & Qwen3-VL-2B          & 0.561 & 0.864 & 0.303 & 0.351 & 0.334 & 0.606 \\
 & Qwen3-VL-8B          & 0.608 & 0.937 & 0.330 & 0.344 & 0.334 & 0.598 \\
 & Qwen3-VL-32B         & 0.654 & 0.975 & 0.320 & 0.333 & 0.321 & 0.563 \\
\cmidrule(lr){2-8}
 & InternVL3-2B         & 0.537 & 0.766 & 0.229 & 0.262 & 0.276 & 0.705 \\
 & InternVL3-8B         & 0.612 & 0.821 & 0.209 & 0.259 & 0.267 & 0.633 \\
 & InternVL3-38B        & 0.614 & 0.746 & 0.132 & 0.192 & 0.211 & 0.711 \\
\cmidrule(lr){2-8}
 & LLaVA-NeXT-7B        & 0.477 & 0.629 & 0.152 & 0.205 & 0.263 & 0.674 \\
 & LLaVA-NeXT-34B       & 0.516 & 0.724 & 0.208 & 0.239 & 0.278 & 0.643 \\
\bottomrule
\end{tabular}
% }
\end{table}

\begin{table}[t]
\centering

\small
\caption{Effect of model scale on accuracy, confidence, and calibration (sampling-based confidence) on SLAKE-EN. Overconf.\ Gap = Mean Conf.\ $-$ Accuracy. $\uparrow$: higher is better, $\downarrow$: lower is better.}
\label{tab:scaling_analysis_slake}
% \resizebox{\textwidth}{!}{%
\begin{tabular}{@{}llccccccc@{}}
\toprule
\textbf{Type} & \textbf{Model} & \textbf{Acc.} ($\uparrow$) & \textbf{Conf.} & \textbf{Gap} ($\downarrow$) & \textbf{ECE} ($\downarrow$) & \textbf{ACE} ($\downarrow$) & \textbf{AUROC} ($\uparrow$) \\
\midrule
\multirow{8}{*}{\rotatebox[origin=c]{90}{\textbf{Closed}}}
 & Qwen3-VL-2B          & 0.786 & 0.974 & 0.188 & 0.199 & 0.189 & 0.607 \\
 & Qwen3-VL-8B          & 0.786 & 0.984 & 0.198 & 0.206 & 0.198 & 0.560 \\
 & Qwen3-VL-32B         & 0.803 & 1.000 & 0.197 & 0.197 & 0.197 & 0.500 \\
\cmidrule(lr){2-8}
 & InternVL3-2B         & 0.738 & 0.870 & 0.132 & 0.175 & 0.181 & 0.686 \\
 & InternVL3-8B         & 0.800 & 0.927 & 0.127 & 0.158 & 0.152 & 0.674 \\
 & InternVL3-38B        & 0.825 & 0.871 & 0.046 & 0.107 & 0.120 & 0.695 \\
\cmidrule(lr){2-8}
 & LLaVA-NeXT-7B        & 0.645 & 0.736 & 0.091 & 0.132 & 0.186 & 0.600 \\
 & LLaVA-NeXT-34B       & 0.656 & 0.845 & 0.188 & 0.202 & 0.217 & 0.602 \\
\midrule
\multirow{8}{*}{\rotatebox[origin=c]{90}{\textbf{Open}}}
 & Qwen3-VL-2B          & 0.582 & 0.840 & 0.258 & 0.298 & 0.280 & 0.690 \\
 & Qwen3-VL-8B          & 0.619 & 0.950 & 0.331 & 0.339 & 0.331 & 0.609 \\
 & Qwen3-VL-32B         & 0.694 & 0.959 & 0.265 & 0.272 & 0.265 & 0.608 \\
\cmidrule(lr){2-8}
 & InternVL3-2B         & 0.605 & 0.825 & 0.220 & 0.230 & 0.232 & 0.732 \\
 & InternVL3-8B         & 0.640 & 0.856 & 0.216 & 0.229 & 0.220 & 0.733 \\
 & InternVL3-38B        & 0.593 & 0.813 & 0.220 & 0.238 & 0.228 & 0.673 \\
\cmidrule(lr){2-8}
 & LLaVA-NeXT-7B        & 0.404 & 0.636 & 0.233 & 0.250 & 0.250 & 0.726 \\
 & LLaVA-NeXT-34B       & 0.487 & 0.744 & 0.257 & 0.289 & 0.299 & 0.659 \\
\midrule
\multirow{8}{*}{\rotatebox[origin=c]{90}{\textbf{Overall}}}
 & Qwen3-VL-2B          & 0.650 & 0.885 & 0.235 & 0.265 & 0.250 & 0.662 \\
 & Qwen3-VL-8B          & 0.675 & 0.962 & 0.287 & 0.294 & 0.287 & 0.593 \\
 & Qwen3-VL-32B         & 0.730 & 0.973 & 0.242 & 0.247 & 0.242 & 0.572 \\
\cmidrule(lr){2-8}
 & InternVL3-2B         & 0.649 & 0.840 & 0.191 & 0.212 & 0.215 & 0.717 \\
 & InternVL3-8B         & 0.694 & 0.880 & 0.186 & 0.205 & 0.198 & 0.713 \\
 & InternVL3-38B        & 0.671 & 0.833 & 0.162 & 0.194 & 0.192 & 0.680 \\
\cmidrule(lr){2-8}
 & LLaVA-NeXT-7B        & 0.484 & 0.670 & 0.185 & 0.210 & 0.229 & 0.684 \\
 & LLaVA-NeXT-34B       & 0.577 & 0.798 & 0.220 & 0.242 & 0.255 & 0.629 \\
\bottomrule
\end{tabular}
% }
\end{table}

\begin{table}[t]
\centering
\small
\caption{Effect of model scale on accuracy, confidence, and calibration (sampling-based confidence) on VQA-Med-2019. Overconf.\ Gap = Mean Conf.\ $-$ Accuracy. $\uparrow$: higher is better, $\downarrow$: lower is better.}
\label{tab:scaling_analysis_vqa_med_2019}
% \resizebox{\textwidth}{!}{%
\begin{tabular}{@{}llccccccc@{}}
\toprule
\textbf{Type} & \textbf{Model} & \textbf{Acc.} ($\uparrow$) & \textbf{Conf.} & \textbf{Gap} ($\downarrow$) & \textbf{ECE} ($\downarrow$) & \textbf{ACE} ($\downarrow$) & \textbf{AUROC} ($\uparrow$) \\
\midrule
\multirow{8}{*}{\rotatebox[origin=c]{90}{\textbf{Closed}}}
 & Qwen3-VL-2B          & 0.688 & 0.973 & 0.285 & 0.286 & 0.286 & 0.567 \\
 & Qwen3-VL-8B          & 0.750 & 0.980 & 0.230 & 0.242 & 0.242 & 0.549 \\
 & Qwen3-VL-32B         & 0.703 & 1.000 & 0.297 & 0.296 & 0.296 & 0.500 \\
\cmidrule(lr){2-8}
 & InternVL3-2B         & 0.703 & 0.866 & 0.162 & 0.210 & 0.199 & 0.739 \\
 & InternVL3-8B         & 0.781 & 0.905 & 0.124 & 0.163 & 0.179 & 0.762 \\
 & InternVL3-38B        & 0.609 & 0.848 & 0.239 & 0.338 & 0.312 & 0.463 \\
\cmidrule(lr){2-8}
 & LLaVA-NeXT-7B        & 0.578 & 0.795 & 0.217 & 0.258 & 0.286 & 0.556 \\
 & LLaVA-NeXT-34B       & 0.609 & 0.835 & 0.225 & 0.291 & 0.283 & 0.569 \\
\midrule
\multirow{8}{*}{\rotatebox[origin=c]{90}{\textbf{Open}}}
 & Qwen3-VL-2B          & 0.328 & 0.777 & 0.449 & 0.461 & 0.450 & 0.704 \\
 & Qwen3-VL-8B          & 0.413 & 0.899 & 0.486 & 0.488 & 0.486 & 0.657 \\
 & Qwen3-VL-32B         & 0.445 & 0.881 & 0.436 & 0.441 & 0.436 & 0.704 \\
\cmidrule(lr){2-8}
 & InternVL3-2B         & 0.319 & 0.608 & 0.290 & 0.294 & 0.292 & 0.812 \\
 & InternVL3-8B         & 0.417 & 0.619 & 0.201 & 0.212 & 0.226 & 0.801 \\
 & InternVL3-38B        & 0.174 & 0.618 & 0.443 & 0.444 & 0.443 & 0.680 \\
\cmidrule(lr){2-8}
 & LLaVA-NeXT-7B        & 0.245 & 0.557 & 0.311 & 0.323 & 0.320 & 0.625 \\
 & LLaVA-NeXT-34B       & 0.259 & 0.688 & 0.429 & 0.436 & 0.433 & 0.607 \\
\midrule
\multirow{8}{*}{\rotatebox[origin=c]{90}{\textbf{Overall}}}
 & Qwen3-VL-2B          & 0.374 & 0.802 & 0.428 & 0.438 & 0.429 & 0.687 \\
 & Qwen3-VL-8B          & 0.456 & 0.909 & 0.453 & 0.457 & 0.455 & 0.643 \\
 & Qwen3-VL-32B         & 0.478 & 0.896 & 0.418 & 0.422 & 0.418 & 0.678 \\
\cmidrule(lr){2-8}
 & InternVL3-2B         & 0.368 & 0.641 & 0.273 & 0.283 & 0.280 & 0.803 \\
 & InternVL3-8B         & 0.464 & 0.655 & 0.191 & 0.206 & 0.220 & 0.796 \\
 & InternVL3-38B        & 0.230 & 0.647 & 0.417 & 0.431 & 0.426 & 0.652 \\
\cmidrule(lr){2-8}
 & LLaVA-NeXT-7B        & 0.288 & 0.587 & 0.299 & 0.315 & 0.316 & 0.617 \\
 & LLaVA-NeXT-34B       & 0.304 & 0.707 & 0.403 & 0.417 & 0.414 & 0.602 \\
\bottomrule
\end{tabular}
% }
\end{table}

\begin{table}[t]
\centering
\caption{Calibration quality of different confidence prompting strategies (uncalibrated) across all models. ECE, ACE, and AUROC are reported per dataset and question type, with Avg computed on the pooled dataset. Dashes indicate degenerate cases where the model achieved 0\% accuracy, making calibration metrics undefined. No single prompting variant consistently outperforms the base sampling or verbalized methods across all settings, reconfirming the finding that prompting strategies alone do not reliably improve calibration. \textbf{Bold}: best, \underline{underline}: second best per column within each model.}
\label{tab:prompting_strategies}
\resizebox{\textwidth}{!}{%
\begin{tabular}{@{}ll|cccccc|cccccc|cccccc|cccccc@{}}
\toprule
 &  & \multicolumn{6}{c|}{VQA-RAD} & \multicolumn{6}{c|}{SLAKE} & \multicolumn{6}{c|}{VQA-Med} & \multicolumn{6}{c}{Avg} \\
\cmidrule(lr){3-8} \cmidrule(lr){9-14} \cmidrule(lr){15-20} \cmidrule(lr){21-26}
 &  & \multicolumn{3}{c}{Closed} & \multicolumn{3}{c|}{Open} & \multicolumn{3}{c}{Closed} & \multicolumn{3}{c|}{Open} & \multicolumn{3}{c}{Closed} & \multicolumn{3}{c|}{Open} & \multicolumn{3}{c}{Closed} & \multicolumn{3}{c}{Open} \\
\cmidrule(lr){3-5} \cmidrule(lr){6-8} \cmidrule(lr){9-11} \cmidrule(lr){12-14} \cmidrule(lr){15-17} \cmidrule(lr){18-20} \cmidrule(lr){21-23} \cmidrule(lr){24-26}
 & & ECE$\downarrow$ & ACE$\downarrow$ & AUC$\uparrow$ & ECE$\downarrow$ & ACE$\downarrow$ & AUC$\uparrow$ & ECE$\downarrow$ & ACE$\downarrow$ & AUC$\uparrow$ & ECE$\downarrow$ & ACE$\downarrow$ & AUC$\uparrow$ & ECE$\downarrow$ & ACE$\downarrow$ & AUC$\uparrow$ & ECE$\downarrow$ & ACE$\downarrow$ & AUC$\uparrow$ & ECE$\downarrow$ & ACE$\downarrow$ & AUC$\uparrow$ & ECE$\downarrow$ & ACE$\downarrow$ & AUC$\uparrow$ \\
\midrule
\multirow{8}{*}{Q-2B} & Samp. (Base) & 0.288 & 0.281 & 0.560 & 0.430 & 0.401 & 0.666 & 0.199 & 0.189 & 0.607 & 0.298 & \underline{0.280} & \underline{0.690} & \textbf{0.286} & \underline{0.286} & \underline{0.567} & 0.461 & 0.450 & 0.704 & 0.235 & 0.224 & 0.586 & \underline{0.356} & \underline{0.345} & \underline{0.695} \\
 & Samp. (CoT) & \underline{0.244} & \textbf{0.221} & \underline{0.637} & \textbf{0.321} & \textbf{0.324} & \textbf{0.782} & 0.233 & 0.231 & \textbf{0.671} & \textbf{0.275} & \textbf{0.262} & \textbf{0.716} & \underline{0.291} & \textbf{0.285} & \textbf{0.782} & \textbf{0.338} & \textbf{0.331} & \underline{0.710} & 0.230 & 0.225 & \textbf{0.668} & \textbf{0.292} & \textbf{0.286} & \textbf{0.734} \\
\cmidrule(lr){2-26}
 & Verb. (Vanilla) & 0.245 & \underline{0.257} & 0.518 & \underline{0.369} & \underline{0.378} & 0.625 & \textbf{0.150} & \underline{0.172} & 0.586 & 0.346 & 0.337 & 0.603 & 0.297 & 0.325 & 0.549 & 0.547 & 0.547 & 0.577 & \textbf{0.200} & \textbf{0.199} & 0.559 & 0.414 & 0.411 & 0.590 \\
 & Verb. (Van.+CoT) & 0.315 & 0.320 & \textbf{0.641} & 0.514 & 0.514 & \underline{0.730} & 0.290 & 0.293 & 0.570 & 0.431 & 0.425 & 0.644 & 0.315 & 0.288 & 0.516 & \underline{0.442} & \underline{0.442} & \textbf{0.720} & 0.302 & 0.299 & \underline{0.596} & 0.447 & 0.446 & 0.684 \\
 & Verb. (Punish) & \textbf{0.238} & 0.258 & 0.585 & 0.414 & 0.440 & 0.619 & \underline{0.157} & \textbf{0.168} & \underline{0.620} & 0.381 & 0.374 & 0.632 & 0.291 & 0.315 & 0.486 & 0.573 & 0.573 & 0.614 & \underline{0.201} & \underline{0.206} & 0.594 & 0.447 & 0.444 & 0.616 \\
 & Verb. (Top-K) & 0.316 & 0.334 & 0.474 & 0.523 & 0.530 & 0.638 & 0.334 & 0.333 & 0.421 & 0.420 & 0.420 & 0.633 & 0.347 & 0.367 & 0.342 & 0.594 & 0.594 & 0.571 & 0.329 & 0.321 & 0.434 & 0.492 & 0.492 & 0.613 \\
 & Verb. (Two-Stage) & 0.333 & 0.321 & 0.557 & 0.738 & 0.738 & 0.648 & 0.271 & 0.275 & 0.530 & 0.851 & 0.846 & 0.473 & 0.392 & 0.362 & 0.420 & 0.888 & 0.888 & 0.575 & 0.306 & 0.299 & 0.533 & 0.846 & 0.842 & 0.514 \\
 & Verb. (Linguistic) & 0.331 & 0.345 & 0.494 & 0.395 & 0.406 & 0.663 & 0.405 & 0.398 & 0.469 & \underline{0.288} & 0.284 & 0.661 & 0.368 & 0.387 & 0.418 & 0.496 & 0.498 & 0.646 & 0.367 & 0.362 & 0.479 & 0.363 & 0.358 & 0.648 \\
\midrule
\multirow{8}{*}{Q-8B} & Samp. (Base) & 0.247 & 0.240 & 0.537 & 0.468 & 0.453 & 0.675 & 0.206 & 0.198 & 0.560 & 0.339 & 0.331 & 0.609 & 0.242 & 0.242 & 0.549 & 0.488 & 0.486 & 0.657 & 0.219 & 0.214 & 0.551 & 0.401 & 0.400 & 0.643 \\
 & Samp. (CoT) & 0.223 & \textbf{0.201} & \underline{0.614} & \textbf{0.337} & \textbf{0.332} & \textbf{0.789} & 0.249 & 0.235 & 0.613 & \underline{0.289} & \underline{0.282} & \textbf{0.755} & 0.222 & 0.224 & 0.584 & \textbf{0.354} & \textbf{0.346} & \underline{0.768} & 0.219 & 0.210 & 0.608 & \textbf{0.309} & \textbf{0.306} & \textbf{0.766} \\
\cmidrule(lr){2-26}
 & Verb. (Vanilla) & 0.230 & 0.250 & 0.575 & 0.389 & 0.406 & \underline{0.753} & \underline{0.171} & \textbf{0.180} & 0.616 & 0.299 & 0.299 & 0.685 & \underline{0.159} & 0.218 & 0.518 & 0.455 & 0.457 & 0.737 & \textbf{0.184} & \textbf{0.180} & 0.600 & 0.363 & 0.363 & \underline{0.710} \\
 & Verb. (Van.+CoT) & \textbf{0.198} & \underline{0.225} & \textbf{0.651} & 0.425 & 0.424 & 0.739 & 0.217 & 0.226 & \textbf{0.796} & 0.342 & 0.343 & 0.686 & 0.232 & 0.248 & \textbf{0.732} & \underline{0.410} & \underline{0.409} & 0.727 & 0.208 & 0.207 & \textbf{0.733} & 0.376 & 0.375 & 0.709 \\
 & Verb. (Punish) & \underline{0.215} & 0.242 & 0.507 & 0.402 & 0.423 & 0.684 & 0.188 & 0.208 & 0.621 & 0.322 & 0.318 & 0.651 & 0.164 & 0.215 & 0.637 & 0.469 & 0.469 & 0.691 & \underline{0.192} & \underline{0.195} & 0.579 & 0.381 & 0.379 & 0.672 \\
 & Verb. (Top-K) & 0.264 & 0.291 & 0.405 & \underline{0.355} & \underline{0.374} & 0.662 & 0.298 & 0.329 & 0.373 & 0.415 & 0.419 & 0.529 & 0.231 & 0.304 & 0.453 & 0.460 & 0.461 & 0.670 & 0.278 & 0.279 & 0.396 & 0.420 & 0.420 & 0.598 \\
 & Verb. (Two-Stage) & 0.245 & 0.265 & 0.500 & 0.731 & 0.731 & 0.749 & 0.199 & 0.193 & 0.654 & 0.784 & 0.779 & 0.603 & 0.162 & \textbf{0.181} & 0.465 & 0.683 & 0.675 & \textbf{0.788} & 0.210 & 0.216 & 0.576 & 0.743 & 0.737 & 0.699 \\
 & Verb. (Linguistic) & 0.300 & 0.335 & 0.603 & 0.386 & 0.413 & 0.702 & \textbf{0.157} & \underline{0.189} & \underline{0.761} & \textbf{0.287} & \textbf{0.279} & \underline{0.695} & \textbf{0.131} & \underline{0.194} & \underline{0.670} & 0.439 & 0.443 & 0.714 & 0.204 & 0.214 & \underline{0.701} & \underline{0.350} & \underline{0.342} & 0.702 \\
\midrule
\multirow{8}{*}{Q-32B} & Samp. (Base) & 0.230 & 0.213 & 0.551 & 0.465 & 0.459 & 0.577 & 0.197 & 0.197 & 0.500 & 0.272 & 0.265 & 0.608 & 0.296 & 0.296 & 0.500 & 0.441 & 0.436 & 0.704 & 0.219 & 0.212 & 0.521 & 0.351 & 0.349 & 0.641 \\
 & Samp. (CoT) & 0.247 & 0.229 & \textbf{0.687} & 0.284 & \underline{0.271} & \underline{0.774} & 0.208 & 0.199 & 0.552 & \textbf{0.199} & \textbf{0.179} & \underline{0.767} & 0.228 & \textbf{0.215} & 0.552 & \textbf{0.294} & \textbf{0.281} & \textbf{0.769} & 0.212 & 0.199 & 0.610 & \textbf{0.223} & \textbf{0.215} & \textbf{0.771} \\
\cmidrule(lr){2-26}
 & Verb. (Vanilla) & \textbf{0.174} & \textbf{0.210} & 0.559 & 0.339 & 0.364 & 0.708 & \textbf{0.143} & \textbf{0.163} & 0.618 & 0.257 & 0.259 & 0.724 & \textbf{0.200} & 0.240 & 0.607 & 0.417 & 0.420 & 0.675 & \textbf{0.155} & \textbf{0.166} & 0.596 & 0.321 & 0.321 & 0.702 \\
 & Verb. (Van.+CoT) & 0.211 & 0.239 & 0.619 & \underline{0.248} & \textbf{0.264} & \textbf{0.856} & \underline{0.148} & 0.174 & \textbf{0.695} & \underline{0.209} & \underline{0.207} & \textbf{0.786} & 0.250 & 0.247 & \underline{0.628} & \underline{0.335} & \underline{0.338} & 0.705 & \underline{0.169} & \underline{0.173} & \textbf{0.661} & \underline{0.256} & \underline{0.253} & \underline{0.771} \\
 & Verb. (Punish) & \underline{0.183} & \underline{0.211} & 0.560 & 0.355 & 0.363 & 0.697 & 0.153 & \underline{0.166} & \underline{0.687} & 0.265 & 0.266 & 0.719 & \underline{0.208} & \underline{0.234} & 0.623 & 0.432 & 0.434 & 0.676 & 0.169 & 0.175 & 0.634 & 0.332 & 0.332 & 0.695 \\
 & Verb. (Top-K) & 0.183 & 0.245 & 0.463 & \textbf{0.228} & 0.300 & 0.613 & 0.378 & 0.384 & 0.463 & 0.275 & 0.287 & 0.484 & 0.232 & 0.298 & 0.389 & 0.405 & 0.420 & 0.580 & 0.269 & 0.272 & 0.430 & 0.295 & 0.287 & 0.533 \\
 & Verb. (Two-Stage) & 0.223 & 0.247 & \underline{0.619} & 0.871 & 0.868 & 0.473 & 0.154 & 0.169 & 0.681 & 0.834 & 0.834 & 0.449 & 0.255 & 0.265 & \textbf{0.633} & 0.903 & 0.903 & 0.556 & 0.186 & 0.188 & \underline{0.654} & 0.861 & 0.861 & 0.466 \\
 & Verb. (Linguistic) & 0.358 & 0.370 & 0.482 & 0.373 & 0.395 & 0.705 & 0.233 & 0.260 & 0.490 & 0.251 & 0.252 & 0.632 & 0.352 & 0.313 & 0.511 & 0.402 & 0.406 & \underline{0.729} & 0.281 & 0.282 & 0.501 & 0.317 & 0.316 & 0.679 \\
\midrule
\multirow{8}{*}{I-2B} & Samp. (Base) & 0.232 & \textbf{0.242} & \textbf{0.687} & \underline{0.300} & \underline{0.319} & 0.727 & \underline{0.175} & \textbf{0.181} & \textbf{0.686} & \textbf{0.230} & \textbf{0.232} & \underline{0.732} & 0.210 & 0.199 & \textbf{0.739} & \textbf{0.294} & \textbf{0.292} & \textbf{0.812} & \textbf{0.181} & \underline{0.194} & \textbf{0.693} & \textbf{0.250} & \textbf{0.249} & \underline{0.782} \\
 & Samp. (CoT) & 0.265 & 0.284 & 0.543 & \textbf{0.284} & \textbf{0.285} & \underline{0.774} & 0.222 & 0.227 & \underline{0.686} & \underline{0.239} & \underline{0.233} & 0.722 & 0.277 & 0.301 & \underline{0.722} & \underline{0.329} & \underline{0.324} & 0.712 & 0.224 & 0.236 & \underline{0.642} & \underline{0.263} & \underline{0.261} & 0.749 \\
\cmidrule(lr){2-26}
 & Verb. (Vanilla) & \underline{0.232} & \underline{0.264} & \underline{0.675} & 0.428 & 0.445 & 0.514 & \textbf{0.170} & \underline{0.194} & 0.554 & 0.340 & 0.339 & 0.613 & \textbf{0.136} & \textbf{0.160} & 0.587 & 0.550 & 0.552 & 0.447 & \underline{0.190} & \textbf{0.192} & 0.599 & 0.421 & 0.421 & 0.543 \\
 & Verb. (Van.+CoT) & 0.295 & 0.303 & 0.575 & 0.515 & 0.522 & 0.656 & 0.339 & 0.329 & 0.610 & 0.396 & 0.395 & 0.595 & 0.267 & 0.294 & 0.544 & 0.624 & 0.624 & 0.564 & 0.314 & 0.303 & 0.590 & 0.488 & 0.487 & 0.597 \\
 & Verb. (Punish) & 0.238 & 0.270 & 0.612 & 0.467 & 0.483 & 0.512 & 0.179 & 0.216 & 0.551 & 0.345 & 0.345 & 0.604 & \underline{0.155} & \underline{0.193} & 0.656 & 0.560 & 0.559 & 0.402 & 0.199 & 0.197 & 0.580 & 0.433 & 0.433 & 0.522 \\
 & Verb. (Top-K) & \textbf{0.221} & 0.266 & 0.601 & 0.374 & 0.399 & 0.630 & 0.211 & 0.233 & 0.583 & 0.326 & 0.326 & 0.668 & 0.255 & 0.297 & 0.556 & 0.517 & 0.514 & 0.518 & 0.216 & 0.213 & 0.586 & 0.393 & 0.393 & 0.621 \\
 & Verb. (Two-Stage) & 0.277 & 0.305 & 0.626 & 0.876 & 0.876 & \textbf{0.904} & 0.416 & 0.416 & 0.512 & 0.896 & 0.896 & \textbf{0.766} & 0.307 & 0.284 & 0.616 & 0.889 & 0.889 & \underline{0.762} & 0.351 & 0.343 & 0.563 & 0.891 & 0.891 & \textbf{0.802} \\
 & Verb. (Linguistic) & 0.645 & 0.631 & 0.169 & 0.867 & 0.867 & 0.503 & - & - & - & 0.820 & 0.820 & 0.531 & - & - & - & - & - & - & 0.645 & 0.631 & 0.169 & 0.831 & 0.831 & 0.525 \\
\midrule
\multirow{8}{*}{I-8B} & Samp. (Base) & \underline{0.188} & \textbf{0.194} & 0.616 & 0.350 & 0.361 & 0.654 & 0.158 & \underline{0.152} & 0.674 & 0.229 & 0.220 & \underline{0.733} & 0.163 & 0.179 & \textbf{0.762} & \textbf{0.212} & \textbf{0.226} & \textbf{0.801} & 0.151 & \underline{0.152} & 0.663 & 0.230 & 0.228 & \textbf{0.764} \\
 & Samp. (CoT) & \textbf{0.179} & 0.220 & \underline{0.645} & \underline{0.346} & \underline{0.335} & \underline{0.654} & \underline{0.143} & \textbf{0.139} & \textbf{0.753} & \underline{0.207} & \underline{0.201} & 0.730 & 0.152 & \textbf{0.146} & 0.694 & 0.236 & 0.235 & \underline{0.800} & \textbf{0.141} & \textbf{0.143} & \textbf{0.714} & \underline{0.217} & \underline{0.210} & \underline{0.762} \\
\cmidrule(lr){2-26}
 & Verb. (Vanilla) & 0.203 & 0.233 & 0.503 & 0.440 & 0.458 & 0.616 & 0.165 & 0.197 & 0.535 & 0.264 & 0.265 & 0.598 & \textbf{0.089} & \underline{0.150} & 0.491 & 0.458 & 0.461 & 0.633 & 0.169 & 0.176 & 0.522 & 0.353 & 0.352 & 0.614 \\
 & Verb. (Van.+CoT) & 0.249 & 0.281 & 0.533 & 0.438 & 0.444 & 0.579 & 0.171 & 0.183 & 0.599 & 0.265 & 0.273 & 0.643 & 0.180 & 0.213 & 0.550 & 0.439 & 0.439 & 0.591 & 0.194 & 0.197 & 0.570 & 0.347 & 0.346 & 0.610 \\
 & Verb. (Punish) & 0.191 & \underline{0.220} & 0.495 & 0.408 & 0.427 & 0.576 & 0.153 & 0.182 & 0.536 & 0.270 & 0.271 & 0.561 & \underline{0.090} & 0.152 & 0.500 & 0.482 & 0.482 & 0.572 & 0.160 & 0.167 & 0.517 & 0.359 & 0.358 & 0.566 \\
 & Verb. (Top-K) & 0.205 & 0.247 & 0.620 & \textbf{0.133} & \textbf{0.224} & \textbf{0.687} & \textbf{0.124} & 0.170 & 0.620 & \textbf{0.152} & \textbf{0.148} & \textbf{0.743} & 0.220 & 0.246 & 0.485 & \underline{0.227} & \underline{0.233} & 0.743 & \underline{0.147} & 0.155 & 0.602 & \textbf{0.140} & \textbf{0.149} & 0.742 \\
 & Verb. (Two-Stage) & 0.269 & 0.299 & \textbf{0.659} & 0.874 & 0.875 & 0.397 & 0.155 & 0.177 & \underline{0.706} & 0.915 & 0.915 & 0.280 & 0.290 & 0.324 & \underline{0.755} & - & - & - & 0.207 & 0.200 & \underline{0.688} & 0.906 & 0.906 & 0.299 \\
 & Verb. (Linguistic) & 0.495 & 0.481 & 0.442 & 0.397 & 0.422 & 0.609 & 0.542 & 0.507 & 0.441 & 0.234 & 0.234 & 0.623 & 0.626 & 0.564 & 0.387 & 0.480 & 0.481 & 0.553 & 0.531 & 0.512 & 0.442 & 0.337 & 0.334 & 0.596 \\
\midrule
\multirow{8}{*}{I-38B} & Samp. (Base) & \textbf{0.124} & \textbf{0.171} & \textbf{0.666} & \textbf{0.279} & \textbf{0.261} & \textbf{0.769} & \textbf{0.107} & \textbf{0.120} & \underline{0.695} & \underline{0.238} & \underline{0.228} & 0.673 & 0.338 & 0.312 & 0.463 & \underline{0.444} & \underline{0.443} & 0.680 & \textbf{0.106} & \textbf{0.122} & \underline{0.662} & \underline{0.294} & \underline{0.294} & \underline{0.740} \\
 & Samp. (CoT) & 0.190 & 0.220 & \underline{0.638} & 0.330 & 0.332 & 0.679 & 0.142 & \underline{0.139} & \textbf{0.736} & \textbf{0.224} & \textbf{0.205} & 0.670 & 0.338 & 0.359 & 0.390 & \textbf{0.262} & \textbf{0.262} & \textbf{0.770} & 0.158 & 0.169 & \textbf{0.667} & \textbf{0.229} & \textbf{0.233} & 0.727 \\
\cmidrule(lr){2-26}
 & Verb. (Vanilla) & 0.183 & 0.232 & 0.598 & 0.411 & 0.427 & 0.720 & 0.154 & 0.175 & 0.632 & 0.353 & 0.351 & 0.655 & \textbf{0.069} & \underline{0.149} & \textbf{0.791} & 0.495 & 0.500 & 0.629 & 0.150 & 0.164 & 0.626 & 0.407 & 0.406 & 0.656 \\
 & Verb. (Van.+CoT) & 0.179 & \underline{0.208} & 0.611 & 0.442 & 0.466 & 0.685 & 0.191 & 0.203 & 0.651 & 0.303 & 0.305 & 0.678 & 0.174 & 0.226 & 0.474 & 0.460 & 0.462 & 0.655 & 0.180 & 0.184 & 0.615 & 0.375 & 0.375 & 0.674 \\
 & Verb. (Punish) & 0.188 & 0.226 & 0.577 & 0.433 & 0.457 & \underline{0.737} & 0.163 & 0.175 & 0.687 & 0.358 & 0.355 & 0.680 & 0.081 & \textbf{0.145} & 0.573 & 0.479 & 0.481 & 0.650 & 0.163 & 0.175 & 0.641 & 0.408 & 0.407 & 0.679 \\
 & Verb. (Top-K) & 0.214 & 0.297 & 0.456 & \underline{0.287} & \underline{0.327} & 0.630 & 0.606 & 0.601 & 0.234 & 0.321 & 0.326 & 0.552 & 0.273 & 0.357 & 0.414 & 0.454 & 0.457 & 0.601 & 0.407 & 0.406 & 0.294 & 0.353 & 0.352 & 0.579 \\
 & Verb. (Two-Stage) & \underline{0.145} & 0.216 & 0.572 & 0.869 & 0.869 & 0.526 & \underline{0.118} & 0.144 & 0.654 & 0.904 & 0.904 & \textbf{0.871} & \underline{0.072} & 0.163 & 0.670 & 0.901 & 0.901 & \underline{0.745} & \underline{0.117} & \underline{0.131} & 0.622 & 0.897 & 0.897 & \textbf{0.753} \\
 & Verb. (Linguistic) & 0.315 & 0.337 & 0.536 & 0.455 & 0.472 & 0.623 & 0.210 & 0.237 & 0.648 & 0.330 & 0.327 & \underline{0.695} & 0.098 & 0.173 & \underline{0.733} & 0.492 & 0.495 & 0.664 & 0.237 & 0.249 & 0.618 & 0.401 & 0.398 & 0.672 \\
\midrule
\multirow{8}{*}{L-7B} & Samp. (Base) & \textbf{0.133} & \textbf{0.229} & 0.618 & \textbf{0.297} & \textbf{0.305} & \textbf{0.745} & \textbf{0.132} & \textbf{0.186} & \underline{0.600} & \textbf{0.250} & \textbf{0.250} & \textbf{0.726} & \underline{0.258} & \underline{0.286} & 0.556 & \textbf{0.323} & \textbf{0.320} & \underline{0.625} & \textbf{0.122} & \textbf{0.165} & \underline{0.598} & \textbf{0.259} & \textbf{0.256} & \underline{0.706} \\
 & Samp. (CoT) & \underline{0.302} & \underline{0.312} & 0.594 & \underline{0.392} & \underline{0.424} & 0.611 & \underline{0.212} & \underline{0.221} & 0.581 & \underline{0.338} & \underline{0.329} & 0.656 & \textbf{0.213} & \textbf{0.260} & \textbf{0.598} & \underline{0.423} & \underline{0.426} & 0.576 & \underline{0.235} & \underline{0.241} & 0.575 & \underline{0.367} & \underline{0.363} & 0.628 \\
\cmidrule(lr){2-26}
 & Verb. (Vanilla) & 0.357 & 0.356 & 0.549 & 0.505 & 0.508 & 0.589 & 0.377 & 0.367 & 0.566 & 0.450 & 0.446 & 0.618 & 0.453 & 0.456 & \underline{0.565} & 0.703 & 0.703 & 0.565 & 0.376 & 0.371 & 0.558 & 0.541 & 0.539 & 0.603 \\
 & Verb. (Van.+CoT) & 0.383 & 0.390 & \underline{0.618} & 0.505 & 0.500 & 0.672 & 0.409 & 0.398 & 0.529 & 0.408 & 0.404 & 0.631 & 0.517 & 0.478 & 0.544 & 0.633 & 0.633 & \textbf{0.672} & 0.408 & 0.401 & 0.573 & 0.495 & 0.488 & 0.642 \\
 & Verb. (Punish) & 0.352 & 0.353 & 0.552 & 0.566 & 0.568 & 0.590 & 0.370 & 0.350 & 0.569 & 0.476 & 0.471 & 0.616 & 0.475 & 0.485 & 0.488 & 0.718 & 0.718 & 0.586 & 0.373 & 0.363 & 0.560 & 0.568 & 0.566 & 0.607 \\
 & Verb. (Top-K) & 0.381 & 0.377 & 0.487 & 0.463 & 0.481 & 0.568 & 0.490 & 0.468 & 0.540 & 0.436 & 0.430 & 0.587 & 0.451 & 0.458 & 0.471 & 0.655 & 0.655 & 0.501 & 0.443 & 0.425 & 0.506 & 0.507 & 0.507 & 0.557 \\
 & Verb. (Two-Stage) & 0.323 & 0.326 & \textbf{0.758} & 0.749 & 0.749 & \underline{0.731} & 0.359 & 0.354 & \textbf{0.631} & 0.734 & 0.734 & \underline{0.720} & 0.444 & 0.364 & 0.453 & - & - & - & 0.349 & 0.347 & \textbf{0.683} & 0.738 & 0.738 & \textbf{0.722} \\
 & Verb. (Linguistic) & 0.471 & 0.457 & 0.525 & 0.527 & 0.536 & 0.576 & 0.442 & 0.429 & 0.594 & 0.464 & 0.449 & 0.606 & 0.374 & 0.381 & 0.478 & 0.710 & 0.710 & 0.611 & 0.423 & 0.420 & 0.582 & 0.551 & 0.537 & 0.583 \\
\midrule
\multirow{8}{*}{L-34B} & Samp. (Base) & \underline{0.212} & \underline{0.261} & \textbf{0.625} & \underline{0.342} & \underline{0.343} & 0.713 & \underline{0.202} & \textbf{0.217} & 0.602 & \underline{0.289} & \underline{0.299} & \underline{0.659} & 0.291 & 0.283 & 0.569 & \underline{0.436} & \underline{0.433} & 0.607 & \underline{0.200} & \underline{0.212} & \textbf{0.618} & \underline{0.355} & \underline{0.360} & \textbf{0.645} \\
 & Samp. (CoT) & 0.366 & 0.360 & 0.567 & 0.435 & 0.439 & 0.614 & 0.256 & 0.265 & 0.594 & 0.366 & 0.353 & 0.648 & 0.269 & \underline{0.264} & 0.474 & 0.482 & 0.484 & 0.490 & 0.283 & 0.277 & 0.562 & 0.422 & 0.415 & 0.568 \\
\cmidrule(lr){2-26}
 & Verb. (Vanilla) & 0.270 & 0.303 & 0.549 & 0.540 & 0.549 & 0.701 & 0.234 & \underline{0.245} & 0.606 & 0.455 & 0.452 & 0.608 & 0.265 & 0.299 & 0.525 & 0.630 & 0.630 & 0.606 & 0.247 & 0.241 & 0.580 & 0.556 & 0.551 & 0.613 \\
 & Verb. (Van.+CoT) & 0.375 & 0.385 & 0.505 & 0.445 & 0.448 & 0.586 & 0.301 & 0.304 & \underline{0.623} & 0.361 & 0.366 & 0.609 & 0.387 & 0.363 & \textbf{0.620} & 0.530 & 0.530 & \textbf{0.700} & 0.324 & 0.318 & 0.582 & 0.456 & 0.456 & \underline{0.642} \\
 & Verb. (Punish) & 0.260 & 0.299 & 0.509 & 0.566 & 0.572 & 0.607 & 0.239 & 0.255 & 0.599 & 0.454 & 0.459 & 0.547 & \underline{0.230} & 0.284 & 0.524 & 0.645 & 0.645 & 0.495 & 0.246 & 0.249 & 0.560 & 0.565 & 0.565 & 0.539 \\
 & Verb. (Top-K) & \textbf{0.092} & \textbf{0.213} & \underline{0.602} & \textbf{0.188} & \textbf{0.234} & \underline{0.730} & \textbf{0.176} & 0.246 & 0.614 & \textbf{0.211} & \textbf{0.261} & 0.576 & \textbf{0.105} & \textbf{0.227} & 0.509 & \textbf{0.366} & \textbf{0.366} & \underline{0.644} & \textbf{0.121} & \textbf{0.161} & \underline{0.601} & \textbf{0.283} & \textbf{0.290} & 0.619 \\
 & Verb. (Two-Stage) & 0.274 & 0.311 & 0.582 & - & - & - & 0.329 & 0.334 & 0.562 & - & - & - & 0.322 & 0.373 & \underline{0.574} & - & - & - & 0.299 & 0.299 & 0.565 & - & - & - \\
 & Verb. (Linguistic) & 0.493 & 0.478 & 0.483 & 0.578 & 0.590 & \textbf{0.735} & 0.445 & 0.446 & \textbf{0.632} & 0.450 & 0.449 & \textbf{0.660} & 0.296 & 0.339 & 0.429 & 0.643 & 0.643 & 0.587 & 0.449 & 0.423 & 0.557 & 0.564 & 0.562 & 0.624 \\
\bottomrule
\end{tabular}}
\end{table}

\clearpage

\section{Full Results on Post-hoc Calibration (Sections~\ref{sec:posthoc} and \ref{sec:hac})} \label{app:posthoc}

\subsection{Predictive Discrimination Gains} \label{app:auroc}
Table~\ref{tab:hac_auroc_all} extends Table~\ref{tab:hac_auroc} to per-dataset and per-model breakdowns. For open-ended questions, we reconfirmed that HAC consistently improved AUROC. For closed-ended questions, performance varied across datasets, with the largest gain on VQA-Med. Exploring a more robust HAC formulation that generalizes across question types remains a meaningful direction.

\begin{table}[h]
\centering
\caption{Post-hoc calibration comparison: AUROC ($\uparrow$) across calibration methods (5-fold CV). Each cell shows sampling\,/\,verbalized confidence. \textbf{Bold} = best per row.}
\label{tab:hac_auroc_all}
\begin{subtable}{\textwidth}
\centering
\caption{Pooled Medical VQA Datasets (mean $\pm$ std)}
\label{tab:hac_auroc_micro_avg}
\resizebox{\textwidth}{!}{%
\begin{tabular}{@{}ll|cc|cc|cc|cc|cc|cc@{}}
\toprule
 &  & \multicolumn{2}{c}{Uncalibrated} & \multicolumn{2}{c}{Platt Scaling} & \multicolumn{2}{c}{Isotonic Regr.} & \multicolumn{2}{c}{Hist. Binning} & \multicolumn{2}{c}{\textbf{HAC-Platt}} & \multicolumn{2}{c}{\textbf{HAC-Gate}} \\
\cmidrule(lr){3-4} \cmidrule(lr){5-6} \cmidrule(lr){7-8} \cmidrule(lr){9-10} \cmidrule(lr){11-12} \cmidrule(lr){13-14}
 &  & Samp. & Verb. & Samp. & Verb. & Samp. & Verb. & Samp. & Verb. & Samp. & Verb. & Samp. & Verb. \\
\midrule
\multirow{8}{*}{\rotatebox[origin=c]{90}{Closed}} & Qwen3-VL-2B & {\small \cellcolor[rgb]{0.85,0.94,0.86} 0.586 $\pm$ 0.054} & {\small \cellcolor[rgb]{0.88,0.95,0.87} 0.559 $\pm$ 0.014} & {\small \cellcolor[rgb]{0.85,0.94,0.86} 0.586 $\pm$ 0.054} & {\small \cellcolor[rgb]{0.88,0.95,0.87} 0.559 $\pm$ 0.014} & {\small \cellcolor[rgb]{0.86,0.95,0.86} 0.577 $\pm$ 0.039} & {\small \cellcolor[rgb]{0.88,0.95,0.87} 0.557 $\pm$ 0.015} & {\small \cellcolor[rgb]{0.86,0.94,0.86} 0.583 $\pm$ 0.041} & {\small \cellcolor[rgb]{0.92,0.97,0.91} 0.503 $\pm$ 0.005} & {\small \cellcolor[rgb]{0.77,0.91,0.85} 0.653 $\pm$ 0.051} & {\small \cellcolor[rgb]{0.79,0.92,0.85} 0.644 $\pm$ 0.035} & {\small \cellcolor[rgb]{0.77,0.91,0.85} \textbf{0.654 $\pm$ 0.051}} & {\small \cellcolor[rgb]{0.78,0.92,0.85} \textbf{0.648 $\pm$ 0.037}} \\
 & Qwen3-VL-8B & {\small \cellcolor[rgb]{0.89,0.96,0.87} 0.551 $\pm$ 0.031} & {\small \cellcolor[rgb]{0.84,0.94,0.85} 0.600 $\pm$ 0.045} & {\small \cellcolor[rgb]{0.89,0.96,0.87} 0.551 $\pm$ 0.031} & {\small \cellcolor[rgb]{0.84,0.94,0.85} 0.600 $\pm$ 0.045} & {\small \cellcolor[rgb]{0.88,0.95,0.87} 0.556 $\pm$ 0.031} & {\small \cellcolor[rgb]{0.83,0.93,0.85} 0.604 $\pm$ 0.044} & {\small \cellcolor[rgb]{0.89,0.96,0.88} 0.544 $\pm$ 0.022} & {\small \cellcolor[rgb]{0.87,0.95,0.87} 0.569 $\pm$ 0.035} & {\small \cellcolor[rgb]{0.80,0.92,0.85} 0.639 $\pm$ 0.044} & {\small \cellcolor[rgb]{0.83,0.93,0.85} 0.608 $\pm$ 0.027} & {\small \cellcolor[rgb]{0.80,0.92,0.85} \textbf{0.639 $\pm$ 0.045}} & {\small \cellcolor[rgb]{0.83,0.93,0.85} \textbf{0.608 $\pm$ 0.030}} \\
 & Qwen3-VL-32B & {\small \cellcolor[rgb]{0.91,0.96,0.89} 0.521 $\pm$ 0.021} & {\small \cellcolor[rgb]{0.84,0.94,0.85} 0.596 $\pm$ 0.032} & {\small \cellcolor[rgb]{0.91,0.96,0.89} 0.521 $\pm$ 0.021} & {\small \cellcolor[rgb]{0.84,0.94,0.85} 0.596 $\pm$ 0.032} & {\small \cellcolor[rgb]{0.91,0.97,0.90} 0.509 $\pm$ 0.014} & {\small \cellcolor[rgb]{0.84,0.94,0.85} 0.595 $\pm$ 0.033} & {\small \cellcolor[rgb]{0.92,0.97,0.90} 0.506 $\pm$ 0.018} & {\small \cellcolor[rgb]{0.91,0.96,0.89} 0.522 $\pm$ 0.041} & {\small \cellcolor[rgb]{0.83,0.93,0.85} \textbf{0.610 $\pm$ 0.027}} & {\small \cellcolor[rgb]{0.78,0.91,0.85} \textbf{0.651 $\pm$ 0.024}} & {\small \cellcolor[rgb]{0.83,0.93,0.85} 0.610 $\pm$ 0.027} & {\small \cellcolor[rgb]{0.78,0.91,0.85} 0.651 $\pm$ 0.023} \\
 & InternVL3-2B & {\small \cellcolor[rgb]{0.73,0.89,0.87} 0.693 $\pm$ 0.061} & {\small \cellcolor[rgb]{0.84,0.94,0.85} 0.599 $\pm$ 0.010} & {\small \cellcolor[rgb]{0.73,0.89,0.87} 0.693 $\pm$ 0.061} & {\small \cellcolor[rgb]{0.84,0.94,0.85} 0.599 $\pm$ 0.010} & {\small \cellcolor[rgb]{0.73,0.89,0.87} 0.691 $\pm$ 0.059} & {\small \cellcolor[rgb]{0.84,0.94,0.85} 0.599 $\pm$ 0.011} & {\small \cellcolor[rgb]{0.73,0.89,0.87} 0.692 $\pm$ 0.048} & {\small \cellcolor[rgb]{0.86,0.95,0.86} 0.577 $\pm$ 0.014} & {\small \cellcolor[rgb]{0.71,0.89,0.87} 0.702 $\pm$ 0.049} & {\small \cellcolor[rgb]{0.78,0.91,0.85} 0.649 $\pm$ 0.024} & {\small \cellcolor[rgb]{0.71,0.89,0.88} \textbf{0.705 $\pm$ 0.052}} & {\small \cellcolor[rgb]{0.77,0.91,0.85} \textbf{0.658 $\pm$ 0.039}} \\
 & InternVL3-8B & {\small \cellcolor[rgb]{0.76,0.91,0.86} 0.663 $\pm$ 0.099} & {\small \cellcolor[rgb]{0.91,0.96,0.89} 0.522 $\pm$ 0.026} & {\small \cellcolor[rgb]{0.76,0.91,0.86} 0.663 $\pm$ 0.099} & {\small \cellcolor[rgb]{0.91,0.96,0.89} 0.522 $\pm$ 0.026} & {\small \cellcolor[rgb]{0.77,0.91,0.86} 0.660 $\pm$ 0.102} & {\small \cellcolor[rgb]{0.91,0.96,0.89} 0.522 $\pm$ 0.026} & {\small \cellcolor[rgb]{0.80,0.92,0.84} 0.632 $\pm$ 0.075} & {\small \cellcolor[rgb]{0.92,0.97,0.91} 0.502 $\pm$ 0.036} & {\small \cellcolor[rgb]{0.74,0.90,0.86} 0.677 $\pm$ 0.088} & {\small \cellcolor[rgb]{0.89,0.96,0.88} \textbf{0.545 $\pm$ 0.036}} & {\small \cellcolor[rgb]{0.74,0.90,0.86} \textbf{0.678 $\pm$ 0.088}} & {\small \cellcolor[rgb]{0.89,0.96,0.88} \textbf{0.545 $\pm$ 0.036}} \\
 & InternVL3-38B & {\small \cellcolor[rgb]{0.77,0.91,0.86} \textbf{0.662 $\pm$ 0.035}} & {\small \cellcolor[rgb]{0.81,0.93,0.84} 0.626 $\pm$ 0.045} & {\small \cellcolor[rgb]{0.77,0.91,0.86} \textbf{0.662 $\pm$ 0.035}} & {\small \cellcolor[rgb]{0.81,0.93,0.84} 0.626 $\pm$ 0.045} & {\small \cellcolor[rgb]{0.77,0.91,0.85} 0.658 $\pm$ 0.040} & {\small \cellcolor[rgb]{0.81,0.93,0.84} 0.626 $\pm$ 0.045} & {\small \cellcolor[rgb]{0.80,0.92,0.85} 0.639 $\pm$ 0.040} & {\small \cellcolor[rgb]{0.84,0.94,0.85} 0.598 $\pm$ 0.055} & {\small \cellcolor[rgb]{0.77,0.91,0.86} 0.661 $\pm$ 0.033} & {\small \cellcolor[rgb]{0.77,0.91,0.85} \textbf{0.657 $\pm$ 0.037}} & {\small \cellcolor[rgb]{0.77,0.91,0.85} 0.660 $\pm$ 0.034} & {\small \cellcolor[rgb]{0.77,0.91,0.85} \textbf{0.657 $\pm$ 0.037}} \\
 & LLaVA-NeXT-7B & {\small \cellcolor[rgb]{0.84,0.94,0.85} \textbf{0.598 $\pm$ 0.051}} & {\small \cellcolor[rgb]{0.88,0.95,0.87} 0.558 $\pm$ 0.043} & {\small \cellcolor[rgb]{0.84,0.94,0.85} \textbf{0.598 $\pm$ 0.051}} & {\small \cellcolor[rgb]{0.88,0.95,0.87} 0.558 $\pm$ 0.043} & {\small \cellcolor[rgb]{0.85,0.94,0.86} 0.587 $\pm$ 0.035} & {\small \cellcolor[rgb]{0.88,0.95,0.87} \textbf{0.562 $\pm$ 0.041}} & {\small \cellcolor[rgb]{0.85,0.94,0.86} 0.591 $\pm$ 0.018} & {\small \cellcolor[rgb]{0.90,0.96,0.88} 0.541 $\pm$ 0.059} & {\small \cellcolor[rgb]{0.84,0.94,0.85} 0.597 $\pm$ 0.050} & {\small \cellcolor[rgb]{0.89,0.96,0.88} 0.547 $\pm$ 0.036} & {\small \cellcolor[rgb]{0.84,0.94,0.85} \textbf{0.598 $\pm$ 0.051}} & {\small \cellcolor[rgb]{0.88,0.95,0.87} 0.556 $\pm$ 0.051} \\
 & LLaVA-NeXT-34B & {\small \cellcolor[rgb]{0.82,0.93,0.84} 0.618 $\pm$ 0.037} & {\small \cellcolor[rgb]{0.86,0.95,0.86} \textbf{0.580 $\pm$ 0.038}} & {\small \cellcolor[rgb]{0.82,0.93,0.84} 0.618 $\pm$ 0.037} & {\small \cellcolor[rgb]{0.86,0.95,0.86} \textbf{0.580 $\pm$ 0.038}} & {\small \cellcolor[rgb]{0.82,0.93,0.84} 0.617 $\pm$ 0.041} & {\small \cellcolor[rgb]{0.86,0.95,0.86} 0.579 $\pm$ 0.038} & {\small \cellcolor[rgb]{0.83,0.93,0.85} 0.608 $\pm$ 0.057} & {\small \cellcolor[rgb]{0.86,0.95,0.86} 0.576 $\pm$ 0.031} & {\small \cellcolor[rgb]{0.82,0.93,0.84} \textbf{0.620 $\pm$ 0.040}} & {\small \cellcolor[rgb]{0.87,0.95,0.86} 0.574 $\pm$ 0.028} & {\small \cellcolor[rgb]{0.82,0.93,0.84} 0.619 $\pm$ 0.038} & {\small \cellcolor[rgb]{0.87,0.95,0.86} 0.574 $\pm$ 0.028} \\
\midrule
\multirow{8}{*}{\rotatebox[origin=c]{90}{Open}} & Qwen3-VL-2B & {\small \cellcolor[rgb]{0.72,0.89,0.87} 0.695 $\pm$ 0.022} & {\small \cellcolor[rgb]{0.85,0.94,0.86} 0.590 $\pm$ 0.013} & {\small \cellcolor[rgb]{0.72,0.89,0.87} 0.695 $\pm$ 0.022} & {\small \cellcolor[rgb]{0.85,0.94,0.86} 0.590 $\pm$ 0.013} & {\small \cellcolor[rgb]{0.72,0.89,0.87} 0.695 $\pm$ 0.021} & {\small \cellcolor[rgb]{0.85,0.94,0.86} 0.590 $\pm$ 0.013} & {\small \cellcolor[rgb]{0.73,0.90,0.87} 0.687 $\pm$ 0.024} & {\small \cellcolor[rgb]{0.88,0.95,0.87} 0.558 $\pm$ 0.006} & {\small \cellcolor[rgb]{0.68,0.87,0.89} 0.729 $\pm$ 0.032} & {\small \cellcolor[rgb]{0.71,0.89,0.87} 0.703 $\pm$ 0.023} & {\small \cellcolor[rgb]{0.68,0.87,0.89} \textbf{0.729 $\pm$ 0.032}} & {\small \cellcolor[rgb]{0.71,0.89,0.87} \textbf{0.704 $\pm$ 0.022}} \\
 & Qwen3-VL-8B & {\small \cellcolor[rgb]{0.79,0.92,0.85} 0.643 $\pm$ 0.023} & {\small \cellcolor[rgb]{0.70,0.88,0.88} 0.710 $\pm$ 0.014} & {\small \cellcolor[rgb]{0.79,0.92,0.85} 0.643 $\pm$ 0.023} & {\small \cellcolor[rgb]{0.70,0.88,0.88} 0.710 $\pm$ 0.014} & {\small \cellcolor[rgb]{0.79,0.92,0.85} 0.642 $\pm$ 0.023} & {\small \cellcolor[rgb]{0.70,0.88,0.88} 0.710 $\pm$ 0.013} & {\small \cellcolor[rgb]{0.81,0.93,0.84} 0.628 $\pm$ 0.019} & {\small \cellcolor[rgb]{0.75,0.90,0.86} 0.674 $\pm$ 0.028} & {\small \cellcolor[rgb]{0.65,0.85,0.89} 0.750 $\pm$ 0.010} & {\small \cellcolor[rgb]{0.62,0.84,0.90} \textbf{0.771 $\pm$ 0.012}} & {\small \cellcolor[rgb]{0.65,0.85,0.89} \textbf{0.750 $\pm$ 0.009}} & {\small \cellcolor[rgb]{0.66,0.86,0.89} 0.743 $\pm$ 0.010} \\
 & Qwen3-VL-32B & {\small \cellcolor[rgb]{0.79,0.92,0.85} 0.641 $\pm$ 0.018} & {\small \cellcolor[rgb]{0.71,0.89,0.87} 0.702 $\pm$ 0.020} & {\small \cellcolor[rgb]{0.79,0.92,0.85} 0.641 $\pm$ 0.018} & {\small \cellcolor[rgb]{0.71,0.89,0.87} 0.702 $\pm$ 0.020} & {\small \cellcolor[rgb]{0.79,0.92,0.85} 0.641 $\pm$ 0.017} & {\small \cellcolor[rgb]{0.72,0.89,0.87} 0.698 $\pm$ 0.022} & {\small \cellcolor[rgb]{0.80,0.92,0.84} 0.634 $\pm$ 0.015} & {\small \cellcolor[rgb]{0.77,0.91,0.85} 0.656 $\pm$ 0.013} & {\small \cellcolor[rgb]{0.65,0.85,0.90} 0.753 $\pm$ 0.035} & {\small \cellcolor[rgb]{0.63,0.84,0.90} \textbf{0.764 $\pm$ 0.015}} & {\small \cellcolor[rgb]{0.65,0.85,0.90} \textbf{0.753 $\pm$ 0.035}} & {\small \cellcolor[rgb]{0.64,0.85,0.90} 0.758 $\pm$ 0.008} \\
 & InternVL3-2B & {\small \cellcolor[rgb]{0.61,0.83,0.90} 0.782 $\pm$ 0.048} & {\small \cellcolor[rgb]{0.89,0.96,0.88} 0.543 $\pm$ 0.021} & {\small \cellcolor[rgb]{0.61,0.83,0.90} 0.782 $\pm$ 0.048} & {\small \cellcolor[rgb]{0.89,0.96,0.88} 0.543 $\pm$ 0.021} & {\small \cellcolor[rgb]{0.61,0.83,0.90} 0.781 $\pm$ 0.048} & {\small \cellcolor[rgb]{0.88,0.95,0.87} 0.560 $\pm$ 0.020} & {\small \cellcolor[rgb]{0.62,0.83,0.90} 0.777 $\pm$ 0.052} & {\small \cellcolor[rgb]{0.88,0.95,0.87} 0.559 $\pm$ 0.020} & {\small \cellcolor[rgb]{0.60,0.81,0.89} 0.796 $\pm$ 0.045} & {\small \cellcolor[rgb]{0.67,0.87,0.89} \textbf{0.731 $\pm$ 0.045}} & {\small \cellcolor[rgb]{0.59,0.81,0.89} \textbf{0.799 $\pm$ 0.044}} & {\small \cellcolor[rgb]{0.67,0.87,0.89} 0.731 $\pm$ 0.044} \\
 & InternVL3-8B & {\small \cellcolor[rgb]{0.63,0.84,0.90} 0.764 $\pm$ 0.032} & {\small \cellcolor[rgb]{0.82,0.93,0.85} 0.614 $\pm$ 0.031} & {\small \cellcolor[rgb]{0.63,0.84,0.90} 0.764 $\pm$ 0.032} & {\small \cellcolor[rgb]{0.82,0.93,0.85} 0.614 $\pm$ 0.031} & {\small \cellcolor[rgb]{0.63,0.85,0.90} 0.763 $\pm$ 0.032} & {\small \cellcolor[rgb]{0.82,0.93,0.84} 0.620 $\pm$ 0.033} & {\small \cellcolor[rgb]{0.65,0.86,0.89} 0.749 $\pm$ 0.036} & {\small \cellcolor[rgb]{0.82,0.93,0.84} 0.619 $\pm$ 0.034} & {\small \cellcolor[rgb]{0.60,0.82,0.90} 0.789 $\pm$ 0.031} & {\small \cellcolor[rgb]{0.66,0.86,0.89} \textbf{0.742 $\pm$ 0.038}} & {\small \cellcolor[rgb]{0.60,0.82,0.89} \textbf{0.793 $\pm$ 0.032}} & {\small \cellcolor[rgb]{0.67,0.86,0.89} 0.736 $\pm$ 0.037} \\
 & InternVL3-38B & {\small \cellcolor[rgb]{0.66,0.86,0.89} 0.740 $\pm$ 0.036} & {\small \cellcolor[rgb]{0.77,0.91,0.85} 0.656 $\pm$ 0.028} & {\small \cellcolor[rgb]{0.66,0.86,0.89} 0.740 $\pm$ 0.036} & {\small \cellcolor[rgb]{0.77,0.91,0.85} 0.656 $\pm$ 0.028} & {\small \cellcolor[rgb]{0.67,0.86,0.89} 0.736 $\pm$ 0.036} & {\small \cellcolor[rgb]{0.76,0.91,0.86} 0.668 $\pm$ 0.023} & {\small \cellcolor[rgb]{0.68,0.87,0.88} 0.728 $\pm$ 0.043} & {\small \cellcolor[rgb]{0.76,0.91,0.86} 0.666 $\pm$ 0.017} & {\small \cellcolor[rgb]{0.64,0.85,0.90} 0.756 $\pm$ 0.027} & {\small \cellcolor[rgb]{0.71,0.89,0.87} \textbf{0.701 $\pm$ 0.044}} & {\small \cellcolor[rgb]{0.64,0.85,0.90} \textbf{0.757 $\pm$ 0.022}} & {\small \cellcolor[rgb]{0.73,0.89,0.87} 0.692 $\pm$ 0.043} \\
 & LLaVA-NeXT-7B & {\small \cellcolor[rgb]{0.71,0.89,0.88} 0.706 $\pm$ 0.035} & {\small \cellcolor[rgb]{0.84,0.94,0.85} 0.603 $\pm$ 0.039} & {\small \cellcolor[rgb]{0.71,0.89,0.88} 0.706 $\pm$ 0.035} & {\small \cellcolor[rgb]{0.84,0.94,0.85} 0.603 $\pm$ 0.039} & {\small \cellcolor[rgb]{0.71,0.89,0.87} 0.703 $\pm$ 0.031} & {\small \cellcolor[rgb]{0.84,0.94,0.85} 0.599 $\pm$ 0.032} & {\small \cellcolor[rgb]{0.72,0.89,0.87} 0.694 $\pm$ 0.025} & {\small \cellcolor[rgb]{0.84,0.94,0.85} 0.601 $\pm$ 0.034} & {\small \cellcolor[rgb]{0.68,0.87,0.88} \textbf{0.725 $\pm$ 0.034}} & {\small \cellcolor[rgb]{0.71,0.89,0.87} \textbf{0.703 $\pm$ 0.022}} & {\small \cellcolor[rgb]{0.69,0.87,0.88} 0.721 $\pm$ 0.034} & {\small \cellcolor[rgb]{0.71,0.89,0.87} 0.702 $\pm$ 0.023} \\
 & LLaVA-NeXT-34B & {\small \cellcolor[rgb]{0.79,0.92,0.85} 0.645 $\pm$ 0.045} & {\small \cellcolor[rgb]{0.83,0.93,0.85} 0.613 $\pm$ 0.030} & {\small \cellcolor[rgb]{0.79,0.92,0.85} 0.645 $\pm$ 0.045} & {\small \cellcolor[rgb]{0.83,0.93,0.85} 0.613 $\pm$ 0.030} & {\small \cellcolor[rgb]{0.80,0.92,0.84} 0.633 $\pm$ 0.052} & {\small \cellcolor[rgb]{0.82,0.93,0.85} 0.613 $\pm$ 0.030} & {\small \cellcolor[rgb]{0.83,0.93,0.85} 0.611 $\pm$ 0.038} & {\small \cellcolor[rgb]{0.83,0.93,0.85} 0.611 $\pm$ 0.030} & {\small \cellcolor[rgb]{0.77,0.91,0.86} \textbf{0.660 $\pm$ 0.044}} & {\small \cellcolor[rgb]{0.68,0.87,0.88} \textbf{0.727 $\pm$ 0.058}} & {\small \cellcolor[rgb]{0.77,0.91,0.85} 0.657 $\pm$ 0.046} & {\small \cellcolor[rgb]{0.68,0.87,0.88} 0.727 $\pm$ 0.058} \\
\bottomrule
\end{tabular}
}
\end{subtable}

\vspace{0.5em}

\begin{subtable}{\textwidth}
\centering
\caption{VQA-RAD (mean $\pm$ std)}
\label{tab:hac_auroc_rad_vqa}
\resizebox{\textwidth}{!}{%
\begin{tabular}{@{}ll|cc|cc|cc|cc|cc|cc@{}}
\toprule
 &  & \multicolumn{2}{c}{Uncalibrated} & \multicolumn{2}{c}{Platt Scaling} & \multicolumn{2}{c}{Isotonic Regr.} & \multicolumn{2}{c}{Hist. Binning} & \multicolumn{2}{c}{\textbf{HAC-Platt}} & \multicolumn{2}{c}{\textbf{HAC-Gate}} \\
\cmidrule(lr){3-4} \cmidrule(lr){5-6} \cmidrule(lr){7-8} \cmidrule(lr){9-10} \cmidrule(lr){11-12} \cmidrule(lr){13-14}
 &  & Samp. & Verb. & Samp. & Verb. & Samp. & Verb. & Samp. & Verb. & Samp. & Verb. & Samp. & Verb. \\
\midrule
\multirow{8}{*}{\rotatebox[origin=c]{90}{Closed}} & Qwen3-VL-2B & {\small \cellcolor[rgb]{0.88,0.95,0.87} 0.560 $\pm$ 0.114} & {\small \cellcolor[rgb]{0.91,0.96,0.90} 0.518 $\pm$ 0.027} & {\small \cellcolor[rgb]{0.88,0.95,0.87} 0.560 $\pm$ 0.114} & {\small \cellcolor[rgb]{0.91,0.96,0.90} 0.518 $\pm$ 0.027} & {\small \cellcolor[rgb]{0.88,0.95,0.87} 0.561 $\pm$ 0.092} & {\small \cellcolor[rgb]{0.91,0.97,0.90} 0.515 $\pm$ 0.027} & {\small \cellcolor[rgb]{0.87,0.95,0.87} 0.566 $\pm$ 0.064} & {\small \cellcolor[rgb]{0.92,0.97,0.90} 0.504 $\pm$ 0.009} & {\small \cellcolor[rgb]{0.82,0.93,0.84} \textbf{0.621 $\pm$ 0.104}} & {\small \cellcolor[rgb]{0.84,0.94,0.85} 0.601 $\pm$ 0.097} & {\small \cellcolor[rgb]{0.82,0.93,0.84} 0.621 $\pm$ 0.104} & {\small \cellcolor[rgb]{0.83,0.93,0.85} \textbf{0.612 $\pm$ 0.104}} \\
 & Qwen3-VL-8B & {\small \cellcolor[rgb]{0.90,0.96,0.88} 0.537 $\pm$ 0.071} & {\small \cellcolor[rgb]{0.87,0.95,0.86} 0.575 $\pm$ 0.061} & {\small \cellcolor[rgb]{0.90,0.96,0.88} 0.537 $\pm$ 0.071} & {\small \cellcolor[rgb]{0.87,0.95,0.86} 0.575 $\pm$ 0.061} & {\small \cellcolor[rgb]{0.90,0.96,0.88} 0.539 $\pm$ 0.069} & {\small \cellcolor[rgb]{0.86,0.95,0.86} 0.580 $\pm$ 0.055} & {\small \cellcolor[rgb]{0.89,0.96,0.88} 0.546 $\pm$ 0.069} & {\small \cellcolor[rgb]{0.85,0.94,0.86} 0.591 $\pm$ 0.012} & {\small \cellcolor[rgb]{0.87,0.95,0.87} \textbf{0.566 $\pm$ 0.077}} & {\small \cellcolor[rgb]{0.88,0.95,0.87} 0.562 $\pm$ 0.059} & {\small \cellcolor[rgb]{0.90,0.96,0.88} 0.534 $\pm$ 0.061} & {\small \cellcolor[rgb]{0.85,0.94,0.86} \textbf{0.592 $\pm$ 0.068}} \\
 & Qwen3-VL-32B & {\small \cellcolor[rgb]{0.89,0.96,0.87} 0.551 $\pm$ 0.052} & {\small \cellcolor[rgb]{0.88,0.95,0.87} 0.559 $\pm$ 0.066} & {\small \cellcolor[rgb]{0.89,0.96,0.87} 0.551 $\pm$ 0.052} & {\small \cellcolor[rgb]{0.88,0.95,0.87} 0.559 $\pm$ 0.066} & {\small \cellcolor[rgb]{0.91,0.96,0.89} 0.522 $\pm$ 0.037} & {\small \cellcolor[rgb]{0.89,0.96,0.87} 0.554 $\pm$ 0.053} & {\small \cellcolor[rgb]{0.91,0.97,0.90} 0.515 $\pm$ 0.047} & {\small \cellcolor[rgb]{0.91,0.97,0.90} 0.511 $\pm$ 0.057} & {\small \cellcolor[rgb]{0.86,0.95,0.86} \textbf{0.576 $\pm$ 0.068}} & {\small \cellcolor[rgb]{0.85,0.94,0.86} \textbf{0.588 $\pm$ 0.062}} & {\small \cellcolor[rgb]{0.90,0.96,0.88} 0.541 $\pm$ 0.056} & {\small \cellcolor[rgb]{0.85,0.94,0.86} \textbf{0.588 $\pm$ 0.062}} \\
 & InternVL3-2B & {\small \cellcolor[rgb]{0.73,0.90,0.87} 0.687 $\pm$ 0.093} & {\small \cellcolor[rgb]{0.75,0.90,0.86} \textbf{0.675 $\pm$ 0.062}} & {\small \cellcolor[rgb]{0.73,0.90,0.87} 0.687 $\pm$ 0.093} & {\small \cellcolor[rgb]{0.75,0.90,0.86} \textbf{0.675 $\pm$ 0.062}} & {\small \cellcolor[rgb]{0.72,0.89,0.87} 0.697 $\pm$ 0.079} & {\small \cellcolor[rgb]{0.75,0.90,0.86} 0.670 $\pm$ 0.065} & {\small \cellcolor[rgb]{0.71,0.89,0.87} \textbf{0.702 $\pm$ 0.068}} & {\small \cellcolor[rgb]{0.79,0.92,0.85} 0.643 $\pm$ 0.074} & {\small \cellcolor[rgb]{0.74,0.90,0.86} 0.680 $\pm$ 0.100} & {\small \cellcolor[rgb]{0.76,0.91,0.86} 0.663 $\pm$ 0.053} & {\small \cellcolor[rgb]{0.74,0.90,0.86} 0.681 $\pm$ 0.101} & {\small \cellcolor[rgb]{0.77,0.91,0.86} 0.662 $\pm$ 0.071} \\
 & InternVL3-8B & {\small \cellcolor[rgb]{0.82,0.93,0.84} \textbf{0.616 $\pm$ 0.188}} & {\small \cellcolor[rgb]{0.92,0.97,0.90} 0.503 $\pm$ 0.055} & {\small \cellcolor[rgb]{0.82,0.93,0.84} \textbf{0.616 $\pm$ 0.188}} & {\small \cellcolor[rgb]{0.92,0.97,0.90} 0.503 $\pm$ 0.055} & {\small \cellcolor[rgb]{0.83,0.93,0.85} 0.610 $\pm$ 0.183} & {\small \cellcolor[rgb]{0.93,0.97,0.92} 0.476 $\pm$ 0.027} & {\small \cellcolor[rgb]{0.87,0.95,0.87} 0.570 $\pm$ 0.106} & {\small \cellcolor[rgb]{0.95,0.98,0.94} 0.449 $\pm$ 0.010} & {\small \cellcolor[rgb]{0.84,0.94,0.85} 0.595 $\pm$ 0.190} & {\small \cellcolor[rgb]{0.92,0.97,0.91} 0.499 $\pm$ 0.046} & {\small \cellcolor[rgb]{0.84,0.94,0.85} 0.600 $\pm$ 0.186} & {\small \cellcolor[rgb]{0.91,0.97,0.90} \textbf{0.512 $\pm$ 0.060}} \\
 & InternVL3-38B & {\small \cellcolor[rgb]{0.76,0.91,0.86} \textbf{0.666 $\pm$ 0.097}} & {\small \cellcolor[rgb]{0.84,0.94,0.85} 0.598 $\pm$ 0.077} & {\small \cellcolor[rgb]{0.76,0.91,0.86} \textbf{0.666 $\pm$ 0.097}} & {\small \cellcolor[rgb]{0.84,0.94,0.85} 0.598 $\pm$ 0.077} & {\small \cellcolor[rgb]{0.79,0.92,0.85} 0.645 $\pm$ 0.079} & {\small \cellcolor[rgb]{0.88,0.95,0.87} 0.562 $\pm$ 0.059} & {\small \cellcolor[rgb]{0.82,0.93,0.84} 0.617 $\pm$ 0.083} & {\small \cellcolor[rgb]{0.90,0.96,0.88} 0.539 $\pm$ 0.051} & {\small \cellcolor[rgb]{0.76,0.91,0.86} \textbf{0.666 $\pm$ 0.097}} & {\small \cellcolor[rgb]{0.84,0.94,0.85} 0.598 $\pm$ 0.077} & {\small \cellcolor[rgb]{0.76,0.91,0.86} \textbf{0.666 $\pm$ 0.097}} & {\small \cellcolor[rgb]{0.84,0.94,0.85} \textbf{0.600 $\pm$ 0.081}} \\
 & LLaVA-NeXT-7B & {\small \cellcolor[rgb]{0.82,0.93,0.84} \textbf{0.618 $\pm$ 0.062}} & {\small \cellcolor[rgb]{0.89,0.96,0.88} 0.549 $\pm$ 0.069} & {\small \cellcolor[rgb]{0.82,0.93,0.84} \textbf{0.618 $\pm$ 0.062}} & {\small \cellcolor[rgb]{0.89,0.96,0.88} 0.549 $\pm$ 0.069} & {\small \cellcolor[rgb]{0.85,0.94,0.86} 0.588 $\pm$ 0.065} & {\small \cellcolor[rgb]{0.87,0.95,0.87} \textbf{0.565 $\pm$ 0.054}} & {\small \cellcolor[rgb]{0.86,0.95,0.86} 0.576 $\pm$ 0.055} & {\small \cellcolor[rgb]{0.88,0.95,0.87} 0.561 $\pm$ 0.066} & {\small \cellcolor[rgb]{0.82,0.93,0.84} \textbf{0.618 $\pm$ 0.062}} & {\small \cellcolor[rgb]{0.93,0.97,0.92} 0.482 $\pm$ 0.047} & {\small \cellcolor[rgb]{0.82,0.93,0.84} \textbf{0.618 $\pm$ 0.062}} & {\small \cellcolor[rgb]{0.89,0.96,0.87} 0.551 $\pm$ 0.079} \\
 & LLaVA-NeXT-34B & {\small \cellcolor[rgb]{0.81,0.93,0.84} \textbf{0.625 $\pm$ 0.065}} & {\small \cellcolor[rgb]{0.89,0.96,0.88} 0.549 $\pm$ 0.034} & {\small \cellcolor[rgb]{0.81,0.93,0.84} \textbf{0.625 $\pm$ 0.065}} & {\small \cellcolor[rgb]{0.89,0.96,0.88} 0.549 $\pm$ 0.034} & {\small \cellcolor[rgb]{0.84,0.94,0.85} 0.600 $\pm$ 0.058} & {\small \cellcolor[rgb]{0.89,0.96,0.88} 0.548 $\pm$ 0.034} & {\small \cellcolor[rgb]{0.83,0.93,0.85} 0.607 $\pm$ 0.102} & {\small \cellcolor[rgb]{0.90,0.96,0.88} 0.538 $\pm$ 0.031} & {\small \cellcolor[rgb]{0.81,0.93,0.84} 0.624 $\pm$ 0.069} & {\small \cellcolor[rgb]{0.89,0.96,0.87} 0.551 $\pm$ 0.042} & {\small \cellcolor[rgb]{0.81,0.93,0.84} 0.624 $\pm$ 0.069} & {\small \cellcolor[rgb]{0.89,0.96,0.87} \textbf{0.553 $\pm$ 0.041}} \\
\midrule
\multirow{8}{*}{\rotatebox[origin=c]{90}{Open}} & Qwen3-VL-2B & {\small \cellcolor[rgb]{0.76,0.91,0.86} 0.666 $\pm$ 0.050} & {\small \cellcolor[rgb]{0.81,0.93,0.84} 0.625 $\pm$ 0.077} & {\small \cellcolor[rgb]{0.76,0.91,0.86} 0.666 $\pm$ 0.050} & {\small \cellcolor[rgb]{0.81,0.93,0.84} 0.625 $\pm$ 0.077} & {\small \cellcolor[rgb]{0.78,0.91,0.85} 0.650 $\pm$ 0.049} & {\small \cellcolor[rgb]{0.84,0.94,0.85} 0.599 $\pm$ 0.050} & {\small \cellcolor[rgb]{0.84,0.94,0.85} 0.601 $\pm$ 0.077} & {\small \cellcolor[rgb]{0.85,0.94,0.86} 0.587 $\pm$ 0.046} & {\small \cellcolor[rgb]{0.75,0.90,0.86} 0.672 $\pm$ 0.031} & {\small \cellcolor[rgb]{0.70,0.88,0.88} 0.716 $\pm$ 0.027} & {\small \cellcolor[rgb]{0.75,0.90,0.86} \textbf{0.676 $\pm$ 0.029}} & {\small \cellcolor[rgb]{0.69,0.88,0.88} \textbf{0.717 $\pm$ 0.027}} \\
 & Qwen3-VL-8B & {\small \cellcolor[rgb]{0.75,0.90,0.86} 0.675 $\pm$ 0.075} & {\small \cellcolor[rgb]{0.65,0.85,0.90} 0.753 $\pm$ 0.089} & {\small \cellcolor[rgb]{0.75,0.90,0.86} 0.675 $\pm$ 0.075} & {\small \cellcolor[rgb]{0.65,0.85,0.90} 0.753 $\pm$ 0.089} & {\small \cellcolor[rgb]{0.75,0.90,0.86} 0.676 $\pm$ 0.075} & {\small \cellcolor[rgb]{0.65,0.85,0.90} 0.753 $\pm$ 0.090} & {\small \cellcolor[rgb]{0.82,0.93,0.85} 0.614 $\pm$ 0.071} & {\small \cellcolor[rgb]{0.67,0.87,0.89} 0.735 $\pm$ 0.090} & {\small \cellcolor[rgb]{0.69,0.88,0.88} 0.717 $\pm$ 0.089} & {\small \cellcolor[rgb]{0.62,0.84,0.90} \textbf{0.774 $\pm$ 0.085}} & {\small \cellcolor[rgb]{0.69,0.88,0.88} \textbf{0.717 $\pm$ 0.090}} & {\small \cellcolor[rgb]{0.66,0.86,0.89} 0.746 $\pm$ 0.084} \\
 & Qwen3-VL-32B & {\small \cellcolor[rgb]{0.86,0.95,0.86} 0.577 $\pm$ 0.038} & {\small \cellcolor[rgb]{0.70,0.88,0.88} 0.708 $\pm$ 0.103} & {\small \cellcolor[rgb]{0.86,0.95,0.86} 0.578 $\pm$ 0.038} & {\small \cellcolor[rgb]{0.70,0.88,0.88} 0.708 $\pm$ 0.103} & {\small \cellcolor[rgb]{0.86,0.95,0.86} 0.579 $\pm$ 0.039} & {\small \cellcolor[rgb]{0.72,0.89,0.87} 0.698 $\pm$ 0.097} & {\small \cellcolor[rgb]{0.89,0.96,0.88} 0.548 $\pm$ 0.062} & {\small \cellcolor[rgb]{0.78,0.91,0.85} 0.653 $\pm$ 0.109} & {\small \cellcolor[rgb]{0.65,0.85,0.90} \textbf{0.753 $\pm$ 0.066}} & {\small \cellcolor[rgb]{0.60,0.82,0.90} 0.788 $\pm$ 0.118} & {\small \cellcolor[rgb]{0.66,0.86,0.89} 0.745 $\pm$ 0.066} & {\small \cellcolor[rgb]{0.60,0.81,0.89} \textbf{0.798 $\pm$ 0.097}} \\
 & InternVL3-2B & {\small \cellcolor[rgb]{0.68,0.87,0.88} 0.727 $\pm$ 0.079} & {\small \cellcolor[rgb]{0.91,0.97,0.90} 0.514 $\pm$ 0.110} & {\small \cellcolor[rgb]{0.68,0.87,0.88} 0.727 $\pm$ 0.079} & {\small \cellcolor[rgb]{0.91,0.97,0.90} 0.514 $\pm$ 0.110} & {\small \cellcolor[rgb]{0.71,0.89,0.87} 0.705 $\pm$ 0.082} & {\small \cellcolor[rgb]{0.94,0.98,0.93} 0.468 $\pm$ 0.055} & {\small \cellcolor[rgb]{0.76,0.91,0.86} 0.663 $\pm$ 0.051} & {\small \cellcolor[rgb]{0.96,0.98,0.94} 0.439 $\pm$ 0.085} & {\small \cellcolor[rgb]{0.59,0.81,0.89} 0.803 $\pm$ 0.047} & {\small \cellcolor[rgb]{0.59,0.80,0.89} 0.806 $\pm$ 0.095} & {\small \cellcolor[rgb]{0.58,0.79,0.88} \textbf{0.815 $\pm$ 0.043}} & {\small \cellcolor[rgb]{0.58,0.80,0.88} \textbf{0.808 $\pm$ 0.097}} \\
 & InternVL3-8B & {\small \cellcolor[rgb]{0.77,0.91,0.85} 0.654 $\pm$ 0.112} & {\small \cellcolor[rgb]{0.82,0.93,0.84} 0.616 $\pm$ 0.056} & {\small \cellcolor[rgb]{0.77,0.91,0.85} 0.654 $\pm$ 0.112} & {\small \cellcolor[rgb]{0.82,0.93,0.84} 0.616 $\pm$ 0.056} & {\small \cellcolor[rgb]{0.79,0.92,0.85} 0.640 $\pm$ 0.111} & {\small \cellcolor[rgb]{0.82,0.93,0.84} 0.617 $\pm$ 0.063} & {\small \cellcolor[rgb]{0.84,0.94,0.85} 0.598 $\pm$ 0.116} & {\small \cellcolor[rgb]{0.83,0.93,0.85} 0.604 $\pm$ 0.051} & {\small \cellcolor[rgb]{0.72,0.89,0.87} 0.699 $\pm$ 0.114} & {\small \cellcolor[rgb]{0.67,0.86,0.89} \textbf{0.739 $\pm$ 0.088}} & {\small \cellcolor[rgb]{0.71,0.89,0.87} \textbf{0.704 $\pm$ 0.107}} & {\small \cellcolor[rgb]{0.70,0.88,0.88} 0.713 $\pm$ 0.085} \\
 & InternVL3-38B & {\small \cellcolor[rgb]{0.63,0.84,0.90} 0.769 $\pm$ 0.062} & {\small \cellcolor[rgb]{0.69,0.88,0.88} 0.720 $\pm$ 0.061} & {\small \cellcolor[rgb]{0.63,0.84,0.90} 0.769 $\pm$ 0.062} & {\small \cellcolor[rgb]{0.69,0.88,0.88} 0.720 $\pm$ 0.061} & {\small \cellcolor[rgb]{0.63,0.84,0.90} 0.768 $\pm$ 0.051} & {\small \cellcolor[rgb]{0.70,0.88,0.88} 0.716 $\pm$ 0.055} & {\small \cellcolor[rgb]{0.67,0.86,0.89} 0.738 $\pm$ 0.082} & {\small \cellcolor[rgb]{0.70,0.88,0.88} 0.715 $\pm$ 0.056} & {\small \cellcolor[rgb]{0.60,0.81,0.89} \textbf{0.796 $\pm$ 0.074}} & {\small \cellcolor[rgb]{0.62,0.84,0.90} \textbf{0.771 $\pm$ 0.070}} & {\small \cellcolor[rgb]{0.60,0.81,0.89} 0.794 $\pm$ 0.069} & {\small \cellcolor[rgb]{0.68,0.87,0.88} 0.726 $\pm$ 0.060} \\
 & LLaVA-NeXT-7B & {\small \cellcolor[rgb]{0.66,0.86,0.89} 0.745 $\pm$ 0.080} & {\small \cellcolor[rgb]{0.85,0.94,0.86} 0.589 $\pm$ 0.093} & {\small \cellcolor[rgb]{0.66,0.86,0.89} 0.745 $\pm$ 0.080} & {\small \cellcolor[rgb]{0.85,0.94,0.86} 0.589 $\pm$ 0.093} & {\small \cellcolor[rgb]{0.65,0.86,0.89} 0.749 $\pm$ 0.059} & {\small \cellcolor[rgb]{0.86,0.94,0.86} 0.581 $\pm$ 0.086} & {\small \cellcolor[rgb]{0.81,0.92,0.84} 0.631 $\pm$ 0.087} & {\small \cellcolor[rgb]{0.90,0.96,0.88} 0.536 $\pm$ 0.050} & {\small \cellcolor[rgb]{0.60,0.82,0.90} \textbf{0.788 $\pm$ 0.062}} & {\small \cellcolor[rgb]{0.67,0.86,0.89} 0.737 $\pm$ 0.094} & {\small \cellcolor[rgb]{0.60,0.82,0.90} 0.788 $\pm$ 0.063} & {\small \cellcolor[rgb]{0.67,0.86,0.89} \textbf{0.737 $\pm$ 0.095}} \\
 & LLaVA-NeXT-34B & {\small \cellcolor[rgb]{0.70,0.88,0.88} 0.713 $\pm$ 0.102} & {\small \cellcolor[rgb]{0.71,0.89,0.87} 0.701 $\pm$ 0.126} & {\small \cellcolor[rgb]{0.70,0.88,0.88} 0.713 $\pm$ 0.102} & {\small \cellcolor[rgb]{0.71,0.89,0.87} 0.701 $\pm$ 0.126} & {\small \cellcolor[rgb]{0.68,0.87,0.88} \textbf{0.728 $\pm$ 0.083}} & {\small \cellcolor[rgb]{0.72,0.89,0.87} 0.699 $\pm$ 0.131} & {\small \cellcolor[rgb]{0.73,0.90,0.87} 0.690 $\pm$ 0.060} & {\small \cellcolor[rgb]{0.77,0.91,0.86} 0.662 $\pm$ 0.128} & {\small \cellcolor[rgb]{0.71,0.89,0.87} 0.704 $\pm$ 0.120} & {\small \cellcolor[rgb]{0.59,0.80,0.89} 0.807 $\pm$ 0.177} & {\small \cellcolor[rgb]{0.72,0.89,0.87} 0.696 $\pm$ 0.131} & {\small \cellcolor[rgb]{0.57,0.79,0.88} \textbf{0.819 $\pm$ 0.202}} \\
\bottomrule
\end{tabular}
}
\end{subtable}

\vspace{0.5em}

\begin{subtable}{\textwidth}
\centering
\caption{SLAKE (mean $\pm$ std)}
\label{tab:hac_auroc_slake}
\resizebox{\textwidth}{!}{%
\begin{tabular}{@{}ll|cc|cc|cc|cc|cc|cc@{}}
\toprule
 &  & \multicolumn{2}{c}{Uncalibrated} & \multicolumn{2}{c}{Platt Scaling} & \multicolumn{2}{c}{Isotonic Regr.} & \multicolumn{2}{c}{Hist. Binning} & \multicolumn{2}{c}{\textbf{HAC-Platt}} & \multicolumn{2}{c}{\textbf{HAC-Gate}} \\
\cmidrule(lr){3-4} \cmidrule(lr){5-6} \cmidrule(lr){7-8} \cmidrule(lr){9-10} \cmidrule(lr){11-12} \cmidrule(lr){13-14}
 &  & Samp. & Verb. & Samp. & Verb. & Samp. & Verb. & Samp. & Verb. & Samp. & Verb. & Samp. & Verb. \\
\midrule
\multirow{8}{*}{\rotatebox[origin=c]{90}{Closed}} & Qwen3-VL-2B & {\small \cellcolor[rgb]{0.83,0.93,0.85} 0.607 $\pm$ 0.044} & {\small \cellcolor[rgb]{0.85,0.94,0.86} 0.586 $\pm$ 0.033} & {\small \cellcolor[rgb]{0.83,0.93,0.85} 0.607 $\pm$ 0.044} & {\small \cellcolor[rgb]{0.85,0.94,0.86} 0.586 $\pm$ 0.033} & {\small \cellcolor[rgb]{0.83,0.93,0.85} 0.607 $\pm$ 0.044} & {\small \cellcolor[rgb]{0.85,0.94,0.86} 0.586 $\pm$ 0.033} & {\small \cellcolor[rgb]{0.87,0.95,0.87} 0.566 $\pm$ 0.030} & {\small \cellcolor[rgb]{0.92,0.97,0.91} 0.500 $\pm$ 0.000} & {\small \cellcolor[rgb]{0.77,0.91,0.85} \textbf{0.656 $\pm$ 0.104}} & {\small \cellcolor[rgb]{0.77,0.91,0.85} \textbf{0.657 $\pm$ 0.067}} & {\small \cellcolor[rgb]{0.77,0.91,0.85} \textbf{0.656 $\pm$ 0.104}} & {\small \cellcolor[rgb]{0.77,0.91,0.85} 0.654 $\pm$ 0.069} \\
 & Qwen3-VL-8B & {\small \cellcolor[rgb]{0.88,0.95,0.87} 0.560 $\pm$ 0.047} & {\small \cellcolor[rgb]{0.82,0.93,0.84} \textbf{0.616 $\pm$ 0.070}} & {\small \cellcolor[rgb]{0.88,0.95,0.87} 0.560 $\pm$ 0.047} & {\small \cellcolor[rgb]{0.82,0.93,0.84} \textbf{0.616 $\pm$ 0.070}} & {\small \cellcolor[rgb]{0.88,0.95,0.87} 0.559 $\pm$ 0.047} & {\small \cellcolor[rgb]{0.84,0.94,0.85} 0.603 $\pm$ 0.051} & {\small \cellcolor[rgb]{0.88,0.95,0.87} 0.557 $\pm$ 0.047} & {\small \cellcolor[rgb]{0.88,0.95,0.87} 0.558 $\pm$ 0.038} & {\small \cellcolor[rgb]{0.73,0.90,0.87} \textbf{0.686 $\pm$ 0.046}} & {\small \cellcolor[rgb]{0.83,0.93,0.85} 0.608 $\pm$ 0.058} & {\small \cellcolor[rgb]{0.73,0.90,0.87} \textbf{0.686 $\pm$ 0.046}} & {\small \cellcolor[rgb]{0.83,0.93,0.85} 0.609 $\pm$ 0.056} \\
 & Qwen3-VL-32B & {\small \cellcolor[rgb]{0.92,0.97,0.91} 0.500 $\pm$ 0.000} & {\small \cellcolor[rgb]{0.82,0.93,0.84} 0.618 $\pm$ 0.038} & {\small \cellcolor[rgb]{0.92,0.97,0.91} 0.500 $\pm$ 0.000} & {\small \cellcolor[rgb]{0.82,0.93,0.84} 0.618 $\pm$ 0.038} & {\small \cellcolor[rgb]{0.92,0.97,0.91} 0.500 $\pm$ 0.000} & {\small \cellcolor[rgb]{0.83,0.93,0.85} 0.611 $\pm$ 0.033} & {\small \cellcolor[rgb]{0.92,0.97,0.91} 0.500 $\pm$ 0.000} & {\small \cellcolor[rgb]{0.92,0.97,0.91} 0.496 $\pm$ 0.046} & {\small \cellcolor[rgb]{0.78,0.91,0.85} \textbf{0.652 $\pm$ 0.045}} & {\small \cellcolor[rgb]{0.73,0.90,0.87} \textbf{0.690 $\pm$ 0.052}} & {\small \cellcolor[rgb]{0.78,0.91,0.85} \textbf{0.652 $\pm$ 0.045}} & {\small \cellcolor[rgb]{0.73,0.90,0.87} 0.690 $\pm$ 0.052} \\
 & InternVL3-2B & {\small \cellcolor[rgb]{0.73,0.90,0.87} 0.686 $\pm$ 0.125} & {\small \cellcolor[rgb]{0.89,0.96,0.87} 0.554 $\pm$ 0.025} & {\small \cellcolor[rgb]{0.73,0.90,0.87} 0.686 $\pm$ 0.125} & {\small \cellcolor[rgb]{0.89,0.96,0.87} 0.554 $\pm$ 0.025} & {\small \cellcolor[rgb]{0.74,0.90,0.87} 0.685 $\pm$ 0.129} & {\small \cellcolor[rgb]{0.84,0.94,0.85} 0.600 $\pm$ 0.036} & {\small \cellcolor[rgb]{0.76,0.91,0.86} 0.668 $\pm$ 0.089} & {\small \cellcolor[rgb]{0.91,0.96,0.89} 0.520 $\pm$ 0.033} & {\small \cellcolor[rgb]{0.67,0.87,0.89} \textbf{0.732 $\pm$ 0.096}} & {\small \cellcolor[rgb]{0.72,0.89,0.87} \textbf{0.699 $\pm$ 0.031}} & {\small \cellcolor[rgb]{0.67,0.87,0.89} 0.732 $\pm$ 0.105} & {\small \cellcolor[rgb]{0.72,0.89,0.87} 0.696 $\pm$ 0.033} \\
 & InternVL3-8B & {\small \cellcolor[rgb]{0.75,0.90,0.86} 0.674 $\pm$ 0.102} & {\small \cellcolor[rgb]{0.90,0.96,0.88} \textbf{0.535 $\pm$ 0.056}} & {\small \cellcolor[rgb]{0.75,0.90,0.86} 0.674 $\pm$ 0.102} & {\small \cellcolor[rgb]{0.90,0.96,0.88} \textbf{0.535 $\pm$ 0.056}} & {\small \cellcolor[rgb]{0.75,0.90,0.86} 0.675 $\pm$ 0.102} & {\small \cellcolor[rgb]{0.90,0.96,0.88} 0.535 $\pm$ 0.056} & {\small \cellcolor[rgb]{0.81,0.92,0.84} 0.631 $\pm$ 0.082} & {\small \cellcolor[rgb]{0.91,0.97,0.90} 0.509 $\pm$ 0.075} & {\small \cellcolor[rgb]{0.73,0.90,0.87} 0.687 $\pm$ 0.092} & {\small \cellcolor[rgb]{0.91,0.97,0.90} 0.509 $\pm$ 0.068} & {\small \cellcolor[rgb]{0.73,0.89,0.87} \textbf{0.692 $\pm$ 0.097}} & {\small \cellcolor[rgb]{0.91,0.97,0.90} 0.509 $\pm$ 0.068} \\
 & InternVL3-38B & {\small \cellcolor[rgb]{0.72,0.89,0.87} 0.695 $\pm$ 0.057} & {\small \cellcolor[rgb]{0.80,0.92,0.84} 0.632 $\pm$ 0.050} & {\small \cellcolor[rgb]{0.72,0.89,0.87} 0.695 $\pm$ 0.057} & {\small \cellcolor[rgb]{0.80,0.92,0.84} 0.632 $\pm$ 0.050} & {\small \cellcolor[rgb]{0.74,0.90,0.87} 0.684 $\pm$ 0.060} & {\small \cellcolor[rgb]{0.80,0.92,0.85} 0.638 $\pm$ 0.049} & {\small \cellcolor[rgb]{0.78,0.91,0.85} 0.651 $\pm$ 0.049} & {\small \cellcolor[rgb]{0.83,0.93,0.85} 0.604 $\pm$ 0.057} & {\small \cellcolor[rgb]{0.73,0.90,0.87} 0.688 $\pm$ 0.056} & {\small \cellcolor[rgb]{0.74,0.90,0.86} \textbf{0.680 $\pm$ 0.036}} & {\small \cellcolor[rgb]{0.71,0.89,0.87} \textbf{0.700 $\pm$ 0.058}} & {\small \cellcolor[rgb]{0.76,0.91,0.86} 0.663 $\pm$ 0.041} \\
 & LLaVA-NeXT-7B & {\small \cellcolor[rgb]{0.84,0.94,0.85} 0.600 $\pm$ 0.075} & {\small \cellcolor[rgb]{0.87,0.95,0.87} 0.566 $\pm$ 0.095} & {\small \cellcolor[rgb]{0.84,0.94,0.85} 0.600 $\pm$ 0.075} & {\small \cellcolor[rgb]{0.87,0.95,0.87} 0.566 $\pm$ 0.095} & {\small \cellcolor[rgb]{0.86,0.95,0.86} 0.576 $\pm$ 0.056} & {\small \cellcolor[rgb]{0.87,0.95,0.87} \textbf{0.568 $\pm$ 0.097}} & {\small \cellcolor[rgb]{0.87,0.95,0.87} 0.567 $\pm$ 0.038} & {\small \cellcolor[rgb]{0.90,0.96,0.89} 0.525 $\pm$ 0.106} & {\small \cellcolor[rgb]{0.84,0.94,0.85} 0.602 $\pm$ 0.076} & {\small \cellcolor[rgb]{0.90,0.96,0.88} 0.536 $\pm$ 0.105} & {\small \cellcolor[rgb]{0.83,0.93,0.85} \textbf{0.605 $\pm$ 0.079}} & {\small \cellcolor[rgb]{0.89,0.96,0.87} 0.552 $\pm$ 0.116} \\
 & LLaVA-NeXT-34B & {\small \cellcolor[rgb]{0.84,0.94,0.85} 0.602 $\pm$ 0.044} & {\small \cellcolor[rgb]{0.83,0.93,0.85} 0.606 $\pm$ 0.060} & {\small \cellcolor[rgb]{0.84,0.94,0.85} 0.602 $\pm$ 0.044} & {\small \cellcolor[rgb]{0.83,0.93,0.85} 0.606 $\pm$ 0.060} & {\small \cellcolor[rgb]{0.83,0.93,0.85} 0.605 $\pm$ 0.044} & {\small \cellcolor[rgb]{0.82,0.93,0.85} 0.614 $\pm$ 0.053} & {\small \cellcolor[rgb]{0.84,0.94,0.85} 0.602 $\pm$ 0.042} & {\small \cellcolor[rgb]{0.82,0.93,0.84} \textbf{0.617 $\pm$ 0.043}} & {\small \cellcolor[rgb]{0.83,0.93,0.85} \textbf{0.610 $\pm$ 0.050}} & {\small \cellcolor[rgb]{0.83,0.93,0.85} 0.610 $\pm$ 0.044} & {\small \cellcolor[rgb]{0.83,0.93,0.85} 0.608 $\pm$ 0.045} & {\small \cellcolor[rgb]{0.83,0.93,0.85} 0.611 $\pm$ 0.045} \\
\midrule
\multirow{8}{*}{\rotatebox[origin=c]{90}{Open}} & Qwen3-VL-2B & {\small \cellcolor[rgb]{0.73,0.90,0.87} 0.690 $\pm$ 0.046} & {\small \cellcolor[rgb]{0.84,0.94,0.85} 0.603 $\pm$ 0.031} & {\small \cellcolor[rgb]{0.73,0.90,0.87} 0.690 $\pm$ 0.046} & {\small \cellcolor[rgb]{0.84,0.94,0.85} 0.603 $\pm$ 0.031} & {\small \cellcolor[rgb]{0.73,0.89,0.87} 0.692 $\pm$ 0.049} & {\small \cellcolor[rgb]{0.84,0.94,0.85} 0.600 $\pm$ 0.032} & {\small \cellcolor[rgb]{0.75,0.90,0.86} 0.676 $\pm$ 0.026} & {\small \cellcolor[rgb]{0.90,0.96,0.88} 0.539 $\pm$ 0.020} & {\small \cellcolor[rgb]{0.69,0.88,0.88} \textbf{0.721 $\pm$ 0.055}} & {\small \cellcolor[rgb]{0.75,0.90,0.86} 0.672 $\pm$ 0.054} & {\small \cellcolor[rgb]{0.69,0.88,0.88} 0.721 $\pm$ 0.055} & {\small \cellcolor[rgb]{0.75,0.90,0.86} \textbf{0.673 $\pm$ 0.054}} \\
 & Qwen3-VL-8B & {\small \cellcolor[rgb]{0.83,0.93,0.85} 0.609 $\pm$ 0.047} & {\small \cellcolor[rgb]{0.74,0.90,0.87} 0.685 $\pm$ 0.042} & {\small \cellcolor[rgb]{0.83,0.93,0.85} 0.609 $\pm$ 0.047} & {\small \cellcolor[rgb]{0.74,0.90,0.87} 0.685 $\pm$ 0.042} & {\small \cellcolor[rgb]{0.83,0.93,0.85} 0.608 $\pm$ 0.048} & {\small \cellcolor[rgb]{0.74,0.90,0.87} 0.685 $\pm$ 0.042} & {\small \cellcolor[rgb]{0.85,0.94,0.86} 0.591 $\pm$ 0.040} & {\small \cellcolor[rgb]{0.77,0.91,0.85} 0.656 $\pm$ 0.033} & {\small \cellcolor[rgb]{0.69,0.88,0.88} 0.720 $\pm$ 0.040} & {\small \cellcolor[rgb]{0.66,0.86,0.89} \textbf{0.742 $\pm$ 0.030}} & {\small \cellcolor[rgb]{0.69,0.87,0.88} \textbf{0.721 $\pm$ 0.039}} & {\small \cellcolor[rgb]{0.68,0.87,0.88} 0.724 $\pm$ 0.025} \\
 & Qwen3-VL-32B & {\small \cellcolor[rgb]{0.83,0.93,0.85} 0.608 $\pm$ 0.042} & {\small \cellcolor[rgb]{0.68,0.87,0.88} 0.724 $\pm$ 0.029} & {\small \cellcolor[rgb]{0.83,0.93,0.85} 0.608 $\pm$ 0.042} & {\small \cellcolor[rgb]{0.68,0.87,0.88} 0.724 $\pm$ 0.029} & {\small \cellcolor[rgb]{0.83,0.93,0.85} 0.609 $\pm$ 0.040} & {\small \cellcolor[rgb]{0.69,0.88,0.88} 0.718 $\pm$ 0.024} & {\small \cellcolor[rgb]{0.84,0.94,0.85} 0.602 $\pm$ 0.037} & {\small \cellcolor[rgb]{0.80,0.92,0.85} 0.639 $\pm$ 0.027} & {\small \cellcolor[rgb]{0.70,0.88,0.88} 0.708 $\pm$ 0.074} & {\small \cellcolor[rgb]{0.64,0.85,0.90} \textbf{0.760 $\pm$ 0.030}} & {\small \cellcolor[rgb]{0.70,0.88,0.88} \textbf{0.710 $\pm$ 0.076}} & {\small \cellcolor[rgb]{0.64,0.85,0.90} 0.754 $\pm$ 0.021} \\
 & InternVL3-2B & {\small \cellcolor[rgb]{0.67,0.87,0.89} 0.732 $\pm$ 0.058} & {\small \cellcolor[rgb]{0.82,0.93,0.85} 0.613 $\pm$ 0.037} & {\small \cellcolor[rgb]{0.67,0.87,0.89} 0.732 $\pm$ 0.058} & {\small \cellcolor[rgb]{0.82,0.93,0.85} 0.613 $\pm$ 0.037} & {\small \cellcolor[rgb]{0.68,0.87,0.88} 0.728 $\pm$ 0.057} & {\small \cellcolor[rgb]{0.82,0.93,0.84} 0.621 $\pm$ 0.036} & {\small \cellcolor[rgb]{0.71,0.89,0.88} 0.707 $\pm$ 0.054} & {\small \cellcolor[rgb]{0.82,0.93,0.85} 0.614 $\pm$ 0.027} & {\small \cellcolor[rgb]{0.66,0.86,0.89} 0.744 $\pm$ 0.064} & {\small \cellcolor[rgb]{0.74,0.90,0.87} \textbf{0.684 $\pm$ 0.016}} & {\small \cellcolor[rgb]{0.66,0.86,0.89} \textbf{0.745 $\pm$ 0.066}} & {\small \cellcolor[rgb]{0.74,0.90,0.86} 0.678 $\pm$ 0.013} \\
 & InternVL3-8B & {\small \cellcolor[rgb]{0.67,0.87,0.89} 0.733 $\pm$ 0.057} & {\small \cellcolor[rgb]{0.84,0.94,0.85} 0.598 $\pm$ 0.033} & {\small \cellcolor[rgb]{0.67,0.87,0.89} 0.733 $\pm$ 0.057} & {\small \cellcolor[rgb]{0.84,0.94,0.85} 0.598 $\pm$ 0.033} & {\small \cellcolor[rgb]{0.67,0.87,0.89} 0.734 $\pm$ 0.058} & {\small \cellcolor[rgb]{0.83,0.93,0.85} 0.609 $\pm$ 0.033} & {\small \cellcolor[rgb]{0.69,0.88,0.88} 0.721 $\pm$ 0.054} & {\small \cellcolor[rgb]{0.83,0.93,0.85} 0.609 $\pm$ 0.034} & {\small \cellcolor[rgb]{0.63,0.84,0.90} 0.763 $\pm$ 0.036} & {\small \cellcolor[rgb]{0.70,0.88,0.88} \textbf{0.714 $\pm$ 0.044}} & {\small \cellcolor[rgb]{0.63,0.84,0.90} \textbf{0.766 $\pm$ 0.037}} & {\small \cellcolor[rgb]{0.70,0.88,0.88} 0.708 $\pm$ 0.044} \\
 & InternVL3-38B & {\small \cellcolor[rgb]{0.75,0.90,0.86} 0.673 $\pm$ 0.055} & {\small \cellcolor[rgb]{0.77,0.91,0.85} 0.655 $\pm$ 0.044} & {\small \cellcolor[rgb]{0.75,0.90,0.86} 0.673 $\pm$ 0.055} & {\small \cellcolor[rgb]{0.77,0.91,0.85} 0.655 $\pm$ 0.044} & {\small \cellcolor[rgb]{0.76,0.91,0.86} 0.669 $\pm$ 0.063} & {\small \cellcolor[rgb]{0.77,0.91,0.85} 0.660 $\pm$ 0.048} & {\small \cellcolor[rgb]{0.75,0.90,0.86} 0.675 $\pm$ 0.063} & {\small \cellcolor[rgb]{0.78,0.92,0.85} 0.647 $\pm$ 0.032} & {\small \cellcolor[rgb]{0.72,0.89,0.87} 0.694 $\pm$ 0.042} & {\small \cellcolor[rgb]{0.74,0.90,0.87} \textbf{0.685 $\pm$ 0.061}} & {\small \cellcolor[rgb]{0.72,0.89,0.87} \textbf{0.699 $\pm$ 0.046}} & {\small \cellcolor[rgb]{0.76,0.91,0.86} 0.663 $\pm$ 0.046} \\
 & LLaVA-NeXT-7B & {\small \cellcolor[rgb]{0.68,0.87,0.88} 0.726 $\pm$ 0.019} & {\small \cellcolor[rgb]{0.82,0.93,0.84} 0.618 $\pm$ 0.041} & {\small \cellcolor[rgb]{0.68,0.87,0.88} 0.726 $\pm$ 0.019} & {\small \cellcolor[rgb]{0.82,0.93,0.84} 0.618 $\pm$ 0.041} & {\small \cellcolor[rgb]{0.69,0.88,0.88} 0.719 $\pm$ 0.022} & {\small \cellcolor[rgb]{0.79,0.92,0.85} 0.639 $\pm$ 0.028} & {\small \cellcolor[rgb]{0.70,0.88,0.88} 0.712 $\pm$ 0.028} & {\small \cellcolor[rgb]{0.80,0.92,0.85} 0.637 $\pm$ 0.031} & {\small \cellcolor[rgb]{0.67,0.87,0.89} 0.733 $\pm$ 0.027} & {\small \cellcolor[rgb]{0.74,0.90,0.86} 0.679 $\pm$ 0.033} & {\small \cellcolor[rgb]{0.67,0.87,0.89} \textbf{0.734 $\pm$ 0.025}} & {\small \cellcolor[rgb]{0.74,0.90,0.87} \textbf{0.681 $\pm$ 0.033}} \\
 & LLaVA-NeXT-34B & {\small \cellcolor[rgb]{0.77,0.91,0.85} \textbf{0.659 $\pm$ 0.106}} & {\small \cellcolor[rgb]{0.83,0.93,0.85} 0.608 $\pm$ 0.025} & {\small \cellcolor[rgb]{0.77,0.91,0.85} \textbf{0.659 $\pm$ 0.106}} & {\small \cellcolor[rgb]{0.83,0.93,0.85} 0.608 $\pm$ 0.025} & {\small \cellcolor[rgb]{0.78,0.91,0.85} 0.649 $\pm$ 0.091} & {\small \cellcolor[rgb]{0.83,0.93,0.85} 0.608 $\pm$ 0.026} & {\small \cellcolor[rgb]{0.84,0.94,0.85} 0.597 $\pm$ 0.047} & {\small \cellcolor[rgb]{0.83,0.93,0.85} 0.608 $\pm$ 0.026} & {\small \cellcolor[rgb]{0.77,0.91,0.85} 0.655 $\pm$ 0.107} & {\small \cellcolor[rgb]{0.73,0.90,0.87} \textbf{0.688 $\pm$ 0.031}} & {\small \cellcolor[rgb]{0.77,0.91,0.85} 0.655 $\pm$ 0.112} & {\small \cellcolor[rgb]{0.73,0.90,0.87} 0.688 $\pm$ 0.028} \\
\bottomrule
\end{tabular}
}
\end{subtable}

\vspace{0.5em}

\begin{subtable}{\textwidth}
\centering
\caption{VQA-Med (mean $\pm$ std)}
\label{tab:hac_auroc_vqa_med_2019}
\resizebox{\textwidth}{!}{%
\begin{tabular}{@{}ll|cc|cc|cc|cc|cc|cc@{}}
\toprule
 &  & \multicolumn{2}{c}{Uncalibrated} & \multicolumn{2}{c}{Platt Scaling} & \multicolumn{2}{c}{Isotonic Regr.} & \multicolumn{2}{c}{Hist. Binning} & \multicolumn{2}{c}{\textbf{HAC-Platt}} & \multicolumn{2}{c}{\textbf{HAC-Gate}} \\
\cmidrule(lr){3-4} \cmidrule(lr){5-6} \cmidrule(lr){7-8} \cmidrule(lr){9-10} \cmidrule(lr){11-12} \cmidrule(lr){13-14}
 &  & Samp. & Verb. & Samp. & Verb. & Samp. & Verb. & Samp. & Verb. & Samp. & Verb. & Samp. & Verb. \\
\midrule
\multirow{8}{*}{\rotatebox[origin=c]{90}{Closed}} & Qwen3-VL-2B & {\small \cellcolor[rgb]{0.87,0.95,0.87} \textbf{0.567 $\pm$ 0.063}} & {\small \cellcolor[rgb]{0.89,0.96,0.88} 0.549 $\pm$ 0.052} & {\small \cellcolor[rgb]{0.87,0.95,0.87} \textbf{0.567 $\pm$ 0.063}} & {\small \cellcolor[rgb]{0.89,0.96,0.88} 0.549 $\pm$ 0.052} & {\small \cellcolor[rgb]{0.87,0.95,0.87} \textbf{0.567 $\pm$ 0.063}} & {\small \cellcolor[rgb]{0.88,0.95,0.87} \textbf{0.560 $\pm$ 0.059}} & {\small \cellcolor[rgb]{0.88,0.95,0.87} 0.564 $\pm$ 0.063} & {\small \cellcolor[rgb]{0.91,0.97,0.90} 0.511 $\pm$ 0.025} & {\small \cellcolor[rgb]{0.92,0.97,0.91} 0.498 $\pm$ 0.079} & {\small \cellcolor[rgb]{0.88,0.95,0.87} 0.557 $\pm$ 0.163} & {\small \cellcolor[rgb]{0.92,0.97,0.91} 0.498 $\pm$ 0.079} & {\small \cellcolor[rgb]{0.93,0.97,0.92} 0.486 $\pm$ 0.147} \\
 & Qwen3-VL-8B & {\small \cellcolor[rgb]{0.89,0.96,0.88} 0.549 $\pm$ 0.093} & {\small \cellcolor[rgb]{0.91,0.96,0.90} 0.518 $\pm$ 0.162} & {\small \cellcolor[rgb]{0.89,0.96,0.88} 0.549 $\pm$ 0.093} & {\small \cellcolor[rgb]{0.91,0.96,0.90} 0.518 $\pm$ 0.162} & {\small \cellcolor[rgb]{0.91,0.96,0.90} 0.516 $\pm$ 0.067} & {\small \cellcolor[rgb]{0.96,0.99,0.95} 0.430 $\pm$ 0.057} & {\small \cellcolor[rgb]{0.92,0.97,0.91} 0.497 $\pm$ 0.108} & {\small \cellcolor[rgb]{0.98,0.99,0.97} 0.342 $\pm$ 0.064} & {\small \cellcolor[rgb]{0.61,0.82,0.90} \textbf{0.786 $\pm$ 0.188}} & {\small \cellcolor[rgb]{0.61,0.83,0.90} \textbf{0.780 $\pm$ 0.137}} & {\small \cellcolor[rgb]{0.62,0.84,0.90} 0.771 $\pm$ 0.169} & {\small \cellcolor[rgb]{0.62,0.84,0.90} 0.773 $\pm$ 0.144} \\
 & Qwen3-VL-32B & {\small \cellcolor[rgb]{0.92,0.97,0.91} \textbf{0.500 $\pm$ 0.000}} & {\small \cellcolor[rgb]{0.83,0.93,0.85} 0.607 $\pm$ 0.155} & {\small \cellcolor[rgb]{0.92,0.97,0.91} \textbf{0.500 $\pm$ 0.000}} & {\small \cellcolor[rgb]{0.83,0.93,0.85} 0.607 $\pm$ 0.155} & {\small \cellcolor[rgb]{0.92,0.97,0.91} \textbf{0.500 $\pm$ 0.000}} & {\small \cellcolor[rgb]{0.88,0.95,0.87} 0.561 $\pm$ 0.099} & {\small \cellcolor[rgb]{0.92,0.97,0.91} \textbf{0.500 $\pm$ 0.000}} & {\small \cellcolor[rgb]{0.98,0.99,0.97} 0.402 $\pm$ 0.072} & {\small \cellcolor[rgb]{0.97,0.99,0.95} 0.422 $\pm$ 0.077} & {\small \cellcolor[rgb]{0.78,0.92,0.85} \textbf{0.647 $\pm$ 0.205}} & {\small \cellcolor[rgb]{0.97,0.99,0.95} 0.422 $\pm$ 0.077} & {\small \cellcolor[rgb]{0.80,0.92,0.84} 0.633 $\pm$ 0.190} \\
 & InternVL3-2B & {\small \cellcolor[rgb]{0.67,0.86,0.89} 0.739 $\pm$ 0.150} & {\small \cellcolor[rgb]{0.85,0.94,0.86} \textbf{0.587 $\pm$ 0.196}} & {\small \cellcolor[rgb]{0.67,0.86,0.89} 0.739 $\pm$ 0.150} & {\small \cellcolor[rgb]{0.85,0.94,0.86} \textbf{0.587 $\pm$ 0.196}} & {\small \cellcolor[rgb]{0.64,0.85,0.90} 0.756 $\pm$ 0.093} & {\small \cellcolor[rgb]{0.91,0.96,0.89} 0.524 $\pm$ 0.160} & {\small \cellcolor[rgb]{0.59,0.81,0.89} \textbf{0.799 $\pm$ 0.041}} & {\small \cellcolor[rgb]{0.89,0.96,0.88} 0.547 $\pm$ 0.118} & {\small \cellcolor[rgb]{0.67,0.87,0.89} 0.733 $\pm$ 0.155} & {\small \cellcolor[rgb]{0.85,0.94,0.86} \textbf{0.587 $\pm$ 0.196}} & {\small \cellcolor[rgb]{0.67,0.87,0.89} 0.733 $\pm$ 0.155} & {\small \cellcolor[rgb]{0.87,0.95,0.86} 0.574 $\pm$ 0.210} \\
 & InternVL3-8B & {\small \cellcolor[rgb]{0.63,0.85,0.90} 0.762 $\pm$ 0.151} & {\small \cellcolor[rgb]{0.92,0.97,0.91} 0.491 $\pm$ 0.020} & {\small \cellcolor[rgb]{0.63,0.85,0.90} 0.762 $\pm$ 0.151} & {\small \cellcolor[rgb]{0.92,0.97,0.91} 0.491 $\pm$ 0.020} & {\small \cellcolor[rgb]{0.68,0.87,0.88} 0.725 $\pm$ 0.146} & {\small \cellcolor[rgb]{0.92,0.97,0.91} 0.500 $\pm$ 0.000} & {\small \cellcolor[rgb]{0.77,0.91,0.85} 0.653 $\pm$ 0.188} & {\small \cellcolor[rgb]{0.91,0.97,0.90} 0.509 $\pm$ 0.020} & {\small \cellcolor[rgb]{0.60,0.82,0.89} \textbf{0.793 $\pm$ 0.167}} & {\small \cellcolor[rgb]{0.79,0.92,0.85} \textbf{0.641 $\pm$ 0.197}} & {\small \cellcolor[rgb]{0.60,0.82,0.89} \textbf{0.793 $\pm$ 0.167}} & {\small \cellcolor[rgb]{0.81,0.93,0.84} 0.623 $\pm$ 0.180} \\
 & InternVL3-38B & {\small \cellcolor[rgb]{0.94,0.98,0.93} 0.463 $\pm$ 0.251} & {\small \cellcolor[rgb]{0.60,0.82,0.90} 0.791 $\pm$ 0.100} & {\small \cellcolor[rgb]{0.94,0.98,0.93} 0.463 $\pm$ 0.251} & {\small \cellcolor[rgb]{0.60,0.82,0.90} 0.791 $\pm$ 0.100} & {\small \cellcolor[rgb]{0.95,0.98,0.94} 0.444 $\pm$ 0.140} & {\small \cellcolor[rgb]{0.62,0.83,0.90} 0.777 $\pm$ 0.127} & {\small \cellcolor[rgb]{0.85,0.94,0.86} \textbf{0.587 $\pm$ 0.176}} & {\small \cellcolor[rgb]{0.67,0.86,0.89} 0.736 $\pm$ 0.214} & {\small \cellcolor[rgb]{0.97,0.99,0.95} 0.423 $\pm$ 0.172} & {\small \cellcolor[rgb]{0.57,0.79,0.88} \textbf{0.818 $\pm$ 0.140}} & {\small \cellcolor[rgb]{0.94,0.98,0.93} 0.463 $\pm$ 0.251} & {\small \cellcolor[rgb]{0.57,0.79,0.88} \textbf{0.818 $\pm$ 0.140}} \\
 & LLaVA-NeXT-7B & {\small \cellcolor[rgb]{0.88,0.95,0.87} 0.556 $\pm$ 0.184} & {\small \cellcolor[rgb]{0.87,0.95,0.87} 0.565 $\pm$ 0.164} & {\small \cellcolor[rgb]{0.88,0.95,0.87} 0.556 $\pm$ 0.184} & {\small \cellcolor[rgb]{0.87,0.95,0.87} 0.565 $\pm$ 0.164} & {\small \cellcolor[rgb]{0.88,0.95,0.87} \textbf{0.560 $\pm$ 0.153}} & {\small \cellcolor[rgb]{0.92,0.97,0.90} 0.507 $\pm$ 0.124} & {\small \cellcolor[rgb]{0.90,0.96,0.89} 0.530 $\pm$ 0.199} & {\small \cellcolor[rgb]{0.94,0.98,0.93} 0.471 $\pm$ 0.121} & {\small \cellcolor[rgb]{0.90,0.96,0.89} 0.524 $\pm$ 0.147} & {\small \cellcolor[rgb]{0.81,0.92,0.84} 0.631 $\pm$ 0.170} & {\small \cellcolor[rgb]{0.88,0.95,0.87} 0.559 $\pm$ 0.185} & {\small \cellcolor[rgb]{0.80,0.92,0.85} \textbf{0.637 $\pm$ 0.170}} \\
 & LLaVA-NeXT-34B & {\small \cellcolor[rgb]{0.87,0.95,0.87} 0.569 $\pm$ 0.173} & {\small \cellcolor[rgb]{0.90,0.96,0.89} 0.525 $\pm$ 0.034} & {\small \cellcolor[rgb]{0.87,0.95,0.87} 0.569 $\pm$ 0.173} & {\small \cellcolor[rgb]{0.90,0.96,0.89} 0.525 $\pm$ 0.034} & {\small \cellcolor[rgb]{0.93,0.97,0.92} 0.481 $\pm$ 0.154} & {\small \cellcolor[rgb]{0.90,0.96,0.88} \textbf{0.536 $\pm$ 0.033}} & {\small \cellcolor[rgb]{0.91,0.96,0.90} 0.518 $\pm$ 0.137} & {\small \cellcolor[rgb]{0.90,0.96,0.89} 0.525 $\pm$ 0.034} & {\small \cellcolor[rgb]{0.94,0.98,0.93} 0.461 $\pm$ 0.242} & {\small \cellcolor[rgb]{0.96,0.99,0.95} 0.431 $\pm$ 0.098} & {\small \cellcolor[rgb]{0.87,0.95,0.86} \textbf{0.571 $\pm$ 0.162}} & {\small \cellcolor[rgb]{0.95,0.98,0.93} 0.458 $\pm$ 0.104} \\
\midrule
\multirow{8}{*}{\rotatebox[origin=c]{90}{Open}} & Qwen3-VL-2B & {\small \cellcolor[rgb]{0.71,0.89,0.87} \textbf{0.704 $\pm$ 0.046}} & {\small \cellcolor[rgb]{0.86,0.95,0.86} 0.577 $\pm$ 0.031} & {\small \cellcolor[rgb]{0.71,0.89,0.87} \textbf{0.704 $\pm$ 0.046}} & {\small \cellcolor[rgb]{0.86,0.95,0.86} 0.577 $\pm$ 0.031} & {\small \cellcolor[rgb]{0.72,0.89,0.87} 0.693 $\pm$ 0.048} & {\small \cellcolor[rgb]{0.86,0.95,0.86} 0.577 $\pm$ 0.031} & {\small \cellcolor[rgb]{0.75,0.90,0.86} 0.671 $\pm$ 0.049} & {\small \cellcolor[rgb]{0.89,0.96,0.87} 0.554 $\pm$ 0.024} & {\small \cellcolor[rgb]{0.73,0.90,0.87} 0.688 $\pm$ 0.050} & {\small \cellcolor[rgb]{0.72,0.89,0.87} \textbf{0.696 $\pm$ 0.065}} & {\small \cellcolor[rgb]{0.73,0.90,0.87} 0.689 $\pm$ 0.048} & {\small \cellcolor[rgb]{0.72,0.89,0.87} 0.696 $\pm$ 0.065} \\
 & Qwen3-VL-8B & {\small \cellcolor[rgb]{0.77,0.91,0.85} 0.657 $\pm$ 0.026} & {\small \cellcolor[rgb]{0.67,0.86,0.89} 0.737 $\pm$ 0.055} & {\small \cellcolor[rgb]{0.77,0.91,0.85} 0.657 $\pm$ 0.026} & {\small \cellcolor[rgb]{0.67,0.86,0.89} 0.737 $\pm$ 0.055} & {\small \cellcolor[rgb]{0.77,0.91,0.85} 0.657 $\pm$ 0.026} & {\small \cellcolor[rgb]{0.67,0.86,0.89} 0.737 $\pm$ 0.055} & {\small \cellcolor[rgb]{0.78,0.91,0.85} 0.650 $\pm$ 0.030} & {\small \cellcolor[rgb]{0.75,0.90,0.86} 0.675 $\pm$ 0.044} & {\small \cellcolor[rgb]{0.64,0.85,0.90} \textbf{0.760 $\pm$ 0.053}} & {\small \cellcolor[rgb]{0.60,0.82,0.89} \textbf{0.793 $\pm$ 0.072}} & {\small \cellcolor[rgb]{0.64,0.85,0.90} 0.759 $\pm$ 0.053} & {\small \cellcolor[rgb]{0.63,0.84,0.90} 0.768 $\pm$ 0.053} \\
 & Qwen3-VL-32B & {\small \cellcolor[rgb]{0.71,0.89,0.87} 0.704 $\pm$ 0.024} & {\small \cellcolor[rgb]{0.75,0.90,0.86} 0.675 $\pm$ 0.059} & {\small \cellcolor[rgb]{0.71,0.89,0.87} 0.704 $\pm$ 0.024} & {\small \cellcolor[rgb]{0.75,0.90,0.86} 0.675 $\pm$ 0.059} & {\small \cellcolor[rgb]{0.71,0.89,0.87} 0.702 $\pm$ 0.022} & {\small \cellcolor[rgb]{0.73,0.90,0.87} 0.690 $\pm$ 0.054} & {\small \cellcolor[rgb]{0.73,0.90,0.87} 0.689 $\pm$ 0.014} & {\small \cellcolor[rgb]{0.74,0.90,0.86} 0.677 $\pm$ 0.038} & {\small \cellcolor[rgb]{0.61,0.83,0.90} 0.784 $\pm$ 0.043} & {\small \cellcolor[rgb]{0.67,0.86,0.89} 0.736 $\pm$ 0.053} & {\small \cellcolor[rgb]{0.61,0.83,0.90} \textbf{0.784 $\pm$ 0.044}} & {\small \cellcolor[rgb]{0.66,0.86,0.89} \textbf{0.741 $\pm$ 0.055}} \\
 & InternVL3-2B & {\small \cellcolor[rgb]{0.58,0.79,0.88} 0.812 $\pm$ 0.070} & {\small \cellcolor[rgb]{0.95,0.98,0.94} 0.447 $\pm$ 0.051} & {\small \cellcolor[rgb]{0.58,0.79,0.88} 0.812 $\pm$ 0.070} & {\small \cellcolor[rgb]{0.95,0.98,0.94} 0.447 $\pm$ 0.051} & {\small \cellcolor[rgb]{0.59,0.80,0.89} 0.804 $\pm$ 0.069} & {\small \cellcolor[rgb]{0.90,0.96,0.89} 0.532 $\pm$ 0.030} & {\small \cellcolor[rgb]{0.60,0.81,0.89} 0.798 $\pm$ 0.074} & {\small \cellcolor[rgb]{0.90,0.96,0.89} 0.528 $\pm$ 0.027} & {\small \cellcolor[rgb]{0.57,0.79,0.88} 0.818 $\pm$ 0.060} & {\small \cellcolor[rgb]{0.63,0.85,0.90} \textbf{0.762 $\pm$ 0.084}} & {\small \cellcolor[rgb]{0.57,0.78,0.88} \textbf{0.821 $\pm$ 0.057}} & {\small \cellcolor[rgb]{0.65,0.85,0.90} 0.753 $\pm$ 0.084} \\
 & InternVL3-8B & {\small \cellcolor[rgb]{0.59,0.81,0.89} 0.801 $\pm$ 0.061} & {\small \cellcolor[rgb]{0.80,0.92,0.84} 0.633 $\pm$ 0.029} & {\small \cellcolor[rgb]{0.59,0.81,0.89} 0.801 $\pm$ 0.061} & {\small \cellcolor[rgb]{0.80,0.92,0.84} 0.633 $\pm$ 0.029} & {\small \cellcolor[rgb]{0.59,0.81,0.89} 0.801 $\pm$ 0.059} & {\small \cellcolor[rgb]{0.80,0.92,0.84} 0.635 $\pm$ 0.028} & {\small \cellcolor[rgb]{0.61,0.82,0.90} 0.785 $\pm$ 0.063} & {\small \cellcolor[rgb]{0.80,0.92,0.84} 0.635 $\pm$ 0.028} & {\small \cellcolor[rgb]{0.57,0.78,0.88} 0.821 $\pm$ 0.060} & {\small \cellcolor[rgb]{0.64,0.85,0.90} 0.757 $\pm$ 0.052} & {\small \cellcolor[rgb]{0.57,0.78,0.87} \textbf{0.825 $\pm$ 0.063}} & {\small \cellcolor[rgb]{0.64,0.85,0.90} \textbf{0.760 $\pm$ 0.045}} \\
 & InternVL3-38B & {\small \cellcolor[rgb]{0.74,0.90,0.86} 0.680 $\pm$ 0.043} & {\small \cellcolor[rgb]{0.81,0.93,0.84} 0.629 $\pm$ 0.034} & {\small \cellcolor[rgb]{0.74,0.90,0.86} 0.680 $\pm$ 0.043} & {\small \cellcolor[rgb]{0.81,0.93,0.84} 0.629 $\pm$ 0.034} & {\small \cellcolor[rgb]{0.75,0.90,0.86} 0.676 $\pm$ 0.047} & {\small \cellcolor[rgb]{0.75,0.90,0.86} 0.677 $\pm$ 0.018} & {\small \cellcolor[rgb]{0.78,0.92,0.85} 0.647 $\pm$ 0.038} & {\small \cellcolor[rgb]{0.76,0.91,0.86} 0.668 $\pm$ 0.019} & {\small \cellcolor[rgb]{0.65,0.85,0.89} 0.751 $\pm$ 0.041} & {\small \cellcolor[rgb]{0.72,0.89,0.87} 0.696 $\pm$ 0.058} & {\small \cellcolor[rgb]{0.65,0.85,0.90} \textbf{0.752 $\pm$ 0.039}} & {\small \cellcolor[rgb]{0.69,0.88,0.88} \textbf{0.720 $\pm$ 0.073}} \\
 & LLaVA-NeXT-7B & {\small \cellcolor[rgb]{0.81,0.93,0.84} 0.625 $\pm$ 0.084} & {\small \cellcolor[rgb]{0.87,0.95,0.87} 0.565 $\pm$ 0.070} & {\small \cellcolor[rgb]{0.81,0.93,0.84} 0.625 $\pm$ 0.084} & {\small \cellcolor[rgb]{0.87,0.95,0.87} 0.565 $\pm$ 0.070} & {\small \cellcolor[rgb]{0.77,0.91,0.85} 0.654 $\pm$ 0.049} & {\small \cellcolor[rgb]{0.89,0.96,0.88} 0.550 $\pm$ 0.056} & {\small \cellcolor[rgb]{0.73,0.90,0.87} \textbf{0.688 $\pm$ 0.020}} & {\small \cellcolor[rgb]{0.87,0.95,0.87} 0.565 $\pm$ 0.043} & {\small \cellcolor[rgb]{0.78,0.92,0.85} 0.648 $\pm$ 0.097} & {\small \cellcolor[rgb]{0.75,0.90,0.86} 0.675 $\pm$ 0.050} & {\small \cellcolor[rgb]{0.81,0.93,0.84} 0.625 $\pm$ 0.085} & {\small \cellcolor[rgb]{0.74,0.90,0.86} \textbf{0.678 $\pm$ 0.051}} \\
 & LLaVA-NeXT-34B & {\small \cellcolor[rgb]{0.83,0.93,0.85} 0.607 $\pm$ 0.049} & {\small \cellcolor[rgb]{0.83,0.93,0.85} 0.606 $\pm$ 0.055} & {\small \cellcolor[rgb]{0.83,0.93,0.85} 0.607 $\pm$ 0.049} & {\small \cellcolor[rgb]{0.83,0.93,0.85} 0.606 $\pm$ 0.055} & {\small \cellcolor[rgb]{0.85,0.94,0.86} 0.588 $\pm$ 0.055} & {\small \cellcolor[rgb]{0.83,0.93,0.85} 0.606 $\pm$ 0.055} & {\small \cellcolor[rgb]{0.89,0.96,0.87} 0.553 $\pm$ 0.047} & {\small \cellcolor[rgb]{0.83,0.93,0.85} 0.606 $\pm$ 0.055} & {\small \cellcolor[rgb]{0.77,0.91,0.85} \textbf{0.659 $\pm$ 0.045}} & {\small \cellcolor[rgb]{0.69,0.88,0.88} 0.719 $\pm$ 0.085} & {\small \cellcolor[rgb]{0.80,0.92,0.84} 0.633 $\pm$ 0.051} & {\small \cellcolor[rgb]{0.68,0.87,0.88} \textbf{0.725 $\pm$ 0.092}} \\
\bottomrule
\end{tabular}
}
\end{subtable}
\end{table}

\subsection{Calibration Gains} \label{app:calibration}

Tables~\ref{tab:hac_ace_all} and \ref{tab:hac_ece_all} show the full results comparing post-hoc calibration methods: both standard post-hoc methods and the proposed HAC variants successfully addressed the miscalibration and overconfidence issues. We did not observe a single method that outperformed all others; their final calibration errors remained in a comparable range.

\vspace{1em}

\begin{table}[h]
\centering
\caption{Post-hoc calibration comparison: ACE ($\downarrow$) across calibration methods (5-fold CV). Each cell shows sampling or verbalized confidence. \textbf{Bold} = best per row.}
\label{tab:hac_ace_all}
\begin{subtable}{\textwidth}
\centering
\caption{Pooled Medical VQA Datasets (mean $\pm$ std)}
\label{tab:hac_ace_micro_avg}
\resizebox{\textwidth}{!}{%
\begin{tabular}{@{}ll|cc|cc|cc|cc|cc|cc@{}}
\toprule
 &  & \multicolumn{2}{c}{Uncalibrated} & \multicolumn{2}{c}{Platt Scaling} & \multicolumn{2}{c}{Isotonic Regr.} & \multicolumn{2}{c}{Hist. Binning} & \multicolumn{2}{c}{\textbf{HAC-Platt}} & \multicolumn{2}{c}{\textbf{HAC-Gate}} \\
\cmidrule(lr){3-4} \cmidrule(lr){5-6} \cmidrule(lr){7-8} \cmidrule(lr){9-10} \cmidrule(lr){11-12} \cmidrule(lr){13-14}
 &  & Samp. & Verb. & Samp. & Verb. & Samp. & Verb. & Samp. & Verb. & Samp. & Verb. & Samp. & Verb. \\
\midrule
\multirow{8}{*}{\rotatebox[origin=c]{90}{Closed}} & Qwen3-VL-2B & {\small \cellcolor[rgb]{0.96,0.98,0.80} 0.224 $\pm$ 0.018} & {\small \cellcolor[rgb]{0.92,0.97,0.75} 0.199 $\pm$ 0.003} & {\small \cellcolor[rgb]{0.68,0.86,0.66} 0.104 $\pm$ 0.019} & {\small \cellcolor[rgb]{0.68,0.86,0.66} 0.104 $\pm$ 0.020} & {\small \cellcolor[rgb]{0.66,0.85,0.66} \textbf{0.098 $\pm$ 0.014}} & {\small \cellcolor[rgb]{0.65,0.85,0.66} \textbf{0.095 $\pm$ 0.020}} & {\small \cellcolor[rgb]{0.70,0.87,0.67} 0.113 $\pm$ 0.014} & {\small \cellcolor[rgb]{0.66,0.85,0.66} 0.097 $\pm$ 0.019} & {\small \cellcolor[rgb]{0.70,0.87,0.67} 0.110 $\pm$ 0.032} & {\small \cellcolor[rgb]{0.73,0.89,0.67} 0.124 $\pm$ 0.025} & {\small \cellcolor[rgb]{0.70,0.87,0.67} 0.112 $\pm$ 0.034} & {\small \cellcolor[rgb]{0.71,0.87,0.67} 0.114 $\pm$ 0.019} \\
\bottomrule
\end{tabular}
}
\end{subtable}

\vspace{0.5em}

\begin{subtable}{\textwidth}
\centering
\caption{VQA-RAD (mean $\pm$ std)}
\label{tab:hac_ace_rad_vqa}
\resizebox{\textwidth}{!}{%
\begin{tabular}{@{}ll|cc|cc|cc|cc|cc|cc@{}}
\toprule
 &  & \multicolumn{2}{c}{Uncalibrated} & \multicolumn{2}{c}{Platt Scaling} & \multicolumn{2}{c}{Isotonic Regr.} & \multicolumn{2}{c}{Hist. Binning} & \multicolumn{2}{c}{\textbf{HAC-Platt}} & \multicolumn{2}{c}{\textbf{HAC-Gate}} \\
\cmidrule(lr){3-4} \cmidrule(lr){5-6} \cmidrule(lr){7-8} \cmidrule(lr){9-10} \cmidrule(lr){11-12} \cmidrule(lr){13-14}
 &  & Samp. & Verb. & Samp. & Verb. & Samp. & Verb. & Samp. & Verb. & Samp. & Verb. & Samp. & Verb. \\
\midrule
\multirow{8}{*}{\rotatebox[origin=c]{90}{Closed}} & Qwen3-VL-2B & {\small \cellcolor[rgb]{1.00,0.96,0.79} 0.281 $\pm$ 0.036} & {\small \cellcolor[rgb]{1.00,0.99,0.85} 0.257 $\pm$ 0.019} & {\small \cellcolor[rgb]{0.96,0.98,0.81} 0.228 $\pm$ 0.047} & {\small \cellcolor[rgb]{0.81,0.92,0.68} 0.150 $\pm$ 0.032} & {\small \cellcolor[rgb]{0.95,0.98,0.80} 0.223 $\pm$ 0.042} & {\small \cellcolor[rgb]{0.81,0.92,0.68} 0.154 $\pm$ 0.032} & {\small \cellcolor[rgb]{0.92,0.97,0.75} \textbf{0.201 $\pm$ 0.052}} & {\small \cellcolor[rgb]{0.81,0.92,0.68} \textbf{0.149 $\pm$ 0.023}} & {\small \cellcolor[rgb]{0.95,0.98,0.80} 0.221 $\pm$ 0.041} & {\small \cellcolor[rgb]{0.88,0.95,0.73} 0.185 $\pm$ 0.026} & {\small \cellcolor[rgb]{0.95,0.98,0.79} 0.220 $\pm$ 0.041} & {\small \cellcolor[rgb]{0.91,0.96,0.74} 0.196 $\pm$ 0.024} \\
\bottomrule
\end{tabular}
}
\end{subtable}

\vspace{0.5em}

\begin{subtable}{\textwidth}
\centering
\caption{SLAKE (mean $\pm$ std)}
\label{tab:hac_ace_slake}
\resizebox{\textwidth}{!}{%
\begin{tabular}{@{}ll|cc|cc|cc|cc|cc|cc@{}}
\toprule
 &  & \multicolumn{2}{c}{Uncalibrated} & \multicolumn{2}{c}{Platt Scaling} & \multicolumn{2}{c}{Isotonic Regr.} & \multicolumn{2}{c}{Hist. Binning} & \multicolumn{2}{c}{\textbf{HAC-Platt}} & \multicolumn{2}{c}{\textbf{HAC-Gate}} \\
\cmidrule(lr){3-4} \cmidrule(lr){5-6} \cmidrule(lr){7-8} \cmidrule(lr){9-10} \cmidrule(lr){11-12} \cmidrule(lr){13-14}
 &  & Samp. & Verb. & Samp. & Verb. & Samp. & Verb. & Samp. & Verb. & Samp. & Verb. & Samp. & Verb. \\
\midrule
\multirow{8}{*}{\rotatebox[origin=c]{90}{Closed}} & Qwen3-VL-2B & {\small \cellcolor[rgb]{0.89,0.95,0.73} 0.189 $\pm$ 0.011} & {\small \cellcolor[rgb]{0.86,0.94,0.71} 0.172 $\pm$ 0.011} & {\small \cellcolor[rgb]{0.76,0.90,0.67} \textbf{0.133 $\pm$ 0.014}} & {\small \cellcolor[rgb]{0.73,0.89,0.67} 0.125 $\pm$ 0.020} & {\small \cellcolor[rgb]{0.77,0.90,0.67} 0.137 $\pm$ 0.012} & {\small \cellcolor[rgb]{0.73,0.89,0.67} 0.124 $\pm$ 0.019} & {\small \cellcolor[rgb]{0.79,0.91,0.68} 0.146 $\pm$ 0.026} & {\small \cellcolor[rgb]{0.71,0.88,0.67} \textbf{0.117 $\pm$ 0.027}} & {\small \cellcolor[rgb]{0.77,0.90,0.67} 0.137 $\pm$ 0.031} & {\small \cellcolor[rgb]{0.75,0.89,0.67} 0.131 $\pm$ 0.018} & {\small \cellcolor[rgb]{0.78,0.91,0.68} 0.139 $\pm$ 0.031} & {\small \cellcolor[rgb]{0.77,0.90,0.67} 0.138 $\pm$ 0.013} \\
\bottomrule
\end{tabular}
}
\end{subtable}

\vspace{0.5em}

\begin{subtable}{\textwidth}
\centering
\caption{VQA-Med (mean $\pm$ std)}
\label{tab:hac_ace_vqa_med_2019}
\resizebox{\textwidth}{!}{%
\begin{tabular}{@{}ll|cc|cc|cc|cc|cc|cc@{}}
\toprule
 &  & \multicolumn{2}{c}{Uncalibrated} & \multicolumn{2}{c}{Platt Scaling} & \multicolumn{2}{c}{Isotonic Regr.} & \multicolumn{2}{c}{Hist. Binning} & \multicolumn{2}{c}{\textbf{HAC-Platt}} & \multicolumn{2}{c}{\textbf{HAC-Gate}} \\
\cmidrule(lr){3-4} \cmidrule(lr){5-6} \cmidrule(lr){7-8} \cmidrule(lr){9-10} \cmidrule(lr){11-12} \cmidrule(lr){13-14}
 &  & Samp. & Verb. & Samp. & Verb. & Samp. & Verb. & Samp. & Verb. & Samp. & Verb. & Samp. & Verb. \\
\midrule
\multirow{8}{*}{\rotatebox[origin=c]{90}{Closed}} & Qwen3-VL-2B & {\small \cellcolor[rgb]{1.00,0.95,0.78} 0.286 $\pm$ 0.035} & {\small \cellcolor[rgb]{1.00,0.88,0.70} 0.325 $\pm$ 0.035} & {\small \cellcolor[rgb]{0.98,0.99,0.83} 0.238 $\pm$ 0.033} & {\small \cellcolor[rgb]{0.99,1.00,0.85} 0.243 $\pm$ 0.056} & {\small \cellcolor[rgb]{1.00,1.00,0.86} 0.246 $\pm$ 0.025} & {\small \cellcolor[rgb]{0.97,0.99,0.82} 0.233 $\pm$ 0.064} & {\small \cellcolor[rgb]{0.98,0.99,0.83} 0.237 $\pm$ 0.034} & {\small \cellcolor[rgb]{0.98,0.99,0.84} 0.240 $\pm$ 0.059} & {\small \cellcolor[rgb]{0.85,0.94,0.71} 0.169 $\pm$ 0.045} & {\small \cellcolor[rgb]{0.98,0.99,0.83} 0.237 $\pm$ 0.034} & {\small \cellcolor[rgb]{0.81,0.92,0.68} \textbf{0.152 $\pm$ 0.024}} & {\small \cellcolor[rgb]{0.94,0.98,0.78} \textbf{0.215 $\pm$ 0.030}} \\
\bottomrule
\end{tabular}
}
\end{subtable}
\end{table}

\begin{table}[h]
\centering
\caption{Post-hoc calibration comparison: ECE ($\downarrow$) across calibration methods (5-fold CV). Each cell shows sampling\,/\,verbalized confidence. \textbf{Bold} = best per row.}
\label{tab:hac_ece_all}
\begin{subtable}{\textwidth}
\centering
\caption{Pooled Medical VQA Datasets (mean $\pm$ std)}
\label{tab:hac_ece_micro_avg}
\resizebox{\textwidth}{!}{%
\begin{tabular}{@{}ll|cc|cc|cc|cc|cc|cc@{}}
\toprule
 &  & \multicolumn{2}{c}{Uncalibrated} & \multicolumn{2}{c}{Platt Scaling} & \multicolumn{2}{c}{Isotonic Regr.} & \multicolumn{2}{c}{Hist. Binning} & \multicolumn{2}{c}{\textbf{HAC-Platt}} & \multicolumn{2}{c}{\textbf{HAC-Gate}} \\
\cmidrule(lr){3-4} \cmidrule(lr){5-6} \cmidrule(lr){7-8} \cmidrule(lr){9-10} \cmidrule(lr){11-12} \cmidrule(lr){13-14}
 &  & Samp. & Verb. & Samp. & Verb. & Samp. & Verb. & Samp. & Verb. & Samp. & Verb. & Samp. & Verb. \\
\midrule
\multirow{8}{*}{\rotatebox[origin=c]{90}{Closed}} & Qwen3-VL-2B & {\small \cellcolor[rgb]{0.98,0.99,0.83} 0.235 $\pm$ 0.024} & {\small \cellcolor[rgb]{0.92,0.97,0.75} 0.200 $\pm$ 0.004} & {\small \cellcolor[rgb]{0.53,0.79,0.63} 0.057 $\pm$ 0.022} & {\small \cellcolor[rgb]{0.45,0.68,0.57} \textbf{0.005 $\pm$ 0.002}} & {\small \cellcolor[rgb]{0.49,0.76,0.61} \textbf{0.040 $\pm$ 0.019}} & {\small \cellcolor[rgb]{0.46,0.69,0.58} 0.009 $\pm$ 0.008} & {\small \cellcolor[rgb]{0.52,0.78,0.62} 0.053 $\pm$ 0.025} & {\small \cellcolor[rgb]{0.45,0.68,0.57} 0.006 $\pm$ 0.003} & {\small \cellcolor[rgb]{0.55,0.80,0.63} 0.063 $\pm$ 0.018} & {\small \cellcolor[rgb]{0.50,0.77,0.62} 0.046 $\pm$ 0.023} & {\small \cellcolor[rgb]{0.51,0.78,0.62} 0.052 $\pm$ 0.022} & {\small \cellcolor[rgb]{0.50,0.77,0.62} 0.046 $\pm$ 0.018} \\
\bottomrule
\end{tabular}
}
\end{subtable}

\vspace{0.5em}

\begin{subtable}{\textwidth}
\centering
\caption{VQA-RAD (mean $\pm$ std)}
\label{tab:hac_ece_rad_vqa}
\resizebox{\textwidth}{!}{%
\begin{tabular}{@{}ll|cc|cc|cc|cc|cc|cc@{}}
\toprule
 &  & \multicolumn{2}{c}{Uncalibrated} & \multicolumn{2}{c}{Platt Scaling} & \multicolumn{2}{c}{Isotonic Regr.} & \multicolumn{2}{c}{Hist. Binning} & \multicolumn{2}{c}{\textbf{HAC-Platt}} & \multicolumn{2}{c}{\textbf{HAC-Gate}} \\
\cmidrule(lr){3-4} \cmidrule(lr){5-6} \cmidrule(lr){7-8} \cmidrule(lr){9-10} \cmidrule(lr){11-12} \cmidrule(lr){13-14}
 &  & Samp. & Verb. & Samp. & Verb. & Samp. & Verb. & Samp. & Verb. & Samp. & Verb. & Samp. & Verb. \\
\midrule
\multirow{8}{*}{\rotatebox[origin=c]{90}{Closed}} & Qwen3-VL-2B & {\small \cellcolor[rgb]{1.00,0.95,0.78} 0.288 $\pm$ 0.040} & {\small \cellcolor[rgb]{0.99,1.00,0.85} 0.245 $\pm$ 0.010} & {\small \cellcolor[rgb]{0.71,0.88,0.67} 0.115 $\pm$ 0.040} & {\small \cellcolor[rgb]{0.46,0.70,0.58} 0.013 $\pm$ 0.006} & {\small \cellcolor[rgb]{0.63,0.84,0.65} \textbf{0.086 $\pm$ 0.034}} & {\small \cellcolor[rgb]{0.47,0.70,0.58} 0.014 $\pm$ 0.004} & {\small \cellcolor[rgb]{0.71,0.88,0.67} 0.117 $\pm$ 0.051} & {\small \cellcolor[rgb]{0.46,0.70,0.58} \textbf{0.013 $\pm$ 0.006}} & {\small \cellcolor[rgb]{0.72,0.88,0.67} 0.118 $\pm$ 0.019} & {\small \cellcolor[rgb]{0.50,0.77,0.62} 0.047 $\pm$ 0.030} & {\small \cellcolor[rgb]{0.70,0.87,0.67} 0.110 $\pm$ 0.021} & {\small \cellcolor[rgb]{0.49,0.74,0.60} 0.032 $\pm$ 0.021} \\
\bottomrule
\end{tabular}
}
\end{subtable}

\vspace{0.5em}

\begin{subtable}{\textwidth}
\centering
\caption{SLAKE (mean $\pm$ std)}
\label{tab:hac_ece_slake}
\resizebox{\textwidth}{!}{%
\begin{tabular}{@{}ll|cc|cc|cc|cc|cc|cc@{}}
\toprule
 &  & \multicolumn{2}{c}{Uncalibrated} & \multicolumn{2}{c}{Platt Scaling} & \multicolumn{2}{c}{Isotonic Regr.} & \multicolumn{2}{c}{Hist. Binning} & \multicolumn{2}{c}{\textbf{HAC-Platt}} & \multicolumn{2}{c}{\textbf{HAC-Gate}} \\
\cmidrule(lr){3-4} \cmidrule(lr){5-6} \cmidrule(lr){7-8} \cmidrule(lr){9-10} \cmidrule(lr){11-12} \cmidrule(lr){13-14}
 &  & Samp. & Verb. & Samp. & Verb. & Samp. & Verb. & Samp. & Verb. & Samp. & Verb. & Samp. & Verb. \\
\midrule
\multirow{8}{*}{\rotatebox[origin=c]{90}{Closed}} & Qwen3-VL-2B & {\small \cellcolor[rgb]{0.91,0.96,0.75} 0.199 $\pm$ 0.017} & {\small \cellcolor[rgb]{0.81,0.92,0.68} 0.150 $\pm$ 0.006} & {\small \cellcolor[rgb]{0.50,0.77,0.62} 0.047 $\pm$ 0.008} & {\small \cellcolor[rgb]{0.47,0.71,0.59} 0.019 $\pm$ 0.009} & {\small \cellcolor[rgb]{0.49,0.76,0.61} \textbf{0.040 $\pm$ 0.017}} & {\small \cellcolor[rgb]{0.47,0.71,0.59} 0.018 $\pm$ 0.009} & {\small \cellcolor[rgb]{0.50,0.76,0.62} 0.044 $\pm$ 0.014} & {\small \cellcolor[rgb]{0.46,0.69,0.58} \textbf{0.008 $\pm$ 0.002}} & {\small \cellcolor[rgb]{0.56,0.80,0.64} 0.068 $\pm$ 0.024} & {\small \cellcolor[rgb]{0.47,0.72,0.59} 0.022 $\pm$ 0.009} & {\small \cellcolor[rgb]{0.59,0.82,0.64} 0.077 $\pm$ 0.026} & {\small \cellcolor[rgb]{0.50,0.76,0.61} 0.042 $\pm$ 0.034} \\
\bottomrule
\end{tabular}
}
\end{subtable}

\vspace{0.5em}

\begin{subtable}{\textwidth}
\centering
\caption{VQA-Med (mean $\pm$ std)}
\label{tab:hac_ece_vqa_med_2019}
\resizebox{\textwidth}{!}{%
\begin{tabular}{@{}ll|cc|cc|cc|cc|cc|cc@{}}
\toprule
 &  & \multicolumn{2}{c}{Uncalibrated} & \multicolumn{2}{c}{Platt Scaling} & \multicolumn{2}{c}{Isotonic Regr.} & \multicolumn{2}{c}{Hist. Binning} & \multicolumn{2}{c}{\textbf{HAC-Platt}} & \multicolumn{2}{c}{\textbf{HAC-Gate}} \\
\cmidrule(lr){3-4} \cmidrule(lr){5-6} \cmidrule(lr){7-8} \cmidrule(lr){9-10} \cmidrule(lr){11-12} \cmidrule(lr){13-14}
 &  & Samp. & Verb. & Samp. & Verb. & Samp. & Verb. & Samp. & Verb. & Samp. & Verb. & Samp. & Verb. \\
\midrule
\multirow{8}{*}{\rotatebox[origin=c]{90}{Closed}} & Qwen3-VL-2B & {\small \cellcolor[rgb]{1.00,0.95,0.78} 0.286 $\pm$ 0.035} & {\small \cellcolor[rgb]{1.00,0.94,0.76} 0.297 $\pm$ 0.040} & {\small \cellcolor[rgb]{0.54,0.80,0.63} 0.061 $\pm$ 0.020} & {\small \cellcolor[rgb]{0.54,0.80,0.63} 0.062 $\pm$ 0.034} & {\small \cellcolor[rgb]{0.50,0.78,0.62} \textbf{0.051 $\pm$ 0.023}} & {\small \cellcolor[rgb]{0.50,0.76,0.61} 0.042 $\pm$ 0.041} & {\small \cellcolor[rgb]{0.54,0.80,0.63} 0.062 $\pm$ 0.020} & {\small \cellcolor[rgb]{0.49,0.75,0.61} \textbf{0.037 $\pm$ 0.014}} & {\small \cellcolor[rgb]{0.55,0.80,0.63} 0.064 $\pm$ 0.021} & {\small \cellcolor[rgb]{0.61,0.83,0.65} 0.083 $\pm$ 0.041} & {\small \cellcolor[rgb]{0.55,0.80,0.63} 0.065 $\pm$ 0.026} & {\small \cellcolor[rgb]{0.50,0.76,0.61} 0.042 $\pm$ 0.019} \\
\bottomrule
\end{tabular}
}
\end{subtable}
\end{table}

\clearpage

\section{Ablation and Further Analysis of HAC} \label{app:hac_ablation}

\subsection{Cross-Dataset Calibration Transfer}
\label{app:cross_dataset_transfer}

\begin{figure}[h]
\centering
\begin{subfigure}[t]{\textwidth}
\centering
\includegraphics[width=0.7\textwidth]{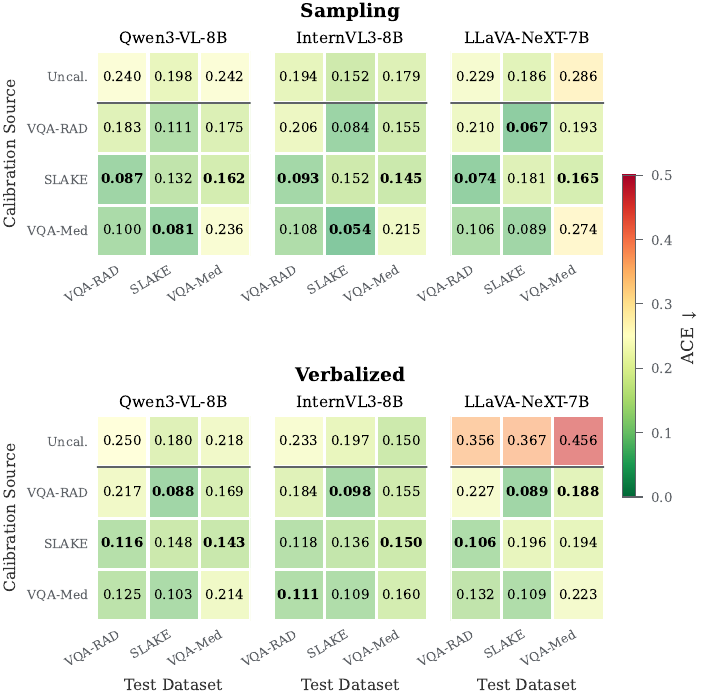}
\caption{Closed-ended}
\label{fig:transfer_ace_closed}
\end{subfigure}

\begin{subfigure}[t!]{\textwidth}
\centering
\includegraphics[width=0.7\textwidth]{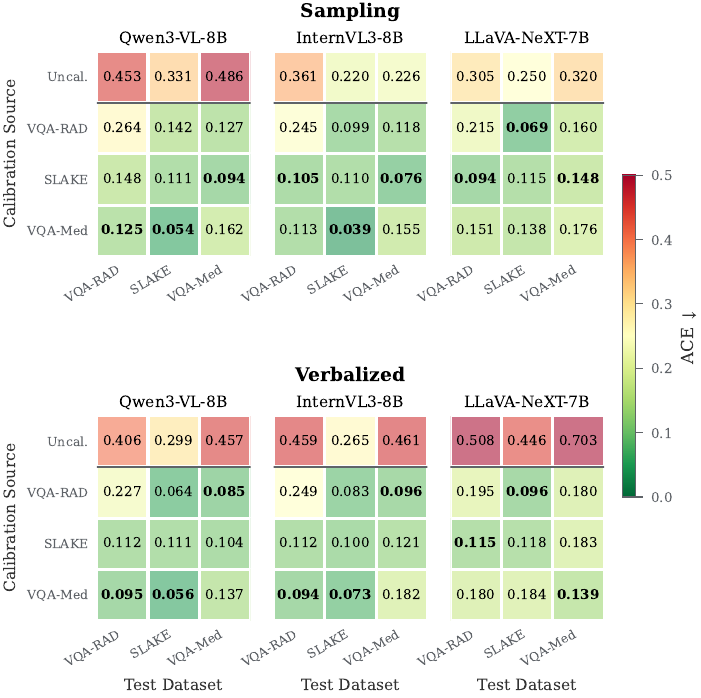}
\caption{Open-ended}
\label{fig:transfer_ace_open}
\end{subfigure}
\caption{Cross-dataset ACE transfer for HAC-Platt. Each cell shows the ACE when calibration is fitted on one dataset (row) and evaluated on another (column). Bold = best per column. The ``Uncal.'' row shows raw confidence without calibration.}
\label{fig:transfer_ace}
\end{figure}

\begin{figure}[t!]
\centering
\begin{subfigure}[t]{\textwidth}
\centering
\includegraphics[width=0.7\textwidth]{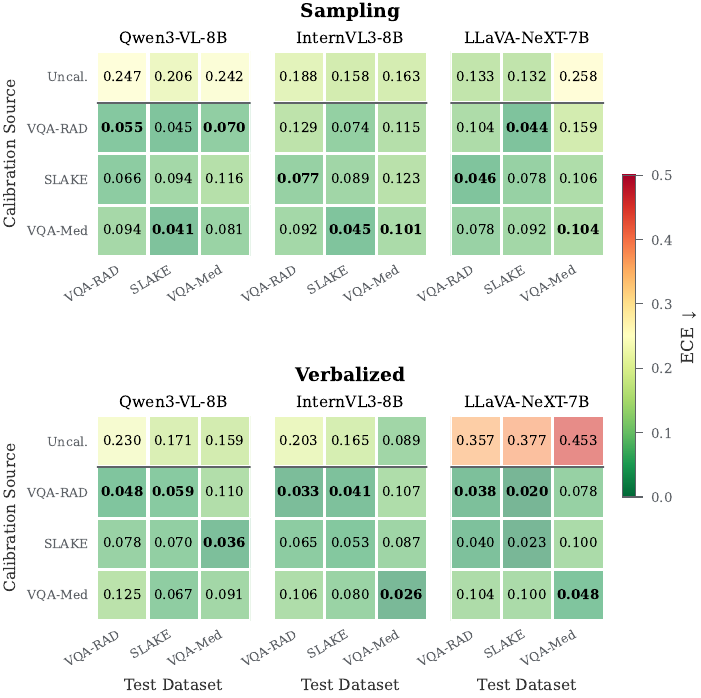}
\caption{Closed-ended}
\label{fig:transfer_ece_closed}
\end{subfigure}

\vspace{1em}

\begin{subfigure}[t]{\textwidth}
\centering
\includegraphics[width=0.7\textwidth]{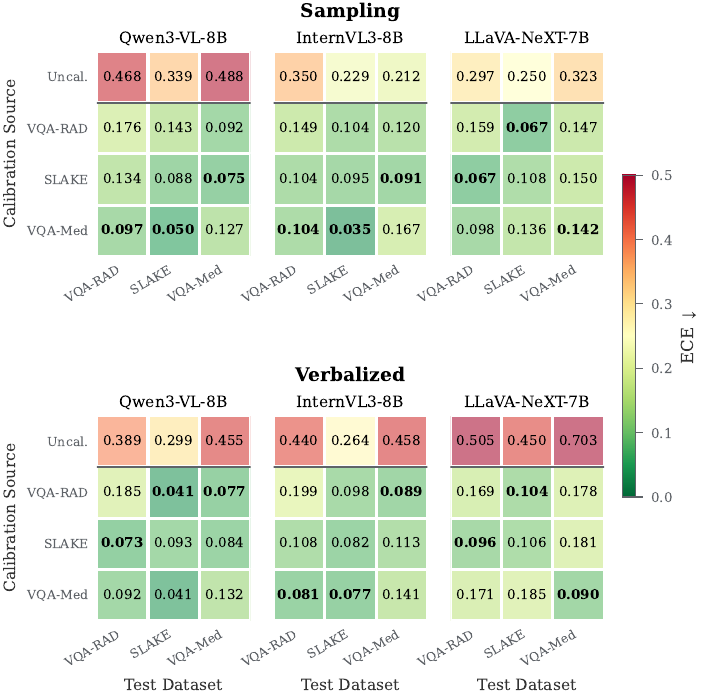}
\caption{Open-ended}
\label{fig:transfer_ece_open}
\end{subfigure}
\caption{Cross-dataset ECE transfer for HAC-Platt. Same layout as Figure~\ref{fig:transfer_ace}.}
\label{fig:transfer_ece}
\end{figure}

\begin{figure}[h]
\centering
\begin{subfigure}[t]{\textwidth}
\centering
\includegraphics[width=0.7\textwidth]{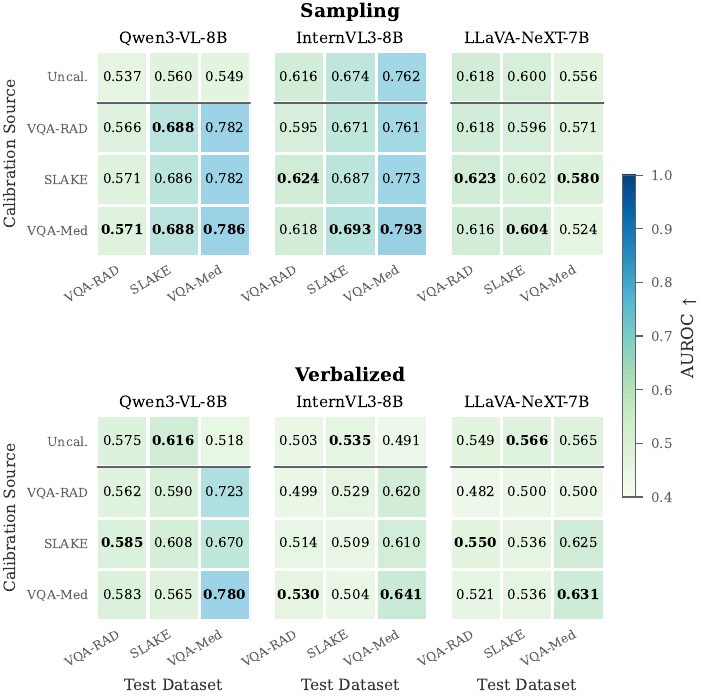}
\caption{Closed-ended}
\label{fig:transfer_auroc_closed}
\end{subfigure}

\vspace{1em}

\begin{subfigure}[t]{\textwidth}
\centering
\includegraphics[width=0.7\textwidth]{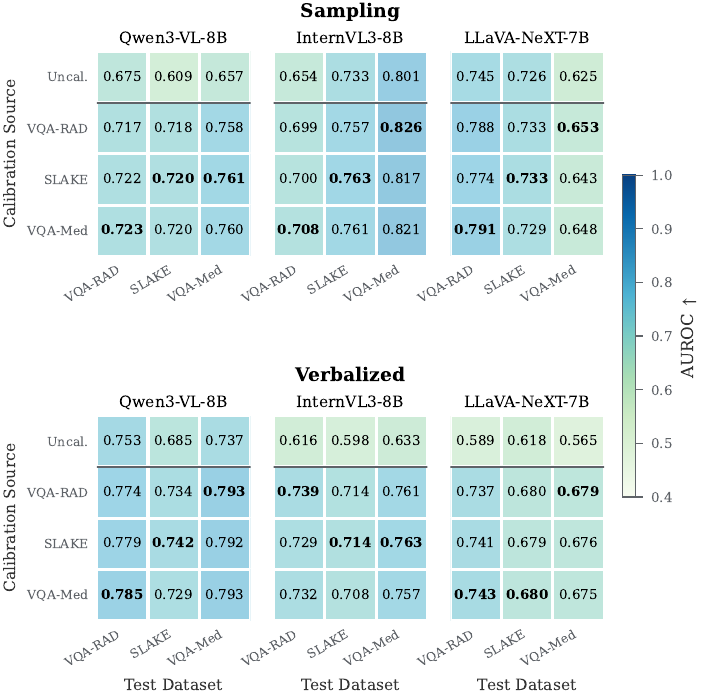}
\caption{Open-ended}
\label{fig:transfer_auroc_open}
\end{subfigure}
\caption{Cross-dataset AUROC transfer for HAC-Platt. Same layout as Figure~\ref{fig:transfer_ece}.}
\label{fig:transfer_auroc}
\end{figure}

We evaluate whether HAC-Platt calibration parameters learned on one dataset transfer to another. Specifically, we tune HAC parameters on each dataset and test on the remaining ones. Figures~\ref{fig:transfer_ace}--\ref{fig:transfer_auroc} show transfer matrices for ACE, ECE, and AUROC with both sampling-based and verbalized confidence. Overall, we observed that HAC is relatively insensitive to the choice of calibration dataset. Another notable finding is that parameters tuned on a dataset's own calibration set do not always yield the best performance, likely due to overfitting during the calibration stage. Exploring more robust calibration tuning methods remains an important direction.

\subsection{Analysis of Learned HAC-Platt Parameters}
\label{app:hac_params}
Table~\ref{tab:hac_params} reports the learned HAC-Platt parameters $\hat{a}$, $\hat{b}$, $\hat{d}$ for the scoring function $s(c,h) = \sigma(\hat{a} \cdot c + \hat{b} \cdot h + \hat{d})$, averaged across 5-fold CV on the pooled dataset. The signs $\hat{a} \geq 0$ and $\hat{b} \leq 0$ held across all 160 fits (8 models $\times$ 5 folds $\times$ 2 confidence types $\times$ 2 question types), confirming that higher base confidence consistently increases and higher hallucination scores consistently decrease the calibrated output.
For sampling-based confidence, parameters were stable across models ($\hat{a} \approx 2$--$5$, $\hat{b} \approx -0.04$ to $-1.00$), with low cross-fold variance. The hallucination penalty $\hat{b}$ was on average stronger for open-ended questions ($\bar{\hat{b}} = -0.63$) than for closed-ended ones ($\bar{\hat{b}} = -0.30$), consistent with the greater risk of hallucination with free-form answers. For verbalized confidence, $\hat{a}$ and $\hat{d}$ exhibited larger magnitudes and higher variance, particularly for closed-ended questions.

\begin{table}[h]
\centering
\caption{Learned HAC-Platt parameters $\hat{a}$, $\hat{b}$, $\hat{d}$ for $s(c,h) = \sigma(\hat{a} \cdot c + \hat{b} \cdot h + \hat{d})$, reported as mean{\tiny$\pm$}std across 5-fold CV on the pooled dataset. $\hat{a} \geq 0$ and $\hat{b} \leq 0$ are satisfied in all cases.}
\label{tab:hac_params}
\vspace{0.3em}
\small
\setlength{\tabcolsep}{4pt}
\begin{tabular}{l rrr rrr}
\toprule
& \multicolumn{3}{c}{\textbf{Sampling (Closed)}} & \multicolumn{3}{c}{\textbf{Sampling (Open)}} \\
\cmidrule(lr){2-4} \cmidrule(lr){5-7}
\textbf{Model} & $\hat{a}$ & $\hat{b}$ & $\hat{d}$ & $\hat{a}$ & $\hat{b}$ & $\hat{d}$ \\
\midrule
Qwen3-VL-2B   & 4.47{\tiny$\pm$0.48} & --0.49{\tiny$\pm$0.08} & --3.12{\tiny$\pm$0.45} & 2.51{\tiny$\pm$0.10} & --0.49{\tiny$\pm$0.04} & --1.81{\tiny$\pm$0.08} \\
Qwen3-VL-8B   & 4.62{\tiny$\pm$0.47} & --0.60{\tiny$\pm$0.06} & --3.21{\tiny$\pm$0.45} & 3.34{\tiny$\pm$0.17} & --0.94{\tiny$\pm$0.02} & --2.37{\tiny$\pm$0.17} \\
Qwen3-VL-32B  & 3.26{\tiny$\pm$0.63} & --0.48{\tiny$\pm$0.05} & --1.90{\tiny$\pm$0.62} & 3.51{\tiny$\pm$0.64} & --1.00{\tiny$\pm$0.03} & --2.44{\tiny$\pm$0.63} \\
InternVL3-2B  & 3.40{\tiny$\pm$0.42} & --0.48{\tiny$\pm$0.06} & --1.78{\tiny$\pm$0.38} & 3.70{\tiny$\pm$0.30} & --0.58{\tiny$\pm$0.07} & --2.02{\tiny$\pm$0.30} \\
InternVL3-8B  & 3.73{\tiny$\pm$0.55} & --0.13{\tiny$\pm$0.08} & --2.00{\tiny$\pm$0.48} & 2.65{\tiny$\pm$0.25} & --0.70{\tiny$\pm$0.07} & --1.15{\tiny$\pm$0.26} \\
InternVL3-38B & 3.28{\tiny$\pm$0.19} & --0.07{\tiny$\pm$0.07} & --1.42{\tiny$\pm$0.16} & 3.48{\tiny$\pm$0.22} & --0.50{\tiny$\pm$0.04} & --2.15{\tiny$\pm$0.22} \\
LLaVA-NeXT-7B  & 2.50{\tiny$\pm$0.28} & --0.04{\tiny$\pm$0.03} & --1.28{\tiny$\pm$0.20} & 2.02{\tiny$\pm$0.18} & --0.52{\tiny$\pm$0.06} & --1.12{\tiny$\pm$0.18} \\
LLaVA-NeXT-34B & 2.52{\tiny$\pm$0.23} & --0.08{\tiny$\pm$0.06} & --1.49{\tiny$\pm$0.16} & 2.13{\tiny$\pm$0.22} & --0.33{\tiny$\pm$0.02} & --1.60{\tiny$\pm$0.17} \\
\midrule
& \multicolumn{3}{c}{\textbf{Verbalized (Closed)}} & \multicolumn{3}{c}{\textbf{Verbalized (Open)}} \\
\cmidrule(lr){2-4} \cmidrule(lr){5-7}
\textbf{Model} & $\hat{a}$ & $\hat{b}$ & $\hat{d}$ & $\hat{a}$ & $\hat{b}$ & $\hat{d}$ \\
\midrule
Qwen3-VL-2B   & 7.32{\tiny$\pm$11.0} & --0.67{\tiny$\pm$0.07} & --5.71{\tiny$\pm$10.5} & 2.43{\tiny$\pm$0.20} & --0.76{\tiny$\pm$0.04} & --1.64{\tiny$\pm$0.22} \\
Qwen3-VL-8B   & 4.95{\tiny$\pm$0.98} & --0.20{\tiny$\pm$0.08} & --3.43{\tiny$\pm$0.93} & 12.9{\tiny$\pm$1.7} & --0.79{\tiny$\pm$0.03} & --11.3{\tiny$\pm$1.6} \\
Qwen3-VL-32B  & 9.55{\tiny$\pm$1.44} & --0.40{\tiny$\pm$0.06} & --7.61{\tiny$\pm$1.36} & 12.0{\tiny$\pm$1.0} & --0.87{\tiny$\pm$0.03} & --10.3{\tiny$\pm$0.9} \\
InternVL3-2B  & 7.44{\tiny$\pm$0.38} & --0.63{\tiny$\pm$0.06} & --5.52{\tiny$\pm$0.38} & 1.83{\tiny$\pm$0.35} & --0.94{\tiny$\pm$0.05} & --0.21{\tiny$\pm$0.31} \\
InternVL3-8B  & 11.1{\tiny$\pm$3.4} & --0.17{\tiny$\pm$0.08} & --9.22{\tiny$\pm$3.23} & 10.5{\tiny$\pm$1.1} & --0.84{\tiny$\pm$0.04} & --8.62{\tiny$\pm$1.04} \\
InternVL3-38B & 12.5{\tiny$\pm$1.2} & --0.23{\tiny$\pm$0.10} & --10.2{\tiny$\pm$1.2} & 12.9{\tiny$\pm$0.7} & --0.55{\tiny$\pm$0.05} & --11.1{\tiny$\pm$0.6} \\
LLaVA-NeXT-7B  & 0.25{\tiny$\pm$0.16} & --0.06{\tiny$\pm$0.04} & 0.21{\tiny$\pm$0.16} & 1.96{\tiny$\pm$0.31} & --0.69{\tiny$\pm$0.02} & --1.34{\tiny$\pm$0.32} \\
LLaVA-NeXT-34B & 2.62{\tiny$\pm$0.78} & --0.18{\tiny$\pm$0.07} & --1.59{\tiny$\pm$0.70} & 2.21{\tiny$\pm$0.39} & --0.85{\tiny$\pm$0.08} & --1.32{\tiny$\pm$0.44} \\
\bottomrule
\end{tabular}
\end{table}

\subsection{Ablation on Hallucination Metrics}
\label{app:ablation_hedge}

\begin{figure}[h]
\centering
\begin{subfigure}[t]{\textwidth}
\centering
\includegraphics[width=\textwidth]{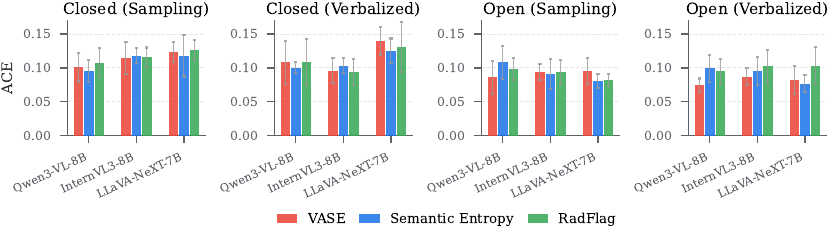}
\caption{ACE ($\downarrow$)}
\label{fig:ablation_hedge_ace}
\end{subfigure}

\vspace{0.5em}

\begin{subfigure}[t]{\textwidth}
\centering
\includegraphics[width=\textwidth]{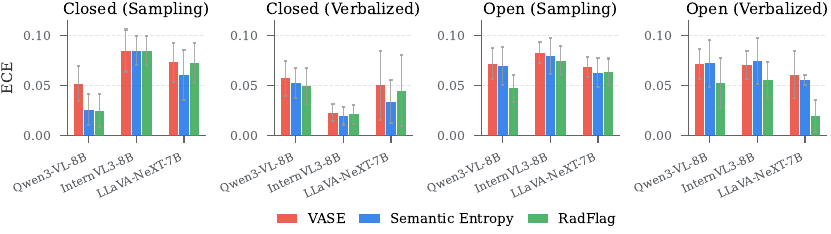}
\caption{ECE ($\downarrow$)}
\label{fig:ablation_hedge_ece}
\end{subfigure}

\vspace{0.5em}

\begin{subfigure}[t]{\textwidth}
\centering
\includegraphics[width=\textwidth]{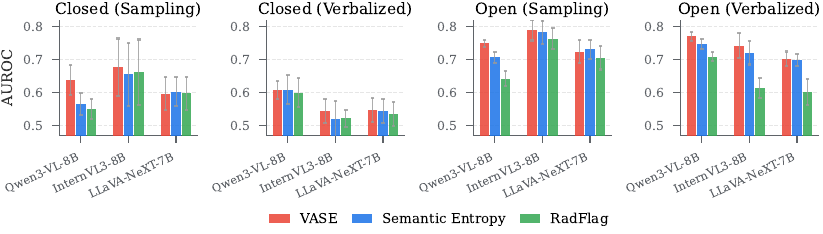}
\caption{AUROC ($\uparrow$)}
\label{fig:ablation_hedge_auroc}
\end{subfigure}
\caption{Ablation on hallucination detection metrics used in HAC-Platt. The pooled dataset is used.}
\label{fig:ablation_hedge}
\end{figure}

We compare three hallucination detection metrics as input to HAC-Platt: VASE, Semantic Entropy (SE)~\citep{farquhar2024detecting}, and RadFlag~\citep{zhang2024radflag}.
VASE consistently achieves the highest AUROC, particularly on open-ended questions. On closed-ended questions, SE and RadFlag show comparable or slightly better ACE in some cases, but their AUROC is notably weaker. These results support VASE as the default hallucination metric for HAC due to its superior discriminative ability across both question types.

\clearpage

\subsection{Auxiliary Signal Substitution for Closed-Ended Questions}
\label{app:hac_logit}
Section~\ref{sec:hac_baseline} attributes HAC's inconsistent
closed-ended gains to the weak correlation between VASE and correctness
on yes/no questions. To test whether the limitation lies in the signal
rather than the framework, we replace VASE with the model's internal
yes/no logit margin, $-|p_{\text{yes}}-p_{\text{no}}|$ (HAC-Logit),
keeping all other components fixed. Table~\ref{tab:logit-margin}
confirms this that HAC-Logit achieves the best closed-ended AUROC on all
three models, while HAC-VASE retains lower ACE on two of three,
indicating the logit margin is the stronger discriminative signal for
this format but does not uniformly dominate in calibration error.

\begin{table}[h]
\centering
\small
\caption{Auxiliary signal substitution on closed-ended questions on pooled datasets. HAC-Logit replaces VASE with the model's
internal yes/no logit margin $-|p_{\text{yes}}-p_{\text{no}}|$, keeping
all other components fixed.}
\label{tab:hac_logit_closed}
\begin{subtable}{\linewidth}
\centering
\caption{AUROC ($\uparrow$)}
\begin{tabular}{lccc}
\toprule
Method & Qwen3-VL-8B & InternVL3-8B & LLaVA-NeXT-7B \\
\midrule
Sampling   & $0.551$ & $0.663$ & $0.598$ \\
Verbalized & $0.602$ & $0.524$ & $0.560$ \\
HAC-VASE   & $0.639$ & $0.677$ & $0.597$ \\
HAC-Logit  & $\mathbf{0.789}$ & $\mathbf{0.740}$ & $\mathbf{0.607}$ \\
\bottomrule
\end{tabular}
\end{subtable}

\vspace{0.8em}

\begin{subtable}{\linewidth}
\centering
\caption{ACE ($\downarrow$)}
\begin{tabular}{lccc}
\toprule
Method & Qwen3-VL-8B & InternVL3-8B & LLaVA-NeXT-7B \\
\midrule
Sampling   & $0.214$ & $0.152$ & $0.165$ \\
Verbalized & $0.181$ & $0.177$ & $0.364$ \\
HAC-VASE   & $\mathbf{0.101}$ & $\mathbf{0.114}$ & $0.124$ \\
HAC-Logit  & $0.149$ & $0.134$ & $\mathbf{0.117}$ \\
\bottomrule
\end{tabular}
\end{subtable}
\label{tab:logit-margin}
\end{table}

\subsection{Calibration-Set Size Ablation}
\label{app:calsize}
\begin{figure}[t]
\centering
\begin{subfigure}[t]{0.68\textwidth}
\centering
\includegraphics[width=\textwidth]{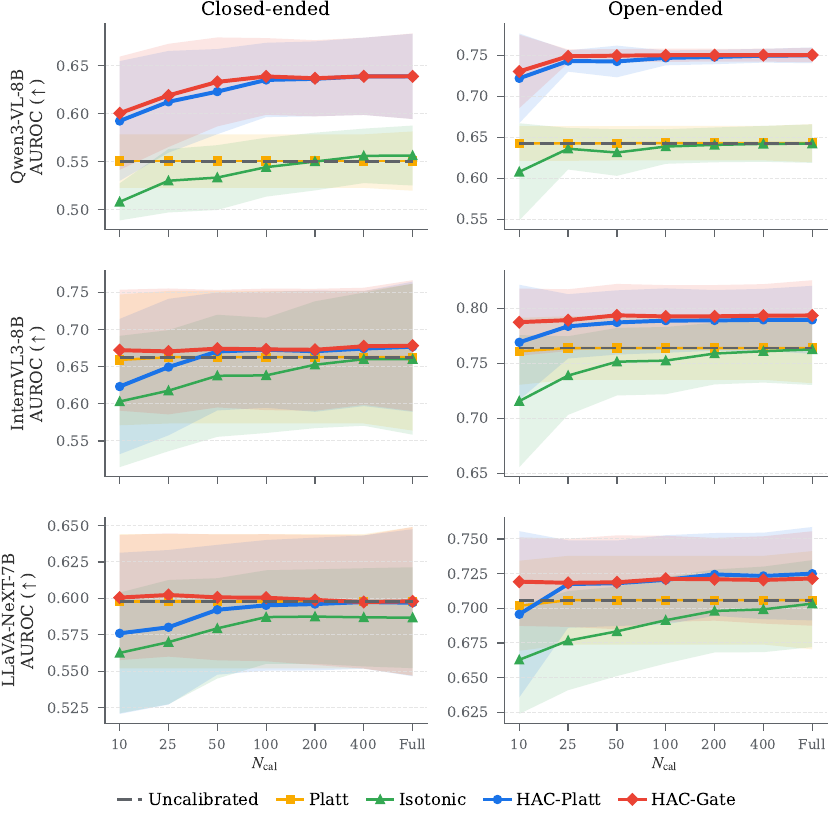}
\caption{Sampling base confidence}
\label{fig:calsize_auroc_sampling}
\end{subfigure}

\vspace{0.6em}

\begin{subfigure}[t]{0.68\textwidth}
\centering
\includegraphics[width=\textwidth]{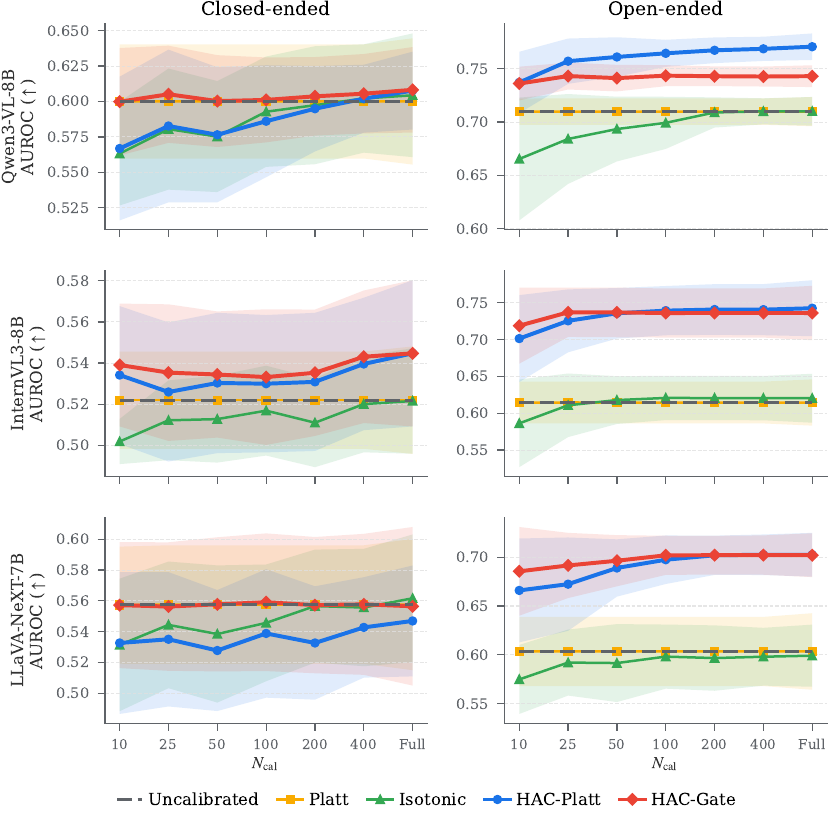}
\caption{Verbalized base confidence}
\label{fig:calsize_auroc_verbalized}
\end{subfigure}
\caption{AUROC ($\uparrow$) versus calibration-set size $N_{\text{cal}}$, per model (rows) and question type (columns), pooled over the three datasets. Uncalibrated and Platt share the same AUROC, so their curves coincide and are flat in $N_{\text{cal}}$.}
\label{fig:calsize_auroc}
\end{figure}

\begin{figure}[t]
\centering
\begin{subfigure}[t]{0.68\textwidth}
\centering
\includegraphics[width=\textwidth]{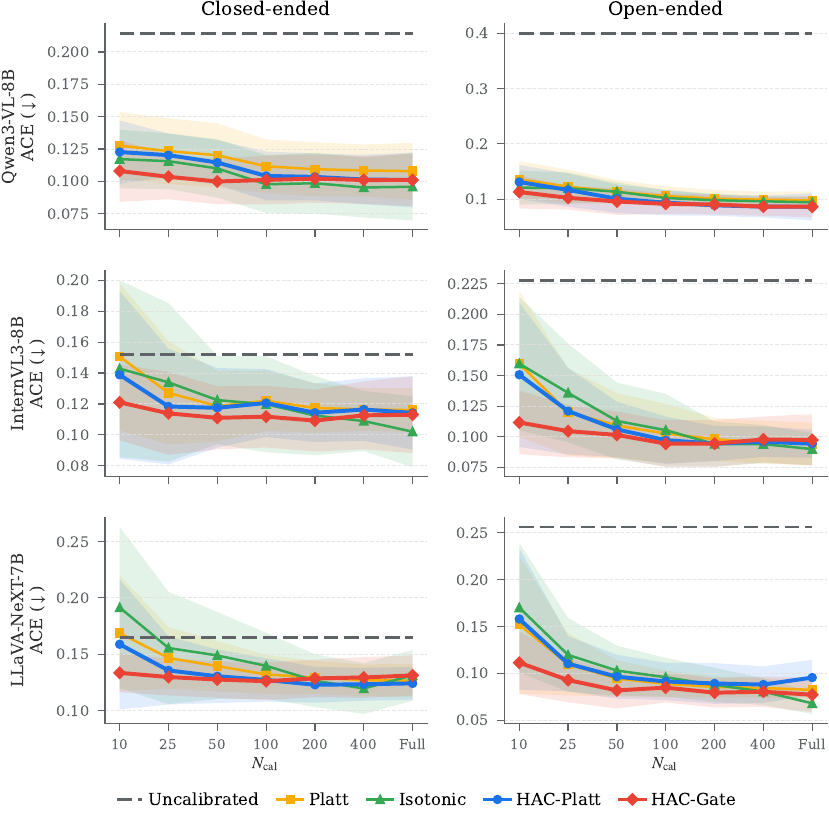}
\caption{Sampling base confidence}
\label{fig:calsize_ace_sampling}
\end{subfigure}

\vspace{0.6em}

\begin{subfigure}[t]{0.68\textwidth}
\centering
\includegraphics[width=\textwidth]{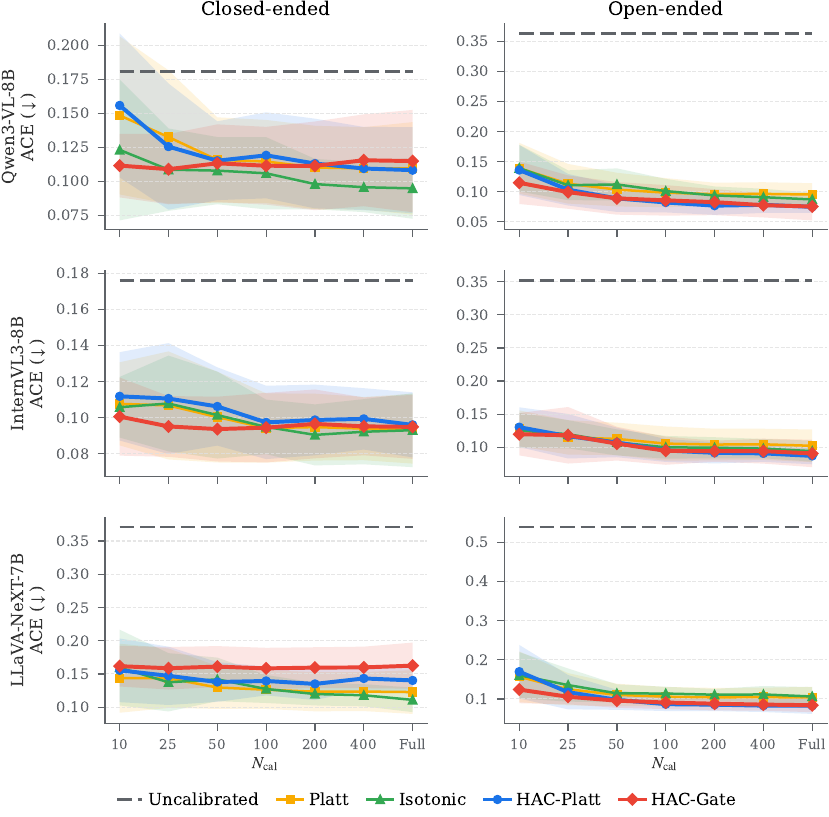}
\caption{Verbalized base confidence}
\label{fig:calsize_ace_verbalized}
\end{subfigure}
\caption{Adaptive Calibration Error (ACE, $\downarrow$) versus calibration-set size $N_{\text{cal}}$, per model (rows) and question type (columns), pooled over the three datasets. The dashed line is the uncalibrated baseline.}
\label{fig:calsize_ace}
\end{figure}

We subsample each fold's calibration partition to
$N_{\text{cal}}\in\{10,25,50,100,200,400,\text{Full}\}$, refitting every calibrator on the subsample while holding the test fold fixed, and report
AUROC (Figure~\ref{fig:calsize_auroc}) and ACE (Figure~\ref{fig:calsize_ace}) for both base confidences.
HAC remains competitive even with very small calibration sets---HAC-Platt and HAC-Gate already exceed the uncalibrated AUROC with only 10 labels, often sufficing by 25--50---and 100--200 samples suffice in most cases, indicating that HAC requires only modest labeled calibration data.

\end{document}